\documentclass[onehalfspace]{easychithesis}

\usepackage{amsmath}
\usepackage{amssymb}
\usepackage{xspace}
\usepackage{epsfig}
\usepackage{graphics}
\usepackage{graphicx}
\usepackage{natbib}
\usepackage{times}
\bibpunct();A{},
\let\cite=\citep

\newcommand{\ignore}[1]{}
\ignore{

}

\DeclareMathOperator*{\argmax}{argmax}
\DeclareMathOperator*{\argmin}{argmin}

\newcommand{\parens}[1]{\left(#1\right)}

\newcommand{\nn}{\nonumber \\}

\newcommand{\expect}[1]{\mathrm{E}\left[#1\right]}

\newcommand{\expectsub}[2]{\mathrm{E}_{#1}\left[{#2}\right]}

\newcommand{\nbrs}{\Gamma}
\newcommand{\dhat}[1]{\widetilde{#1}}

\begin{document}
\title{A Machine Learning Approach to Recovery of Scene Geometry from Images}
\author{Hoang Trinh}
\date{May 2010}
\department{Computer Science}
\division{Physical Sciences}
\degree{Doctor of Philosophy in Computer Science}

\maketitle
\maketitlewithsignature

\topmatter{Abstract}

\indent Recovering the 3D structure of the scene from images yields useful information for tasks such as shape and scene recognition,
object detection, or motion planning and object grasping in robotics.

In this thesis, we introduce a general machine learning approach called unsupervised CRF learning based on maximizing the conditional likelihood.
We describe the application of our machine learning approach to computer vision systems that recover the 3-D scene geometry from images.
We focus on recovering 3D geometry from single images, stereo pairs and video sequences. 
Building these systems requires algorithms for doing inference as well as learning
the parameters of conditional Markov random fields (MRF). Unlike previous
work, our system is trained unsupervisedly without using ground-truth labeled data. 

We employ a slanted-plane stereo vision model in which we use
a fixed over-segmentation to segment the left image into coherent regions called superpixels.
We then assign a disparity plane for each superpixel. We formulate the problem of
inferring plane parameters as an MRF labelling problem, which can be solved by an energy
minimization method. The MRF is a graphical model in which superpixels define nodes
and the adjacency between superpixels define edges. Our stereo energy function balances
between a data matching term and a smoothness term.

For systems with continuous valued variables, or
discrete-valued variables with very large state space, it is impossible to directly use a standard
discrete MRF inference techniques such as Loopy BP, graph cuts or tree-reweighted message passing.
For such systems,
we propose to use a generic Particle-based Belief Propagation (PBP) algorithm closely related
to previous work, which we then formulate specifically for our MRF labeling problems.
Although we only describe a specific use of
this generic PBP algorithm, we believe it can be used as an approximate inference scheme
for a wide variety of problems that can be formulated by a probabilistic graphical model,
especially those containing many random variables with very large or continuous domains.

We demonstrate the use of our unsupervised CRF learning algorithm for a parameterized slanted-plane stereo vision model involving
shape from texture cues. This unsupervised learning algorithm implicitly trains
shape from texture monocular surface orientation cues. We exhibit that training monocular cues from stereo pair data improves stereo depth estimation. Our stereo model with texture cues, only by unsupervised training, outperform the results in related work on the same stereo dataset.

Our unsupervised learning method is also implemented for the monocular depth estimation (MDE) problem.
The MDE model, learned using stereo pairs only, demonstrates a modest improvement after a few training steps, and achieve performance comparable to previous work on the same dataset. The use of MDE in combination with the dense stereo model also introduces a small boost in depth estimation over the initial stereo model.

In this thesis, we also address the use of stereo video sequences.
We formulate structure and motion estimation as an energy minimization problem, in which the model is an extension of our slanted-plane stereo vision model that also handles surface velocity.
Surface estimation is done using our own slanted-plane stereo algorithm. Velocity estimation is achieved by solving an MRF labeling problem using Loopy BP. 
Performance analysis is done using our novel evaluation metrics based on the notion of view prediction error.
Experiments on road-driving stereo sequences show encouraging results.

\topmatter{Acknowledgments}

It has been an honor for me to be advised by Prof. David McAllester, who has been giving me his outstanding mentorship, his patience and unconditional support from the first day to the last day of my graduate program at TTI-C. Through him, I also got to understand more about how to think and act as a true scientist. I would never have come this far without his guidance and encouragement.

Many thanks go to my thesis committee members: Greg Shakhnarovich, Raquel Urtasun, Ronen Basri for their valuable and constructive advices and feedbacks.

During the last few years, I have received a lot of help and support from people in TTI-C, CS Department and elsewhere that I want to give thanks to: 
Pedro Felzenszwalb for his inspiration that fired up my passion in Computer Vision; Alex Ihler, Ronen Basri, Deva Ramanan, Xiaofeng Ren, Cristian Sminchisescu, Yali Amit, John Langford for helping me find answers to lots of my questions and puzzles; Wonseok Chae, Andy Cotter and Karthik Sridharan for being my friend and making me feel less alone as a TTI-C student; Gary Hamburg, Carole Flemming, Christina Novak, Adam Bohlander, Julia MacGlashan, Dawn Gardner and others, who have created a family-like environment and atmosphere at TTI-C, where I consider my second home away from home during the last few years; Nathan Srebro and Umut Acar who gave me advices as Academic advisors; Seiichi Mita and TTI-J Nagoya for welcoming me and giving an warm experience in Nagoya; Mr. Kien Pham and Adam Kalai, through whom I got to know about TTI-C and then became one of the first students here; Robinson, KC Lee, Katharine, Paul King, Dan Prochaska for their support and guidance during my summer internship.
Special thanks go to my friends and my family. 
Last but not least, I want to thank my parents and my beloved $Nhung$.

This thesis was partly funded by a grant from the Vietnam Education Foundation (VEF). 

%
%

\tableofcontents

\listoffigures

\listoftables

\mainmatter
\doublespacing

\chapter{Introduction}
\label{ch:intro}

\indent One of the most important characteristic of natural scenes is their underlying 3D geometric structure. Recovering the 3D structure of the scene from images yields a great amount of useful information for other computer vision tasks such as shape and scene recognition, object detection, or motion planning and object grasping in robotics. 
In this thesis, we focus mainly on the problem of scene geometry recovery and motion estimation.
Concretely we want to solve the problems of monocular depth estimation, stereo vision with monocular cues, and structure and motion estimation from stereo video.
We attempt to solve these problems at different levels, using different forms of input image data and different types of image features.

Markov random fields (MRFs) are a very powerful tool used for modeling a wide range of Vision problems.
Throughout this thesis, we will demonstrate that all the problems we are interested in can be modeled using MRFs or conditional random fields (CRFs), a variant of MRFs. A major part of this thesis consists of formulating general learning and inference algorithms for MRFs. The probabilistic inference and learning techniques we develop to solve our specific Vision problems can also be generalized to other problems modeled by MRFs and CRFs.

\section{Markov Random Fields and Related Formalisms}
\label{sec:MRFCRF}

\subsection{MRF}

An MRF is considered a particular type of graphical model that allows us to efficiently compute marginal distributions.
One representation of an MRF is a factor graph, or hypergraph, in which each hyperedge can connect more than one (regular) node.
MRFs have been widely used for decades as statistical models in a variety of
application areas \cite{HCTheorem,MRF,MRFVision}.  An MRF defines a probability distribution on the
joint assignments to (configurations of) a set of random variables.

Let $Y = (y_1, \ldots, y_N)$ be a set of $N$ random variables. Let $Y_1, \ldots, Y_M$ be $M$ subsets of Y. Let $F = (f_1, 
\ldots, f_M)$ be factors where $f_{\alpha}$ is a function on assignments of values to the variables in $Y_{\alpha}$. 

\begin{eqnarray}
\label{eq:pwMRF}
p_{\beta}(y) = \frac{1}{Z_{\beta}}\prod_{\alpha = 1}^{M} f_{\beta,\alpha}(y_{\alpha})
\end{eqnarray}

where 

\begin{eqnarray}
\label{eq:partitionMRF}
Z_{\beta} = \sum_{y} \prod_{\alpha = 1}^{M} f_{\beta,\alpha}(y_{\alpha})
\end{eqnarray}

Here $y$ is an assignment of values to all random variables in $Y$, each $y_{\alpha}$ is the projection of $y$ onto the variables in the subset $Y_{\alpha} \subseteq (y_1, \ldots, y_N)$. $\beta$ is the mode parameter. Each function $f_{\alpha}$ is called a factor or sometimes called a local function, and is defined only on the subset $Y_{\alpha}$. The set of variables $Y$ and the set of factors $F$ define the MRF.

For inference in MRFs, the goal is to find the assignment of values to all variables maximizing the joint probability:

\begin{eqnarray}
\label{eq:pwMRFinf}
y^{*} = \argmax_{y}(p_{\beta}(y))
\end{eqnarray}

For training, given the i.i.d sampled training data: ${y^1, \ldots, y^T}$, the objective is to maximize the joint probability over the whole training corpus:

\begin{eqnarray}
\label{eq:pwMRFtrain}
\beta^{*} = \argmax_{\beta} \prod_{t = 1}^{T} (p_{\beta}(y^t))
\end{eqnarray}

The most popular version of MRFs in computer vision is the pairwise MRF model, in which each factor is defined over at most two random variables. 

\subsection{CRF}

One notable variant of an MRF is a conditional random field (CRF), introduced by Lafferty et al. in \cite{Pereira01}. 
While MRFs model joint probabilities, CRFs model conditional probabilities.  
Similar to an MRF, a CRF is an undirected graphical model representing random variables and dependencies between random variables. 
In a CRF the variables are divided into two groups --- exogenous and dependent.  
A CRF defines a conditional probability of the dependent variables given the exogenous variables and (importantly) does not model the distribution of the exogenous variables. The advantage is that CRFs can estimate the distribution of interest more accurately than MRFs, since there is no danger of corrupting the model by modeling the exogenous variables poorly.
For instance, in the case of stereo vision one might expect that it is easier to model the probability distribution of the right image given the
left image than to model a probability distribution over images.

Let $x = (x_1, \ldots x_N)$ represent a set of exogenous variables. In a CRF, we want to compute the conditional probability distribution of the dependent variables $y$ with respect to the exogenous variables $x$.

\begin{eqnarray}
\label{eq:pwCRF}
p_{\beta}(y|x) = \frac{1}{Z_{\beta}(x)}\prod_{\alpha = 1}^{M} f_{\beta, \alpha}(y_{\alpha}, x_{\alpha})
\end{eqnarray}

where the partition function $Z_{\beta}(x)$ is defined in a similar way to (\ref{eq:partitionMRF}).

The CRF inference equation:

\begin{eqnarray}
\label{eq:pwCRFinfer}
y^* = \argmax_{y} p_{\beta}(y|x)
\end{eqnarray}

The CRF training equation:

\begin{eqnarray}
\label{eq:pwCRFtrain}
\beta^{*} = \argmax_{\beta} \prod_{t = 1}^{T} (p_{\beta}(y^t|x^t))
\end{eqnarray}


\subsection{Hidden CRF}

A variation of CRFs is the hidden CRFs (HCRF) proposed by Quattoni et al. in \cite{Quattoni07}. In HCRFs, additional latent variables $Z = (z_1, \ldots z_K)$ are introduced, which are not observed in the training data, but are part of the model.

\begin{eqnarray}
\label{eq:hCRF}
p_{\beta}(y, z|x) = \frac{1}{Z_{\beta}(x)}\prod_{\alpha = 1}^{M} f_{\beta, \alpha}(y_{\alpha}, x_{\alpha}, z_{\alpha})
\end{eqnarray}

where the partition function $Z_{\beta}(x)$ is defined in a similar way to (\ref{eq:partitionMRF}).

For inference, HCRFs predict the dependent variable $y$ by computing $p(y|x)$ by marginalizing the hiden variable $z$. The HCRF inference equation:

\begin{eqnarray}
\label{eq:hCRFinfer}
y^* & = & \argmax_{y} p_{\beta}(y|x) \\
\nn
& = & \argmax_{y} \sum_z p_{\beta}(y, z|x)
\end{eqnarray}
 
In training, given the training data: ${(x^1, y^1), \ldots, (x^T, y^T)}$ where $x$ the exogenous variable, and $y$ is the dependent variable, the objective of HCRF is to maximize the following conditional probability:

\begin{eqnarray}
\label{eq:hCRF1}
\beta^* = \argmax_{\beta} \prod_{t = 1}^{T} p_{\beta}(y^t|x^t)
\end{eqnarray}
 
where:

\begin{eqnarray}
\label{eq:hCRF2}
p_{\beta}(y|x) = \sum_{z} p(y,z|x)
\end{eqnarray}

\subsection{Unsupervised CRF}
\label{subsec:uCRF}

We introduce another variant of the CRF, which we call unsupervised CRF (UCRF). 
In unsupervised learning one usually formulates a parameterized probability model and seeks parameter values maximizing
the likelihood of the unlabeled training data. Our approach to unsupervised learning is based on maximizing the conditional likelihood.

The defining equation of UCRF is similar to HCRF up to some changes in variable naming. Here $x$ and $u$ denote the exogenous variables and $y$ is the dependent variable.
In the stereo vision example, $x$ and $u$ correspond to the left and right images of the stereo pair, while $y$ is the hidden-state variable corresponding to the depth map of the left image.

\begin{eqnarray}
\label{eq:uCRF}
p_{\beta}(u, y|x) = \frac{1}{Z_{\beta}(x)}\prod_{\alpha = 1}^{M} f_{\beta, \alpha}(u_{\alpha}, y_{\alpha}, x_{\alpha})
\end{eqnarray}

where the partition function $Z_{\beta}(x)$ is defined in a similar way to (\ref{eq:partitionMRF}).

For inference, UCRF aims at predicting $y$ by computing the following:

\begin{eqnarray}
\label{Eq:uCRFinfer}
y^* = \argmax_{y} p_{\beta}(y|x, u)
\end{eqnarray}

or

\begin{eqnarray}
\label{Eq:uCRFinfer2}
y^* = \argmax_{y} p_{\beta}(y|x)
\end{eqnarray}

Note that (\ref{Eq:uCRFinfer}) looks at both $x$ and $u$, while (\ref{Eq:uCRFinfer2}) only looks at $x$, which in this case $u$ can be considered latent variable. Later we will show that (\ref{Eq:uCRFinfer}) is applied to our stereo inference algorithm, while (\ref{Eq:uCRFinfer2}) is applied to inference in our monocular depth estimator.

In training, given the training data: ${(x^1, u^1), \ldots (x^T, u^T)}$, the objective is to find the set of parameters maximizing the conditional probability of arbitrary random variables, both are observed by the training data. 

\begin{eqnarray}
\label{TrainingEq}
\beta^* = \argmax_{\beta}\;\;\prod_{t=1}^T\;p_{\beta}(u^t|x^t)
\end{eqnarray}

where:

\begin{eqnarray}
\label{eq:uCRF2}
p_{\beta}(u|x) = \sum_{y} p(u, y|x)
\end{eqnarray}

Note that in HCRF, the test time dependent variable $y$ is part of the training data, while the latent variable $z$ is not. In UCRF, the latent variable $u$ is part of the training data, however the test time dependent variable $y$, which needs to be predicted during inference, is not included in the training data.

We will then introduce a hard EM learning algorithm for maximizing the conditional likelihood, where we use our Particle-based Belief Propagation inference algorithm to perform the hard E step, and contrastive divergence to update model parameters in the hard M step. Details are covered in Section~\ref{sec:introlearning}.

Throughout the thesis, we emphasize the employment of unsupervised CRFs learning for our vision models. Section~\ref{sec:introlearning} will succeed this discussion with more details about the components of our learning approach.

A main portion of the thesis involves applying our unsupervised CRF model for the scene geometry estimation problem, in particular the stereo vision problem. Here stereo vision provides a simple setting to investigate unsupervised learning and hence seems a good place to start.
However we believe that our approach to unsupervised learning based on maximizing conditional likelihood can be generalized to other, maybe much more sophisticated models.

\section{MRFs and CRFs for Our Vision Problems}
\label{sec:CRF4Vision}

Many problems in computer vision can be interpreted as pixel-labeling problems: the objective is to assign labels to pixels or groups of pixels. For instance, in two-view stereo vision the labels can be disparity values, in segmentation the labels are indices of different groups that the pixels belong to, etc.

Pixel-labeling problems are usually solved by minimizing an energy function having two terms: a data term penalizing assignments that are inconsistent with the observed data, and a smoothness term enforcing local smoothness (spatial coherence). \cite{geman84} have pointed out that optimizing an energy function of this form is equivalent to estimating the maximum a posteriori probability of an MRF.

\begin{figure*}[t] \centering
\begin{center}
 \includegraphics[width=0.95\textwidth]{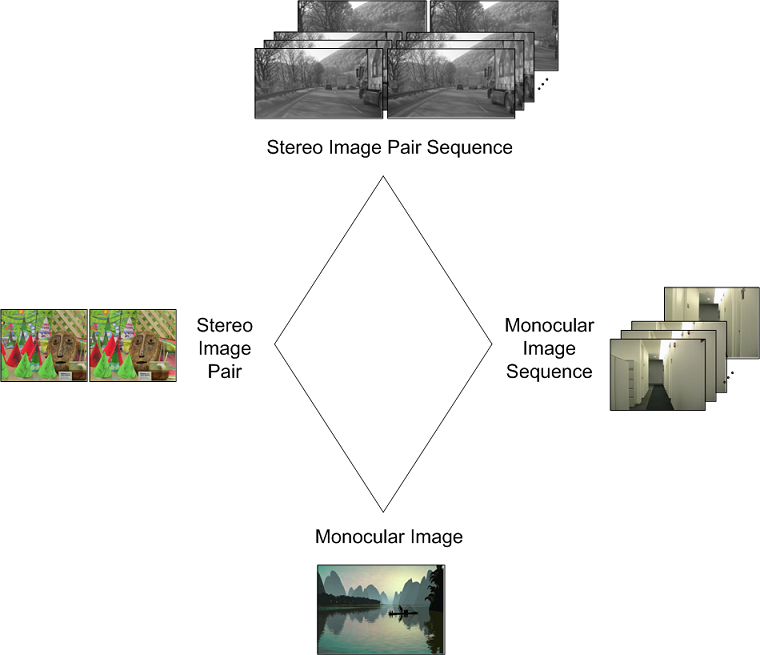}
\end{center}
\caption{Different forms of input image data for recovering scene geometry.}
\label{DataTypes}
\end{figure*}

In this thesis we mainly focus on the problem of recovery of 3D scene structure. 
Scene structure can be recovered from different forms of input image data, ranging from a single image to a multiple view video sequence. The list of most commonly used forms of data is shown in Figure~\ref{DataTypes}. 
All of these image data forms are investigated in this thesis, with main focus on single images, stereo image pairs and stereo image pair sequences. 
Since images are 2-D representations created by perspective projection of 3-D scenes, there is a natural intrinsic ambiguity for recovering 3-D information from local image features. Increasing the amount of input image data can reduce the difficulty of the problem. 

In vision systems, depending on the form of image data used, different types of energy terms may arise. For the vision systems addressed in this thesis, the following types of energy terms are exploited:

\begin{itemize}
	\item Monocular terms: depending on a single image, (e.g: intensity, gradient, SIFT, HOG, etc.)
	\item Stereo terms: involving a stereo pair of images.
	\item Motion terms: involving two consecutive video frames.
	\item Smoothness terms: involving only the output pixel-labeling variables.
\end{itemize}

As we mentioned earlier, many computer vision problems can be interpreted as pixel-labeling problems. The types of the output labels (dependent variables) are defined by what the vision problem is trying to solve specifically, i.e. the quantities of interest that we want to infer about the underlying scene. An image denoising algorithm would want to assign an intensity value or a color to each pixel, while a stereo algorithm would want to assign a disparity value. In this thesis, our output label types of interest are:

\begin{itemize}
  \item 3D point locations: for sparse Structure from Motion.
	\item Disparity values: for monocular depth estimation and dense stereo depth estimation. 
	\item Disparity plane parameter vectors: for slanted-plane stereo model with monocular surface orientation cues.
	\item Disparity plane parameter vectors and motion vectors: for structure and motion estimation from video sequences.
\end{itemize}
 


Any vision problem we address in this thesis can be formulated by a combination of the subsets of each of the three elements above (input image data, types of energy terms, types of output labels). However, in any case, the problem can still be modeled by an MRF.

\section{Our CRF Formulations for Scene Geometry Recovery Problems}

\subsection{Unsupervised CRF Formulation for Stereo Vision}

A major part of the thesis focuses on the problem of recovering scene structure from stereo image pairs, with emphasis on monocular surface orientation cues - evidence of surface orientation based on a single image. For the current discussion, let's assume we are not using the monocular cues.

A simple version of the scene structure recovery problem is dense two-view stereo vision, in which the depth of each pixel is recovered from a stereo image pair taken from a stereo camera. The depth of each pixel can be related to the horizontal disparities between the two stereo images. A dense output is returned - a disparity is assigned to each pixel in the reference image. 
In this problem, the exogenous variable $x$ is the left image, the dependent variable $y$ is the disparity value assignment to all pixels in $x$, and the exogenous variable $u$ is the right image.
A thorough, although not very recent, survey of state of the art dense stereo vision approaches was described in \cite{Sze02}. An online evaluation of the best two-image stereo matching algorithms, along with test images, ground truth, and software can be found in \cite{scharstein01}.

The shift from the dense stereo model to slanted-plane stereo model started with the work by BirchField and Tomasi in \cite{BirchfieldTomasi}. Since then, quite a few other stereo algorithms have been applying this slanted-plane model, among which are state-of-the-art algorithms in the current Middlebury benchmark \cite{KSK06, WangZheng08, yang08}. The model uses the assumption that the scene is composed of a set of planar surfaces. 
The algorithms using this model start by segmenting the image into homogeneous regions called superpixels, using image segmentation. The superpixels found by the segmentation method are coherent groupings of pixels having similar properties (color, texture, etc). Therefore it is just natural to assume that each superpixel corresponds to a planar surface in the scene, which from now on we will call disparity plane. Next, the goal of the stereo algorithm is to infer the location and orientation (or surface normal) of each of these planes - which are fully defined by the 3 plane parameters. After that, the disparity of each pixel can be directly computed from the disparity map equation, with sub-pixel precision.

Our stereo vision model is a slanted plane model in which we use a fixed over-segmentation of the left image into superpixels. The goal of inference is to assign a disparity plane for each superpixel.
In other plane-based stereo algorithms, the authors mostly use a local matching step to extract reliable correspondences, followed by robust plane fitting to fit a disparity plane into each superpixel separately. The process is then repeated for iterative plane refinement.
Our stereo inference algorithm also infers a disparity plane for each superpixel. However we use the Particle BP inference algorithm to simultaneously find the optimal joint assignment of all plane parameters to all superpixels. The optimal assignment is found by minimizing a high-dimensional global energy function. In this Particle BP inference algorithm, each particle, which represents a proposal disparity plane for a specific superpixel, gets updated over time, instead of being fixed after the plane fitting step, as in other slanted-plane stereo algorithms. Chapter~\ref{pbp-inference} will discuss this algorithm in details.

We formulate the problem of inferring plane parameters as an UCRF. The UCRF is a graphical model in which superpixels define nodes and the adjacency between superpixels define edges.  
In this case, the input image data are stereo image pairs; the energy terms involved are stereo and smoothness terms, and the hidden labels here are the parameters of disparity planes corresponding to superpixels.

In this problem, the exogenous variable $x$ is the left image, the dependent variable $y$ is the disparity plane assignment to all superpixels in $x$, and the exogenous variable $u$ is the right image. $x$ and $u$ are in the training data, while $y$ is not.
 


\subsection{Unsupervised CRF Formulation for Stereo Vision Models with Monocular Cues}

Our stereo vision model is a slanted plane model involving monocular shape from texture cues. The introduction of an additional monocular term into the slanted-plane stereo model is a novel idea that has not been investigated.

One interesting point is that monocular cues can easily be incorporated into a stereo model and can be used as an additional monocular term.
The monocular depth cues predicting pixel disparity values directly can be added into a standard dense stereo model, without breaking the UCRF formulation of the model. For this UCRF, $x$ and $u$ form the stereo pair, and $y$ is the assignment of disparity values to all pixels in $x$.
Analogously, the surface orientation cues predicting the surface normal of superpixels can be integrated into our existing slanted-plane stereo model as an additional energy term. For this UCRF, $x$ and $u$ is the stereo pair, $y$ is the assignment of disparity planes to all superpixels in $x$.

Our UCRF models now involve three terms: a data matching energy (stereo term)
measuring the degree to which the left and right images agree under the induced correspondence, 
a smoothness energy (smoothness term) measuring the local smoothness of the assigned scene structure, and a texture
energy (monocular term) measuring the degree to which the assigned labels agree with the labels predicted by 
the monocular predictor.

\subsection{Unsupervised CRF Formulation for Depth Estimation from a Single Image}

Recently, there has been more interest in the problem of estimating 3-D structure from one single image. While this is a very interesting problem, it remains extremely more challenging than the stereo vision problem, due to the high ambiguity between local image features and the 3-D point locations, due to perspective projection. Some methods such as Shape from Shading \cite{zhang99} make specific assumptions about the scene and the formation of the image, and rely mostly on photometric cues such as lighting and shading to recover shapes of surfaces in the scene. Several more efforts by Hoiem \cite{Hoiem05} and Saxena \cite{ashu07} have demonstrated quite encouraging results. Both of these approaches focus on supervised learning to capture the relationship between the local image features and geometric properties (orientation, location) of regions of the scene, as well as the relationship between different regions. 
In this thesis we use UCRF learning to train a monocular depth estimator.
For MDE, we model the disparity of a pixel as a linear function of the monocular feature vector and a parameter vector. The objective of the learning algorithm is to train this monocular feature vector.
A monocular energy term is introduced, penalizing the difference between the assigned disparity and the monocularly predicted disparity.
An energy function formulated by combining this monocular energy term with a smoothness term itself defines an UCRF.
In this UCRF, the exogenous variable $x$ is the left image, or its feature-based representation, the dependent variable $y$ is the disparity map for all pixels in $x$, and there is no latent variable $u$.

For inference, we use equation (\ref{Eq:uCRFinfer}), i.e, we want to infer the depth map $y$ only by looking at the left image $x$.
Training is formulated by incorporating the monocular energy term into a standard pixel-based UCRF stereo model. At each training step, we do inference on this joint model to ontain the new depth map, then we use this updated depth map to retrain the MDE. We then implement our UCRF learning algorithm for monocular depth estimation for learning the monocular parameter vector using stereo image pairs only. 
The experimental results show that the monocular model is able to provide good depth predictions from different scenes in
general, although there were still image parts which were not generalized well, i.e, it tends to perform more poorly in predicting depth for image patches that were too different from those in the training data. 

We also discuss in this thesis the possibility of using histograms of oriented gradients (HOG) feature as monocular surface orientation cues.  
We derive a formal relationship between a
variant of HOG features and surface orientation. Although our observation that there should
be a statistical relationship between HOG features and surface orientation is a simple result
in the area of shape from texture, HOG features have only recently gained
popularity and to our knowledge the idea of using HOG as a surface orientation cue
has not been previously noted.

\subsection{CRF Formulation for Structure and Motion Estimation from Stereo Videos}

Multi-view video sequence is arguably the richest form of image data for scene reconstruction. A subtype in this form that we consider very interesting are stereo sequences. This data form is interesting for two reasons. First, as opposed to a multi-view video sequence, a stereo sequence can easily be captured using only one moving calibrated binocular camera. Second, this data form provides us with enough constraints to conveniently compute depth and motion of scene points simultaneously, since coupling dense stereo matching with motion estimation helps decrease the number of unknowns per image point. More specifically, for stereo sequences, we can formulate structure and motion estimation as an energy minimization problem, in which the model is either an extension of a stereo vision model to also handle scene motions, or an extension of a Structure from Motion or optical flow model under a stereo setup. So far, to our knowledge, this useful data form has hardly been investigated and exploited for the purpose of recovering scene structure and motion. One of the very few research work that has discussed this issue is \cite{huguet07}, in which the authors presented a variational method for scene flow estimation from a calibrated stereo image sequence.
In Chapter~\ref{depth-motion}, we extend the constructed slanted-plane stereo model to handle
structure and motion estimation on road-driving stereo sequences. Based on specific assumptions about the motion
of the camera and the scene, we can reduced the 2D optical flow problem to a 1D velocity value problem.
Our algorithm iteratively and alternately solve for structure (disparity planes) and motion (velocity values). 
Surface estimation is done using our slanted-plane stereo algorithm. 
Velocity estimation is achieved by solving a CRF labeling problem using Loopy BP. 
Experiments on road-driving stereo sequences showed encouraging results. 
Performance analysis was done using our novel evaluation metric based on the notion of view prediction error.

For this CRF, $x$ is the left frame of the stereo pair at time $t$, $u$ is composed of the right frame at time $t$ and the left frame at time $t+1$, and $y$ is the assignment of disparity planes and velocity values to all superpixels in the left frame at time $t$. The dependent variable $y$ is not known in the training video sequences.

%

\section{Training Methods for Unsupervised CRFs}
\label{sec:introlearning}

As mentioned earlier in Section~\ref{subsec:uCRF}, our unsupervised CRF learning approach is based on maximizing conditional likelihood. This section covers more details about our learning approach.
Here we introduce a hard EM algorithm for maximizing the conditional likelihood, where we use contrastive divergence to update model parameters in the hard M step.

\subsection{Hard Conditional EM}

Depending on the specific use of the model, (\ref{eq:uCRF2}) can be further factorized as follows when necessary:

\begin{eqnarray}
\label{TrainingEq3}
P_{\beta}(u|x) = \sum_y P_{\beta}(u,y|x) = \sum_y P_{\beta_y}(y|x)P_{\beta_u}(u|x,y)
\end{eqnarray}

The conditional probability at the right hand side of (\ref{Eq:uCRFinfer}) can be represented as follows.

\begin{eqnarray}
\label{Eq:uCRF-test}
P_{\beta}(y|x, u) = \frac{P_{\beta}(y, u|x)}{P_{\beta}(u|x)} = \frac{P_{\beta}(y, u|x)}{ \sum_y P_{\beta}(u,y|x)}
\end{eqnarray}

For soft conditional EM, the goal of the E step is to estimate the probability distribution over the latent variables given the training data - this distribution can be interpreted using equation (\ref{Eq:uCRF-test}), while the goal of the M step is to maximize the model over complete data. 
Equation (\ref{Eq:uCRF-test}) justifies for our argument that the most likely value of the dependent variables given the model and the observed data, which is the result of the inference step, is also the value maximizing the conditional likelihood.
Note that the transformation from (soft) conditional EM to hard conditional EM is done by replacing the sum in equation (\ref{eq:uCRF2}) by the max. Hard EM works with the single most likely value of $y$ instead of the distribution over $y$.

Hard conditional EM is defined to be the process of iterating the following updates:

\begin{eqnarray}
\label{harde1}
y_i & := & \argmax_y P_{\beta}(u_i,y|x_i) \\
\nn
\label{hardm1}
\beta & := & \argmax_{\beta}\; \sum_{i=1}^N \ln P_{\beta}(u_i,y_i|x_i) 
\end{eqnarray}

We will call (\ref{harde1}) the hard E step and (\ref{hardm1}) the hard M step. 
We implement the hard E step using our Particle-based Belief Propagation inference algorithm that we introduce next in Section~\ref{sec:introinference}. For the M step, we use a learning technique called contrastive divergence, more details of which will follow in Sections~\ref{subsec:CD} and~\ref{sec:CD}.
These two steps are the foundation to the hard conditional EM algorithm that we will elaborate later in Chapter~\ref{uspv-depth-lrn}.

\subsection{Contrastive Divergence}
\label{subsec:CD}

In order to learn the parameters of our unsupervised CRF model, we use contrastive divergence in the hard M step of our hard EM algorithm. Contrastive divergence first introduced by G. Hinton et al. in \cite{contrastiveA,contrastiveB} is a general MRF learning algorithm capable of training large models. Other work that also described learning on MRF using contrastive divergence is the Field of Experts system by Roth and Black in \cite{roth05}.

Briefly speaking, contrastive divergence is a tool that allows us to estimate the gradient of a energy function, even though we cannot evaluate the energy function itself. Using contrastive divergence in the hard M step allows us to compute the gradient of the energy function with respect to the model parameters $\beta$. We can then update $\beta$ using gradient descent.
By running only a few (one or two) MCMC steps, starting from the current estimation of the latent variable, contrastive divergence reduces significantly both the computation and the variance compared to running standard MCMC processes. However, since this is an approximate method (with finite number of samples), there is actually no known guarantee for convergence.


\subsection{Related work}

The most closely related work seems to be that of Zhang and Seitz \cite{Seitz05}.
They give a method for adapting five parameters of a stereo vision model
including the weights for the match and smoothness energies as well as robustness parameters. The five
parameters are tuned to each individual input stereo pair, although the method could be used
to tune a single parameter setting over a corpus of stereo pairs.
The main difference between their work and ours is that we train highly
parameterized monocular depth cues. Another difference is that we formulate a
general CRF-like model for unsupervised learning based on maximizing conditional likelihood
and avoid the need
for the independence assumptions used by Zhang and Seitz by using contrastive divergence --- a general
method for optimizing loopy CRFs \cite{contrastiveA,contrastiveB}.

There is also related work on learning highly parameterized monocular depth cues by Saxena et al. in \cite{ashu07,andrew07}, as well as Hoiem et al. in \cite{Hoiem05}.
The main difference between these methods and ours is that we use unsupervised learning while they
use ground truth data to train their system.  One might argue that stereo pairs constitute supervised
training of monocular depth cues.  A standard stereo depth algorithm could be used to infer a depth map for each pair
which could then be used in a supervised learning mode to train monocular depth cues.  However, we demonstrate that
training monocular depth cues from stereo pair data improves {\em stereo} depth estimation. Hence the method can be
legitimately viewed as unsupervised learning of a stereo depth. 
We also demonstrate later that our stereo model with texture cues, only by unsupervised training, outperformed the results in \cite{andrew07}.


Other related work includes that of Scharstein and Pal \cite{scharstein07} and Kong and Tao \cite{Kong04}.
In these cases somewhat more highly parameterized stereo models are trained using methods developed for general CRFs. 
However, the training uses ground truth depth data rather than unlabeled stereo pairs.

\section{Inference using Particle-based Belief Propagation}
\label{sec:introinference}

Inference problems arise from various scientific areas such as AI, statistical physics, computer vision, coding theory. 
In MRFs,the goal of inference is to find the values of the latent variables achieving the maximum a posteriori of the MRF.
Inference is also a vital component of MRFs learning algorithms in general, and of our learning approach in particular. By performing inference, we want to estimate the most likely value of the latent variables given the model and the observed data, which can be formulated by the following equation.

\begin{eqnarray}
\label{hardInference}
y^* & := & \argmax_y p_{\beta}(y|x,u)
\end{eqnarray}

From equation (\ref{Eq:uCRF-test}), we can see that (\ref{hardInference}) is equivalent to (\ref{harde1}),
which is exactly the hard E step in our learning algorithm described in Chapter~\ref{uspv-depth-lrn}. In that chapter we will show that the inference step minimizes an energy function defined by our MRF model.

Recent energy minimization techniques such as Loopy Belief Propagation (LBP) \cite{pearl88, yedidia00} or graph cuts \cite{boykov01} have shown superior performance on MRFs than were previously possible. According to the widely used Middlebury stereo benchmarks \cite{Sze02, scharstein01}, almost all the best-performing stereo methods rely on graph cuts or LBP.
However, these methods are known to mainly perform well on systems of many variables, each of which has a relatively small state space (number of possible values), or particularly nice parametric forms (such as jointly Gaussian distributions).
For systems with continuous valued variables, or discrete-valued variables with very large domains, it is impossible to directly use a standard discrete MRF inference techniques such as Loopy BP, graph cuts or tree-reweighted message passing. In our slanted plane stereo model, the MRF is a graphical model in which segments define nodes and the adjacency between segments define edges. Since the inference problem is to assign a disparity plane to segments, the state space of each node is a continuous 3-dimensional vector space. Similarly in our MRF model for Structure from Motion, our graphical model consists of two types of nodes, in which each camera node has 6-D state space, and each map node has a 3-D state space.
For such systems, we propose to use a generic Particle-based Belief Propagation (PBP) algorithm closely related to previous work in \cite{koller99}, which we then formulate specifically for our MRF labeling problems. The popularity of particle filtering for inference in Markov chain models defined over random variables with very large or continuous domains makes it natural to consider sample-based versions of belief propagation (BP) for more general (tree structured or loopy) graphs. 
The algorithm creates an initial set of values for each node, representing samples from the posterior marginal. It then continues to improve the current marginal estimate by constructing a new sampling distribution and draw new sets of particles. 
The consistency of PBP has been theoretically justified in \cite{ihler09}. The authors showed that PBP approaches the true BP messages as the number of samples grows large. They used concentration bounds to analyze the finite-sample behavior, giving a convergence rate of $O(1/\sqrt{n})$ where $n$ is the number of samples, independent of the domain size of the variables.
Our empirical results also show that the algorithm is consistent in finite case, i.e. it converges to the optimal values of the message and belief functions with finite samples. 
PBP also provides estimates of uncertainty in the form of an approximate posterior density function. Although we only described a specific use of this generic PBP algorithm, we believe it can be used as an approximate inference scheme for a wide variety of problems that can be formulated by a probabilistic graphical
model, especially those containing many random variables with very large or continuous domains.

Similar to standard belief propagation, there are also two different versions of implementation for PBP - sum-product and max-product. We use the max-product implementation for the hard E step of our learning algorithm.
We have also implemented the sum-product version for some other vision experiments that we describe later in Chapter~\ref{pbp-inference}. 

\section{Applications to Scene Geometry Estimation}
\label{sec:introexperiments}

The problems we focus on in this thesis are monocular depth estimation, stereo vision with monocular cues, scene structure and motion estimation.
We design and implement our own algorithms solving each of these problems.
Quantitative analysis is a crucial step that helps evaluating the performance of our algorithms.
In this thesis, we demonstrate our experimental results for each of our following vision systems:

\begin{itemize}
	\item In Chapter~\ref{pbp-inference} we present our Particle-based Belief Propagation inference algorithm and some of its applications in vision problems.
	\item In Chapter~\ref{uspv-depth-lrn}, we demonstrate our application of our unsupervised learning algorithm for stereo vision with monocular cues - the stereo model that we describe in Chapter~\ref{str-depth-inf}. The experimental results on two different datasets help validating our learning approach, as well as testing the performance of the final trained stereo model.
  \item In Chapter~\ref{mno-depth-est}, we describe our unsupervised CRF learning approach to monocular depth estimation.
	\item In Chapter~\ref{depth-motion}, we propose a novel evaluation metric, based on the notion of view prediction error. This metric can be used to evaluate the performance of structure and motion estimation algorithms on stereo sequences without ground truth data. Experimental results on road-driving stereo sequences demonstrate that our algorithm successfully improves the view prediction error although it was not designed to directly optimize this quantity.	
\end{itemize}

The thesis is organized as follows.
In Chapter~\ref{pbp-inference} we describe in details the application of our Particle-based Belief Propagation (PBP) inference algorithm to several Vision problems that we address throughout this thesis.
Chapter~\ref{str-depth-inf} present our slanted-plane stereo model with monocular cues, and the use of Histogram of Gradients (HOG) monocular feature as surface orientation cues. 
Chapter~\ref{uspv-depth-lrn} explains our unsupervised CRF learning approach using hard conditional EM. Experimental results of both learning and inference on two different stereo datasets are also demonstrated in this chapter.
We present our learning approach to the problem of monocular depth estimation (MDE) in Chapter~\ref{mno-depth-est}.
The unified framework solving for depth and motion simultaneously is introduced in Chapter~\ref{depth-motion}.
We address conclusions in Chapter~\ref{conclude}.
\chapter{Inference using Particle-based Belief Propagation and Its Applications for Vision Problems}
\label{pbp-inference}

Although the main emphasis of the thesis is on our unsupervised CRF learning algorithm, one key component that takes part in both the training process and the testing process of our learning approach is the inference algorithm.
To prepare the audience with necessary technical details in order to have a better understanding of unsupervised CRF learning, we decide to introduce our inference algorithm earlier in the thesis.

In this chapter, we first review some background technical details based on which we construct our PBP inference algorithm. Next we describe in details our PBP inference algorithm that we mentioned earlier in Section~\ref{sec:introinference}, followed by its applications to the following Vision problems:
\begin{itemize}
	\item Structure from Motion (SfM).
	\item The dense stereo vision problem with unknown camera constraints.
	\item The slanted-plane stereo vision.
\end{itemize}

By experimental results we show proofs of the convergence for the accuracy of such a generic PBP algorithm with finite samples. In addition to accuracy, the PBP algorithm also allows us to estimate and represent state uncertainty, although at some computational cost. In particular, an experiment with a synthetic structure from motion arrangement shows
that its accuracy is comparable with the state-of-the-art Bundle Adjustment (BA) while allowing estimates
of state uncertainty in the form of an approximate posterior density function.

\section{Belief Propagation}
\label{sec:BP}

\subsection{Definitions and Notations}

Let $G$ be an undirected graph consisting of nodes $V=\{1,\ldots,k\}$
and edges $E$, and let $\nbrs_s$ denote the set of neighbors of node
$s$ in $G$, i.e., the set of nodes $t$ such that $\{s,t\}$ is an edge
of $G$.  In a probabilistic graphical model, each node $s\in V$ is
associated with a random variable $X_s$ taking on values in some
domain, ${\cal X}_s$.
We assume that each node $s$ and edge $\{s,t\}$ are
associated with potential functions $\Psi_s$ and $\Psi_{s,t}$ respectively,
and given these potential functions we define a probability distribution over assignments of values to nodes as

\begin{align} \label{eqn:P}
 P(\stackrel{\rightarrow}{x}) &= \frac{1}{Z} \left(\prod_s \Psi_s(\stackrel{\rightarrow}{x}_s)\right) \left(\prod_{\{s,t\}\in E} \Psi_{s,t}(\stackrel{\rightarrow}{x}_s,\stackrel{\rightarrow}{x}_t)\right)
\end{align}

Note that (\ref{eqn:P}) corresponds to the pairwise MRF model.
Here $\stackrel{\rightarrow}{x}$ is an assignment of values to all $k$ variables,
$\stackrel{\rightarrow}{x}_s$ is the value assigned to $X_s$ by $\stackrel{\rightarrow}{x}$, and $Z$ is a
scalar chosen to normalize the distribution $P$ (also called the
partition function).
We consider the problem of computing marginal probabilities, defined by

\begin{equation} \label{eqn:B}
  P_s(x_s) = \sum_{\stackrel{\rightarrow}{x} : \stackrel{\rightarrow}{x}_s=x_s} P(\stackrel{\rightarrow}{x})
\end{equation}

In the case where $G$ is a tree and the sets ${\cal X}_s$ are small,
the marginal probabilities can be computed efficiently by belief
propagation~\citep{pearl88}. This is done by computing messages
$m_{t\rightarrow s}$ each of which is a function on the state space
of the target node, ${\cal X}_s$. These messages can be defined
recursively as
\begin{equation} \label{eqn:m}
  m_{t\rightarrow s}(x_s)  = \sum_{x_t\in {\cal X}_t} \; \Psi_{t,s}(x_t,x_s) \Psi_t(x_t) \prod_{u\in \nbrs_t\setminus s} m_{u\rightarrow t}(x_t)
\end{equation}
When $G$ is a tree this recursion is well founded (loop-free) and the
above equation uniquely determines the messages.  We will use an unnormalized
belief function defined as follows.
\begin{equation} \label{eqn:BPB}
 B_s(x_s) = \Psi_s(x_s) \prod_{t \in \nbrs_s} m_{t\rightarrow s}(x_s)
\end{equation}
When $G$ is a tree the belief function is proportional to the
marginal probability $P_s$ defined by (\ref{eqn:B}).
It is sometimes useful to define the ``pre-message''
$M_{t\rightarrow s}$ as follows for $x_t \in {\cal X}_t$.
\begin{equation} \label{def:m}
  M_{t\rightarrow s}(x_t) = \Psi_t(x_t) \prod_{u\in \nbrs_t\setminus s} m_{u\rightarrow t}(x_t)
\end{equation}
Note that the pre-message $M_{t\rightarrow s}$ defines a weighting on
the state space of the source node ${\cal X}_t$, while the message
$m_{t\rightarrow s}$ defines a weighting on the state space of the
destination, ${\cal X}_s$.  We can then
re-express~\eqref{eqn:m}--\eqref{eqn:BPB} as
\begin{align*}
 m_{t\rightarrow s}(x_s)  &=  \sum_{x_t\in {\cal X}_t} \; \Psi_{t,s}(x_t,x_s) M_{t\rightarrow s}(x_t)
 &
 B_t(x_t) &= M_{t\rightarrow s}(x_t) m_{s\rightarrow t}(x_t)
\end{align*}

Although we develop our results for tree--structured graphs, it is
common to apply belief propagation to graphs with cycles as well (``loopy''
belief propagation).  We note connections to and differences for loopy BP
in the text where appropriate.

For reasons of numerical stability, it is common to normalize each message
$m_{t\rightarrow s}$ so that it has unit sum.  However, such normalization
of messages has no other effect on the (normalized) belief
functions~\eqref{eqn:BPB}.  Thus for conceptual simplicity in developing and
analyzing particle belief propagation we avoid any explicit normalization of
the messages; such normalization can be included in the algorithms in practice.

Additionally, for reasons of computational efficiency it is common to use
the alternative expression
$m_{t\rightarrow s}(x_s) = \sum \; \Psi_{t,s}(x_t,x_s) B_t(x_t) / m_{s\rightarrow t}(x_t)$
when computing the messages.  By storing and updating the belief values $B_t(x_t)$
incrementally as incoming messages are re-computed, one can significantly
reduce the number of operations required.  Although our development of
particle belief propagation uses the update form~\eqref{eqn:m}, this
alternative formulation can be applied to improve its efficiency as well.


\subsection{Sum-Product versus Max-Product Algorithm}

There are two different algorithms that can be chosen to implement BP.
The sum-product algorithm computes the marginal distribution at each node of the MRF.
The max-product algorithm, on the other hand, aims at computing the MAP estimate of the whole MRF.
Therefore, max-product belief propagation attempt to solve the same problem as that of Graph Cuts algorithms.


\subsection{Loopy BP and Message Normalization}
Loopy belief propagation maintains a state which stores a numerical value for
each message value $m_{t\rightarrow s}(x_s)$.  In loopy BP one repeatedly
updates one message at a time.  More specifically, one selects a directed edge
$t \rightarrow s$ and updates all values $m_{t\rightarrow s}(x_s)$ using
equation \eqref{eqn:m}.  Loopy BP often involves a large number of message
updates.  As the number of updates increases message values typically diverge
--- usually tending toward zero but possibly toward infinity if the potentials
$\Psi_s$ and $\Psi_{t,s}$ are large.  This typically results in floating point
underflow or overflow.  To avoid floating point errors messages can be
normalized --- the function $m_{t\rightarrow s}$ can be scaled by a constant so
that it sums to one.  We can normalize an entire state by normalizing each
message.  It is important to note that normalization commutes with update.
More specifically, normalizing a message, updating, and then normalizing the
resulting state, is the same as updating (without normalizing) and then
normalizing the resulting state.  This implies that any normalized state of the
system can be viewed as the result of running unnormalized updates and then
normalizing the resulting state only at the end.  For conceptual simplicity we
avoid explicit normalization throughout this paper.  But all algorithms
described here can be normalized and have the property that message updates
commute with normalization in the sense just mentioned.

\subsection{Efficiency of Updates}
Performing a message update using equation \eqref{eqn:m} would seem to involve
an inner loop over the node $u$.  This inner loop can be avoided by using the
following which is equivalent to \eqref{eqn:m}.
\begin{align}
  m_{t\rightarrow s}(x_s) & = \sum_{x_t\in {\cal X}_t} \; \frac{\Psi_{t,s}(x_t,x_s) B_t(x_t)}{m_{s\rightarrow t}(x_t)}
\end{align}
The belief values $B_t$ can be stored as part of the state and maintained
incrementally when updates are made.  This efficiency improvement is possible
with the particle belief algorithm described in the next section.  However, for
conceptual simplicity we will consider only the inefficient update analogous
to \eqref{eqn:m}.

\section{Markov Chain Monte Carlo (MCMC) Methods}
\label{sec:MCMC}

Markov chain Monte Carlo (MCMC) is a class of algorithms for simulating direct draws
from complex probability distributions of interest. MCMC has its name from the fact that it uses the previous samples to randomly generate the next samples, 
constructing a Markov chain that has the desired distribution as its equilibrium distribution. The state of the chain after a large number of steps is then used as a sample from the desired distribution. The quality of the sample improves as a function of the number of steps.	

\subsection{Markov Process and Markov Chains}

Let $X_t$ denote the value
of a random variable at time $t$, and let the state space refer to the range of possible
X values. The random variable is a Markov process if the transition probabilities
between different values in the state space depend only on the random variable's
current state, i.e.,

\begin{eqnarray}
\label{eq_MP}
Pr(X_{t+1}|X_{t}, X_{t-1},...,X_{0}) = Pr(X_{t+1}|X_{t})
\end{eqnarray}

Thus for a Markov random variable the only information about the past needed
to predict the future is the current state of the random variable, knowledge of the
values of earlier states do not change the transition probability. A Markov chain
refers to a sequence of random variables $(X_0;...;X_n)$ generated by a Markov
process.

\subsection{Monte Carlo Integration}

The Monte Carlo approach was originally used as a method to compute integrals approximately by random number generation. 
Suppose we wish to compute a complex integral that can be represented as a product of a function $f(x)$ and a 
probability density function $p(x)$, the integral can then be expressed as an expectation of $f(x)$ over $p(x)$.
This expectation in turn can be approximated by drawing a large number of random variables from $p(x)$:

\begin{eqnarray}
\label{eq_MCI}
\int^{b}_{a}g(x)dx = \int^{b}_{a}f(x)p(x)dx = E_{p(x)}[f(x)] \approx \frac{1}{n}\sum^{n}_{i=1}f(x_i)
\end{eqnarray}

\subsection{Metropolis-Hasting Algorithm}

The most challenging problem of applying Monte Carlo integration is how to sample from some complex probability distribution $p(x)$.
Suppose our goal is to draw samples from some distribution $p(x)$ where
$p(x) = \frac{1}{Z}f(x)$, where the normalizing constant Z may not be known, and very
difficult to compute.

The Metropolis algorithm \cite{metropolis49} generates a sequence of draws from this distribution by first computing

\begin{eqnarray}
\label{eq_metropolis1}
\alpha = \min\left(\frac{f(\theta^*)Q(\theta_{t-1},\theta^*)}{f(\theta_{t-1})Q(\theta^*,\theta_{t-1})}, 1\right)
\end{eqnarray}

at each time step $t$, and then accepting a candidate point with probability $\alpha$. If the candidate is rejected, the current value $\theta_{t-1}$ is retained.
Here $Q(\theta_{t-1},\theta^*)$ is the transition probability function.
The original Metropolis algorithm calls for the transition probability function to be symmetric, i.e. $Q(\theta_{t-1},\theta^*) = Q(\theta^*,\theta_{t-1})$; 
Hastings in \cite{hastings70} generalized the Metropolis algorithm by using an arbitrary transition probability function.

\subsection{Simulated Annealing}

Simulated annealing was developed as an approach for locating a good approximation to the global optimum of a complex multimodal function in a large search space.
This is a good solution for cases where standard hill-climbing approaches such as gradient descent may get stuck at a local optimal peak. 

The idea is that when we
initially start sampling the space, we will accept a reasonable probability of a
down-hill move in order to explore the entire space. As the process proceeds, we
decrease the probability of such down-hill moves.

\begin{eqnarray}
\label{eq_simanl}
\alpha_{SA} = \min\left(\left(\frac{f(\theta^*)Q(\theta_{t-1},\theta^*)}{f(\theta_{t-1})Q(\theta^*,\theta_{t-1})}^{\frac{1}{T(t)}}\right), 1\right)
\end{eqnarray}

We can see that simulated annealing is very closely related to Metropolis sampling, differing only in that the probability $\alpha$ of a move also depends on a temperature $T(t)$, often called the cooling schedule. If $T$ is set to $1$ then equation (\ref{eq_simanl}) is exactly (\ref{eq_metropolis1}).
Simulated annealing usually starts off with a high temperature - high transition probability, and then cool down to a very low temperature 
($T \approx 0$) - low transition probability.

\subsection{Gibbs sampling}

Gibbs sampling is considered a special case of Metropolis-Hastings sampling wherein the
random value is always accepted (i.e, $\alpha = 1$).
The purpose of Gibbs sampling is to generate a sequence of samples from the joint probability distribution of multiple random variables. The purpose of such a sequence is to approximate the joint distribution, or to compute an integral (such as an expected value).

The key to the Gibbs sampler is that one only considers univariate conditional
distributions — the distribution when all of the random variables but one are
assigned fixed values. Such conditional distributions are far easier to simulate than
complex joint distributions and usually have simple forms.
Thus, one simulates $n$ random variables sequentially from the $n$ univariate conditionals rather than generating
a single n-dimensional vector in a single pass using the full joint distribution.

\section{Particle Filters}
\label{sec:PF}

Particle filtering, also known as sequential Monte Carlo methods (SMC), are sophisticated model estimation techniques based on simulation (\cite{arulampalam02, doucet01, godsill01, khan05}).

They are usually used to estimate Bayesian models and are the sequential ('on-line') analogue of Markov chain Monte Carlo (MCMC) batch methods and are often similar to importance sampling methods. Well-designed particle filters can often be much faster than MCMC. They are often an alternative to the Extended Kalman filter (EKF) or Unscented Kalman filter (UKF) (\cite{vandermerwe00}) with the advantage that, with sufficient samples, they approach the Bayesian optimal estimate, so they can be made more accurate than either the EKF or UKF. However, when the sample size is not sufficiently large, they might suffer from sample impoverishment. Particle filters can also be combined by using a version of the Kalman filter as a proposal distribution for the particle filter.

\begin{figure}[t]
\centering
\includegraphics[width = 0.9 \textwidth]{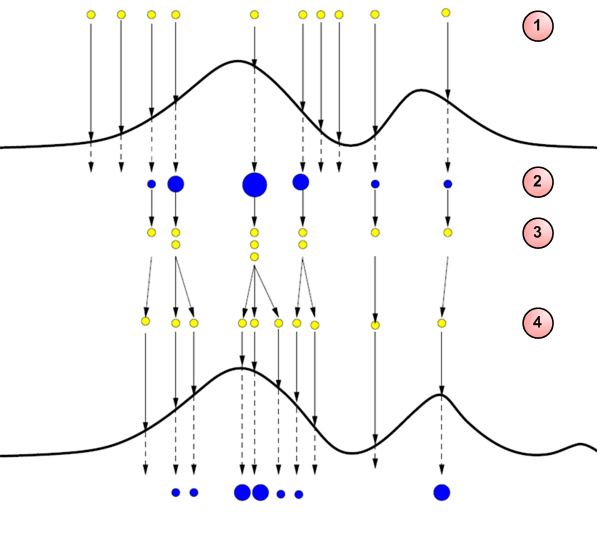}
\caption{Particle filtering approximates the probability distribution of interest using a large set of random samples, named particles.}
\label{fig:PF}
\end{figure}

Particle filtering starts with a large set of random samples called particles and their weights:

$$\left\{\left(x_t^i, w_t^i\right):i=1,..,P\right\}$$
$$\sum_{i=1}^{P}w_t^i = 1$$

Particle filtering then iteratively approximates the probability distribution of interest using the initial set of particles. 

$$\int f(x_t)p(x_t|y_0,..,y_t)dx_t \approx \sum_{i=1}^{P}w_t^if(x_t^i)$$

A single iteration of the algorithm (as illustrated in Figure~\ref{fig:PF})is as follows:

\begin{enumerate}

\item Draw samples from the proposal distribution.
\item Update the importance weights up to a normalizing constant.
\item Draw P new particles from the current particle set with
probabilities proportional to their weights.

\item Set the weight of each particle to be 1/P. Update the proposal
distribution using the new set of particles.
\end{enumerate}

\section{Particle-based Belief Propagation}
\label{sec:PBP}


We now consider the case where $|{\cal X}_s|$ is too large to
enumerate in practice and define a generic particle (sample) based
algorithm.  This algorithm essentially corresponds to a non-iterative
version of the method described in~\cite{koller99}.

The procedure samples a set of particles $x^{(1)}_s$, $\ldots$,
$x^{(n)}_s$ with $x^{(i)}_s \in {\cal X}_s$ at each node $s$ of the
network\footnote{ It is also possible to sample a set of particles
$\{x^{(i)}_{st}\}$ for each \emph{message} $m_{s\rightarrow t}$ in
the network from potentially different distributions $W_{st}(x_s)$,
for which our analysis remains essentially unchanged.  However, for
notational simplicity and to be able to apply the more
computationally efficient message expression described in
Section~\ref{sec:BP}, we use a single distribution and sample set for
each node.}, drawn from a sampling distribution (or weighting)
$W_s(x_s) > 0$ (corresponding to the proposal distribution in
particle filtering).
First we note that \eqref{eqn:m} can be written as the following
importance-sampling corrected expectation.
\begin{align}
\label{eqn:sampling}
m_{t\rightarrow s}(x_s) &= \expectsub{x_t \sim W_t}{\frac{\Psi_{s,t}(x_s,x_t)\Psi_t(x_t) \prod_{u\in \nbrs_t\setminus s} m_{u\rightarrow t}(x_t)}{W_t(x_t)}}
\end{align}
%
Given a sample $x_t^{(1)}$, $\ldots$, $x_t^{(n)}$ of points drawn from $W_t$  we can estimate $m_{t\rightarrow s}(x_s^{(i)})$ as
follows where $w_s^{(i)} = W_s(x_s^{(i)})$. 
\begin{align}
\label{eqn:PBP}
\widehat{m}^{(i)}_{t\rightarrow s} &= \frac{1}{n} \sum_{j=1}^n \frac{\Psi_{t,s}(x_t^{(j)},x_s^{(i)}) \Psi_t(x^{(j)}_t) \prod_{u\in \nbrs_t\setminus s} \widehat{m}^{(j)}_{u\rightarrow t}}{ w^{(j)}_t}
\end{align}
Equation \eqref{eqn:PBP} represents a finite sample estimate for
\eqref{eqn:sampling}. Alternatively, \eqref{eqn:PBP} defines a belief
propagation algorithm where messages are defined on particles rather
than the entire set ${\cal X}_s$.  As in classical belief
propagation, for tree structured graphs and fixed particle locations
there is a unique set of messages satisfying \eqref{eqn:PBP}.
Equation \eqref{eqn:PBP} can also be applied for loopy graphs (again
observing that message normalization can be conceptually ignored). In
this simple version, the sample values $x_s^{(i)}$ and weights
$w_s^{(i)}$ remain unchanged as messages are updated.

We now show that equation \eqref{eqn:PBP} is consistent---it agrees
with \eqref{eqn:m} in the limit as $n \rightarrow \infty$.
For any finite sample, define the particle domain $\widehat{\cal
X}_s$ and the count $c_i(x)$ for $x \in \widehat{\cal X}_s$ as
\begin{align*}
 \widehat{\cal X}_s &= \{x_s\in {\cal X}_s:\;\exists k\;x^{(i)}_s=x_s\}
 &
 c_s(x_s) &= |\{k:\;x^{(i)}_s=x_s\}|
\end{align*}
Equation \eqref{eqn:PBP} has the property that if $x^{(i)}_s =
x^{(i^\prime)}_s$ then $m^{(i)}_{t\rightarrow s} = m^{(i^\prime)}_{t\rightarrow
s}$; thus we can rewrite \eqref{eqn:PBP} as
\begin{align}
\label{eqn:consistent}
 \widehat{m}_{t\rightarrow s}(x_s) &= \frac{1}{n} \sum_{x_t\in \widehat{\cal X}_t} \frac{c_t(x_t)}{W_t(x_t)}
\Psi_{t,s}(x_t,x_s) \Psi_t(x_t) \prod_{u\in \nbrs_t\setminus s} \widehat{m}_{u\rightarrow t}(x_t)
 & x_s &\in \widehat{\cal X}_s
\end{align}
Since we have assumed $W_s(x_s) > 0$, in the limit of an infinite
sample $\widehat{\cal X}_t$ becomes all of ${\cal X}_t$ and the ratio
$(c_t(x_t) / n)$ converges to $W_t(x_t)$. So for sufficiently large
samples \eqref{eqn:consistent} approaches the true message
\eqref{eqn:m}.
Fundamentally, we are interested in particle-based approximations to
belief propagation for their finite--sample behavior, i.e., we hope
that a relatively small collection of samples will give rise to an
accurate estimate of the beliefs - the true marginal $P_t(x_t)$. 
At any stage of BP we can use our current marginal
estimate to construct a new sampling distribution for node $t$ and
draw a new set of particles $\{x^{(i)}_t\}$.  This leads to an
iterative algorithm which continues to improve its estimates as the
sampling distributions become more accurately targeted.  Unfortunately,
such iterative resampling processes often require more work to
analyze; see e.g.~\cite{neal03}.

In~\cite{koller99}, the sampling distributions were constructed using
a density estimation step (fitting mixtures of Gaussians).  However,
the fact that the belief estimate $\dhat{M}_t(x_t)$ can be computed
at any value of $x_t$ allows us to use another approach, which has
also been applied to particle filters~\cite{godsill01,khan05} with
success.  By running a short MCMC simulation such as the Metropolis-Hastings algorithm, one can attempt
to draw samples directly from the weighting $\dhat{M}_t$.  This
manages to avoid any distributional assumptions inherent in density
estimation methods, but has the disadvantage that it can be difficult
to assess convergence during MCMC sampling.

\section{Applications of PBP to vision problems}

In this section we describe the applications of the PBP algorithm to several Vision problems related to scene geometry recovery.

\subsection{Particle Belief Propagation for Structure from Motion}

\begin{figure}[t] \centering
\begin{tabular}{cc}
 \includegraphics[width=0.6\textwidth]{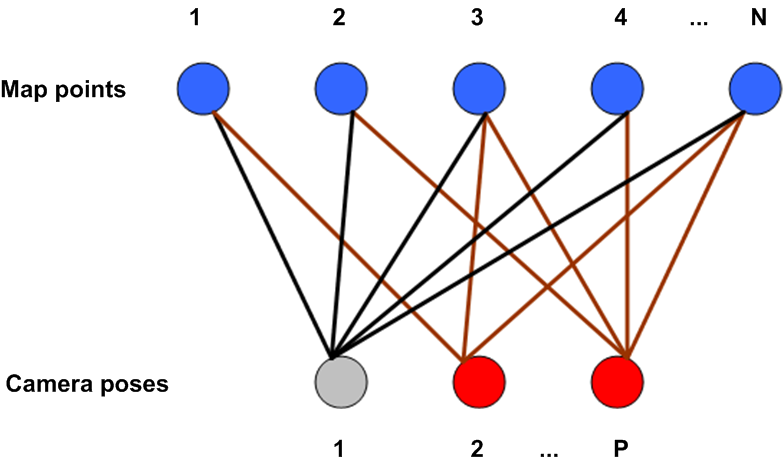}
\end{tabular}
\caption{The representation of the Structure from Motion problem as a bipartite graphical model. } \label{fig:gr}
\end{figure}

In structure from motion(SfM), given a set of 2D images of the same scene, the objective is to simultaneously recover both the camera trajectories and the 3D structure of the scene. For this problem, it is commonly accepted that Bundle Adjustment (BA) \cite{Trigg99} is the Gold standard. There has been related work on this problem \cite{Hartley04,kaucic01,Szeliski06}, mostly using geometric based methods. All of these works demonstrated that their result using BA as the last step is much better than without BA. With each camera pose and each map object 3D location being represented as a finite set of particles, we show that PBP can successfully estimate their true states by estimating their posterior distributions over the state space given the image observations. The algorithm was tested on both real and synthetic data. An experiment on synthetic data allows us to compare the performance of our method with BA.

First, we represent the problem of SfM in the context of PBP.  
Each observed image is converted to a sparse set of image points. These points can be high-level image features, obtained by a feature detector (such as corner detector or SIFT detector \cite{SIFT}). The correspondences of image points between images are automatically obtained using a feature matching method. This also defines the correspondences between the image points and the map points. The scene is represented as a sparse map of 3D points. Each camera pose is also a 3D point, combined with 3 angles of rotation, which define the rotation of the camera about 3 axes. Our method is based on the assumption that given a set of image observations and the estimate of all map points, there exists a probability distribution for each camera pose over its state space. The true state of map points and camera poses are hidden variables that we want to estimate. Let $P_i \in \Re^6$ denote the random variable for the $i^{th}$ camera pose. Let $Y_j \in \Re^3$ denote the random variable for the $j^{th}$ map point. The observed data are image points and their correspondences. We denote $x_{ij} \in \Re^2$ the image point variable associated with map point $Y_j$ as seen from camera $P_i$.  

Our graphical model $G(V,E)$ consists of two types of nodes. Each camera node $i$ is associated with a camera variable $P_i$, each map node $j$ is associated with a map variable $Y_j$. For simplicity of notation we name each node after its variable. For each camera and a map point it observes, we add an edge connecting two respective nodes into the graph. No edge connects any two camera nodes, or any two map nodes. If each map point is seen by all cameras, we have a complete bipartite graph.
For each edge in the graph, there is a binary potential function associated to it, denoted $\Psi_{i,j}(P_i,Y_j)$. More specifically, we have:

\begin{align} \label{eqn:binary}
 \Psi_{i,j}(P_i,Y_j) &= e^{\frac{-\left\|Q(P_i, Y_j)-x_{ij}\right\|^2}{2 * \sigma^2}}
\end{align}

where $Q(P_i, Y_j)$ is the reprojection function that takes a camera pose and a map point and returns the reprojected image point, $x_{ij}$ is the image observation of map point $y_j$ as seen from camera $p_i$, $\sigma^2$ is the variance of the Gaussian image noise. This is equivalent to assuming a normal distribution of the true observation around the given observation. This allows us to represent the uncertainty of the observations.

At this point the message function and belief function on $G$ are well defined using ~\eqref{eqn:m} and ~\eqref{eqn:BPB}. However, the state space of each hidden variable in $G$ is too large to do inference with BP. In order to use PBP, we discretize the state space into a relatively small number of states, each of which is represented by a particle. These states can be sampled from a normal distribution with some initial mean and variance. Now we have at each camera node $P_i$, a set of $M$ particles: $P^{(1)}_i$, $\ldots$, $P^{(M)}_i$, at each map node $Y_j$, a set of $N$ particles: $y^{(1)}_j$, $\ldots$, $y^{(N)}_j$. If we assume no unary potential at each node, we can define the message going from camera node $i$ to map node $j$ in $G$ by ~\eqref{eqn:PBP} as follows:

\begin{align}
\label{eqn:SFM}
\widehat{m}^{(k)}_{i\rightarrow j} &= \frac{1}{M} \sum_{h=1}^M w^{(h)}_t \Psi_{i,j}(P_i^{(h)},Y_j^{(k)}) \prod_{u\in \nbrs_i\setminus j} \widehat{m}^{(h)}_{u\rightarrow i}
\end{align}

At the beginning, all particles in a node are assigned uniform weights and no message has been computed, equation ~\eqref{eqn:SFM} becomes:

\begin{align}
\label{eqn:SFM2}
\widehat{m}^{(k)}_{i\rightarrow j} &= \frac{1}{M} \sum_{h=1}^M \Psi_{i,j}(P_i^{(h)},Y_j^{(k)})
\end{align}

This can be interpreted as the marginal probability $\sum^{M}_{h=1} P(Y_j = Y_j^{(k)}|x_{ij},P_i=P_i^{(h)})$. It follows that the belief of a map node becomes the marginal probability over all assignments of its neighboring camera nodes, conditioned on relevant observations:

\begin{align}
\label{eqn:SFM3}
\widehat{B}_{j}(Y_j^{(k)}) &= \prod_{i\in \nbrs_j} \sum^{M}_{h=1} P(Y_j = Y_j^{(k)}|x,P_i=P_i^{(h)})
&= \sum_{\vec{p}} P(Y_j = Y_j^{(k)}|x,\vec{P})
\end{align}

where $\vec{P}$ is an assignment of values to all neighboring pose nodes.
The message from a map node to a pose node and the belief of a pose node can be interpreted similarly. As BP proceeds, the information from one node is sent to all other nodes in the graph. This allows nodes of the same type to communicate with each other. Then~\eqref{eqn:SFM3} will become exactly ~\eqref{eqn:B}, which is what we want to compute.
As shown in \cite{ihler09}, the posterior marginal estimated by PBP is in most cases guaranteed to converge to the true posterior marginal distribution estimated by BP. In addition to PBP, the samples at each node are iteratively updated by Gibbs sampling from the estimated posterior marginal distribution at that node. Gibbs sampling allows particles to freely explore the state space and thus compensates the inadequacy of representing a large state space by a small number of samples.   
After a sufficient number of alternative PBP and Gibbs sampling iterations, the estimated posterior at each node converges to the true posterior distribution, which gives us the answer for the final state of each camera pose and map point.  

The objective of this experiment is to compare PBP with the well known method BA in terms of reconstruction accuracy, by measuring the deviation of their reconstruction estimates from the ground truth. The model for bundle adjustment is a pairwise graphical model over poses $p_i$ and object
positions $y_j$ given observed locations $x_{ij}$ in the images:

\begin{align*} \label{eqn:binary}
 P(\{p_i\},\{y_j\}) &= \prod \psi_{i,j}(p_i,y_j) &
 \Psi_{i,j}(p_i,y_j) &= e^{\frac{-\left\|Q(p_i, y_j)-x_{ij}\right\|^2}{2 * \sigma^2}}
\end{align*}

Bundle adjustment then searches for a local
maximum (i.e., a mode) of the posterior distribution over the $\{p_i,y_j\}$.
However, if the intent is to minimize the expected squared error loss,
we should prefer to estimate the mean of the posterior distribution rather
than its mode.  These can be very different in problems where the posterior
distribution is very skewed or multi-modal; in such cases,
it may be advantageous to estimate the posterior
distribution and its mean using methods such as belief propagation.
Additionally, an explicit representation of uncertainty can be used to assess
confidence in the estimate.

\begin{figure}[t] \centering
\begin{tabular}{ccc}
 \raisebox{.5in}{
 \begin{tabular}{|l|c|}
   \hline
   Method & Error \\
   \hline\hline
   BA (Map) & $185.81\pm66.43$ \\
   PBP (Map) & $194.00\pm63.02$ \\
   BA (Pose) &  $11.88\pm4.40$ \\
   PBP (Pose) & $11.56\pm4.39$ \\
   \hline
 \end{tabular}} &
 \includegraphics[width=0.25\textwidth]{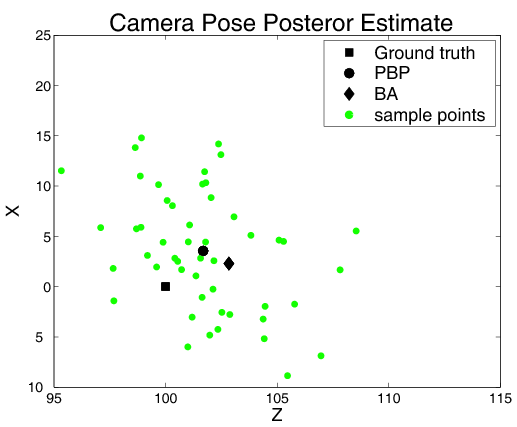} &
 \includegraphics[width=0.25\textwidth]{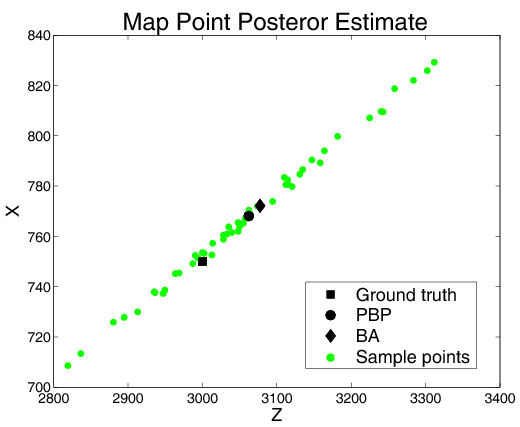}  \\
 (a) & (b) & (c)
\end{tabular}
\caption{Comparing bundle adjustment to PBP in structure from motion.
(a) Estimated mean and standard deviation of reconstruction errors for
camera pose and map positions; (b) example posterior for one camera pose and
(c) for one map point using PBP.} \label{fig:sfm}
\end{figure}

We show a comparison between estimates given by BA\footnote{We use the
\emph{sba} package in \cite{lourakis04}.} and PBP on synthetic
structure from motion data in Figure~\ref{fig:sfm}.  For the data,
we generate a fixed set of map points and a few camera poses. Each point's
observations are obtained by projecting the point on each camera's image
planes and adding Gaussian noise.  We assume that the camera calibration
parameters and the image correspondences are known, and initialize the
estimates for both algorithms to ground truth, then run each to
convergence and compare the resulting error.

Figure~\ref{fig:sfm}(a) shows the estimated mean and standard deviation of
reconstruction errors of BA and PBP from 200 different runs (units
are synthetic coordinates).  This shows that, at the same level of
image noise and from the same initial conditions, PBP produces essentially
the same accuracy as BA for both camera and map points.  However, we
expect that in less idealized cases (including, for example, incorrect
feature associations or outliers measurements), PBP may perform much
better. In addition, PBP provides an estimate of
state uncertainty (although at some computational cost).
Figure~\ref{fig:sfm}(b)--(c) shows the estimated posterior
distributions given by PBP for a camera pose and map point, respectively,
along with the mean found via PBP (circle), mode found via BA (diamond), and
true position (square).

\subsection{PBP for Dense Stereo Vision with Uncalibrated Cameras}

\begin{figure}[t] \centering
\begin{tabular}{cc}
 \includegraphics[width=0.5\textwidth]{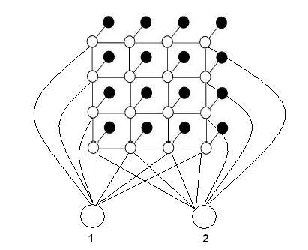}
\end{tabular}
\caption{The representation of the Dense Stereo Vision problem as a graphical model. } \label{fig:densegr}
\end{figure}

Classical dense two-frame Stereo algorithms compute a dense disparity map for each pixel given a pair of images under known camera constraints (i.e, the orientation of the 2 cameras and the distance between them are known) \cite{Jian03,Pedro06,scharstein01}. In this experiment, given a pair of stereo images with unknown camera constraints, we use PBP to simultaneously compute the dense depth map for each pixel, and the configuration of the second camera relative to the first. 

\begin{figure}[t] \centering
\begin{tabular}{cc}
 \includegraphics[width=0.45\textwidth]{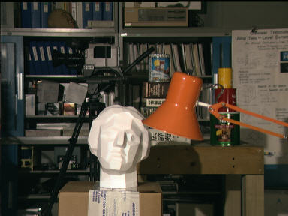} &
 \includegraphics[width=0.45\textwidth]{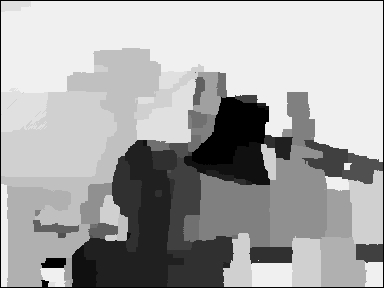}  \\
 (a) & (b)
\end{tabular}
\caption{(a) Ground truth; (b) estimated depth map after a few iterations of PBP.} \label{fig:stro}
\end{figure}

%
\begin{figure}[t] \centering
\begin{tabular}{ccc}
 \includegraphics[width=0.33\textwidth]{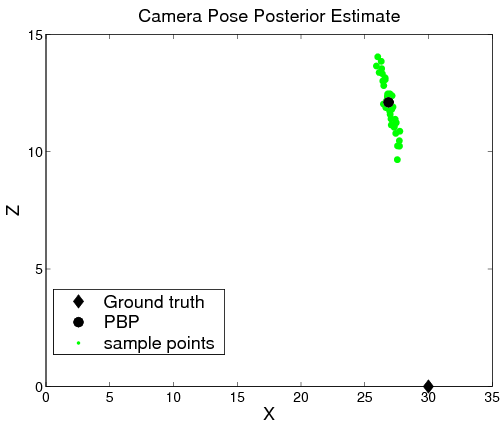} &
 \includegraphics[width=0.33\textwidth]{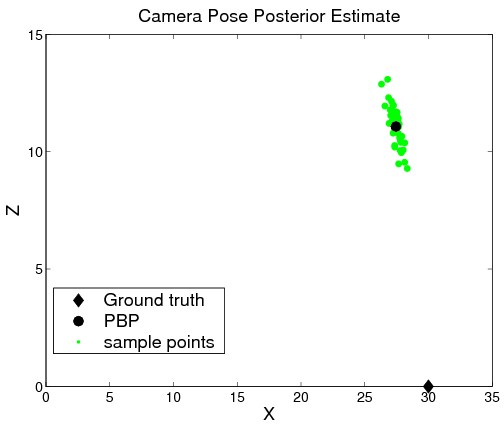} &
 \includegraphics[width=0.33\textwidth]{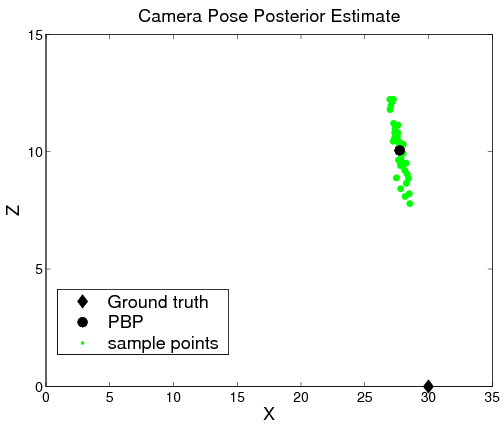} \\
 \includegraphics[width=0.33\textwidth]{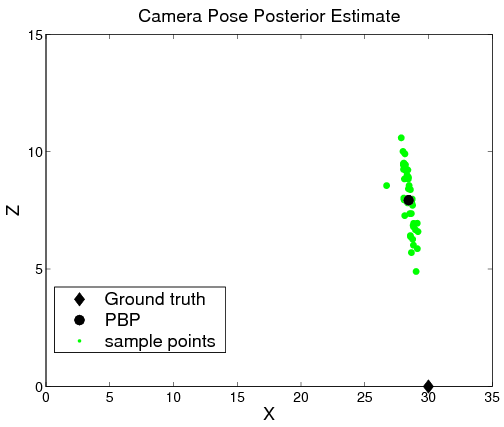} &
 \includegraphics[width=0.33\textwidth]{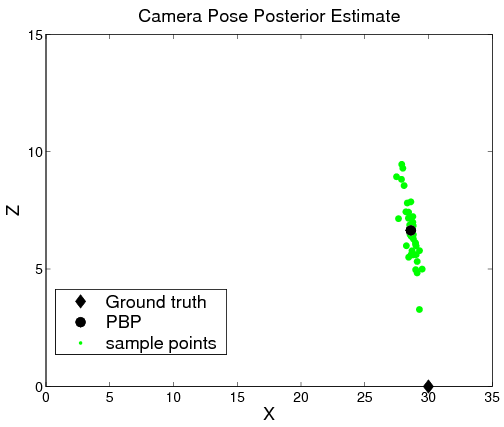}
\end{tabular}
\caption{Estimated posterior distribution of the camera pose over time} 
\label{fig:plt}
\end{figure}

The formulation of the graphical model in this case is quite similar to the previous problem. However there are only 2 camera nodes, and the number of map nodes equals the number of pixels in the first image. An edge is added between every pixel and every camera. (see Figure~\ref{fig:densegr})
Figure~\ref{fig:stro} shows the estimated depth map after a few iterations. We do not show evaluation results of our output depth map in comparison with ground truth and other stereo algorithm (using for example the Middlebury stereo benchmark \cite{scharstein01}), because such a comparison will not be very meaningful, as the labels we use are the true 3D depths of each pixel instead of their disparities. However, in Figure~\ref{fig:plt}, we plot the estimated posterior distribution of the second camera pose at each iteration, thus show that the distribution gradually approaches the true state over time.

\section{PBP for Slanted-Plane Stereo Estimation}

For this task, given a pair of images $(X, Y)$, and a given segmentation of $X$, and a given setting of the model parameters,
the inference problem is to find an assignment $Z$ of plane parameters to superpixels so as to minimize the total energy in (\ref{StereoEq2}). 
The energy defines a Markov random field.  More specifically, the match energy defines the unary potential on each superpixel
independently and the smoothness energy defines the binary potential on pairs of adjacent superpixels.

The complete inference algorithm is described in Section~\ref{sec:planeinference}.
\chapter{Slanted Plane Model with Surface Orientation Cues for Stereo Depth Estimation}
\label{str-depth-inf}

In this chapter, we present our slanted-plane stereo vision model with monocular cues for surface orientation estimation.
We also explain our idea of using HOG features as surface orientation cues, and present an analytical justification for this idea.

\section{The Slanted Plane Stereo Model}
\label{sec:slantedplane}

As previously mentioned in our introduction, the two-view stereo vision problem is commonly treated as a pixel-labeling problem, where labels are disparity values, usually quantized to be integers. The disparity is equivalent to the horizontal displacement of the pixel between the left and the right image, or the inverse depth of that pixel. Specifically, the corresponding pixel in the right image of a pixel $p = (x_p, y_p)$ in the left image is given by $p' = p - d(p) = (x_p - d(p), y_p)$. The labeling is then resolved by using an energy minimization technique. The standard form of the global energy function is formulated in equation (\ref{StereoEq1}). 

\begin{eqnarray}
\label{StereoEq1}
E(d) = E_M(d) + \lambda E_S(d)
\end{eqnarray}

where $E_M$ is the match energy measuring how well the disparity assignment $d$ agrees with the input image pair, $E_S$ is the smoothness energy enforcing the smoothness assumption. Stereo methods aiming at finding $d$ that optimizes the global energy function are called global methods. The best-performing algorithms in the Middlebury benchmark \cite{scharstein01} are all global methods.

\begin{figure*}[t] \centering
\begin{center}
 \includegraphics[width=0.95\textwidth]{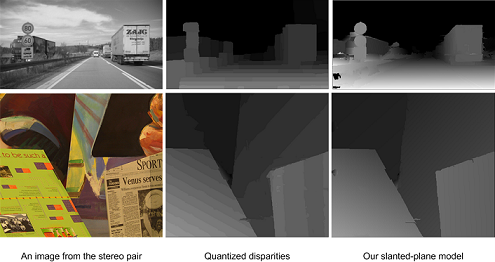}
\end{center}
\caption{The scene geometry is much more accurately captured by the slanted-plane model than by the pixel-based stereo model.}
\label{depthmodels}
\end{figure*}

The quantization of disparity values $d(p)$ helps reduce considerably the state space for searching. All techniques discussed in \cite{Sze02} follow a common framework - producing the quantized disparity map as output.
However, this output has the effect that the scene seems to be split into a series of fronto-parallel surfaces, as illustrated in Figure~\ref{depthmodels}. In other words, piecewise continuity of the scene is replaced by piecewise constancy. This is a poor explanation of the scene geometry, especially when the scene contains slanted surfaces.

The use of slanted-plane model for stereo vision was first proposed by BirchField and Tomasi in \cite{BirchfieldTomasi} as a special case of the affine model. Since then the model has been widely applied in many other stereo algorithms, including most of state-of-the-art algorithms in the current Middlebury benchmark. \cite{KSK06, WangZheng08, yang08}. The model assumes that the scene structure is locally planar - the scene is composed of a set of planar regions. This is a reasonable assumption, based on the fact that most surfaces found in a natural environment can be approximated by a plane or a set of planes. 
These algorithms start by segmenting the image into homogeneous regions called superpixels, using an image segmentation method. The superpixels found by the segmentation method are coherent groupings of pixels having similar properties (e.g: color, texture). Therefore it is just natural to assume that each superpixel corresponds to a disparity plane. Next, the goal of the stereo algorithm is to infer the location and orientation (or surface normal) of each of these planes - which are fully defined by the 3 plane parameters. After that, the disparity of each pixel can be directly computed from the disparity map equation, with sub-pixel precision.

Our stereo vision model is a slanted plane model involving monocular shape from texture cues. The introduction of an additional monocular term into the slanted-plane stereo model is a novel idea that has not been investigated. Another difference of our work is in the inference algorithm.
Other related work that also employ the plane-based stereo model use different approach in the inference step. One of the best methods in the Middlebury benchmark is \cite{KSK06}. The authors used a local matching step to extract reliable correspondences, followed by robust plane fitting to fit a disparity plane into each superpixel separately. The process were repeated for iterative plane refinement. For plane fitting, instead of trying to estimate all plane parameters jointly, the authors proposed a decomposition method to solve each parameter, again separately. The estimated planes are fixed after this step. Other slanted-plane stereo algorithms as in \cite{WangZheng08, yang08} also follow a very similar approach for plane estimation.

Our stereo inference algorithm also infers a disparity plane for each superpixel. However we use the Particle BP inference algorithm to simultaneously find the optimal joint assignment of all plane parameters to all superpixels. The optimal assignment is found by minimizing a high-dimensional global energy function. 
With this PBP inference algorithm, each particle, which represents a proposal disparity plane for a specific superpixel, gets updated over time, instead of being fixed after the plane fitting step, as in previously mentioned methods.

The energy function involves three terms --- a correspondence energy measuring the degree to which the left and right images agree under the induced disparity map, a smoothness energy measuring the smoothness of the induced depth map, and a texture energy measuring the degree to which the surface orientation at each point agrees with a certain (monocular) texture based surface orientation cue.
Specifically, the global energy in equation (\ref{StereoEq1}) is replaced by the following:
 
\begin{eqnarray}
\label{StereoEq2}
E(Z) = E_M(Z) + E_S(Z) + E_T(Z)
\end{eqnarray}

where $Z$ is the assignment of a disparity plane to each superpixel in the left image.
We denote $X$ to be the left image, $Y$ to be the right image. For each superpixel $i$ of $X$
we have that $Z$ specifies three plane parameters $A_i$ $B_i$, and $C_i$.  Given an assignment $Z$ of plane parameters to superpixels
we define the disparity $d(p)$ for any pixel $p$ as follows (where $i(p)$ is the superpixel containing $p$ and $x_p$ and $y_p$ are the image
coordinates of $p$).

\begin{eqnarray}
\label{dispEq}
d(p) = A_{i(p)} x_p + B_{i(p)} y_p + C_{i(p)}
\end{eqnarray}

So by equation (\ref{dispEq}) we have that $Z$ assigns a real valued disparity to each pixel.

\begin{figure*}[t] \centering
\begin{center}
 \includegraphics[width=0.8\textwidth]{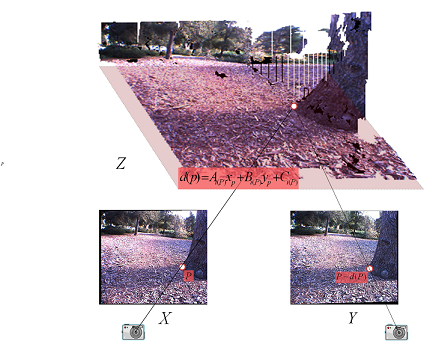}
\end{center}
\caption{The assigned disparity plane defines a disparity value for any pixel in the superpixel.}
\label{reproject2}
\end{figure*}

To define the smoothness energy we write $(p,q) \in B_{i,j}$ if $p$ is a pixel in superpixel $i$, $q$ is a pixel in superpixel $j$, and $p$ and $q$ are adjacent pixels
($p$ is directly above, below, left or right of $q$).
The smoothness energy is defined as follows (where $\tau_S$ and $\lambda_S$ are parameters of the energy, $i$ ranges over all superpixels, $N(i)$ is the set of superpixels adjacent to $i$, and $z$ is the overall assignment of three dimensional vectors of plane parameters $(A_i,B_i,C_i)$ for all superpixels $i$):

\begin{equation}
\label{smooth}
E_S(Z) = \sum_{i,j\in N(i)} \min\parens{\tau_S, \sum_{(p,q)\in B_{i,j}}\;\;\lambda_S|d(p)-d(q)|}
\end{equation}

Intuitively the minimization with $\tau_S$ corresponds to interpreting the entire boundary between $i$ and $j$ as either an occlusion boundary or as a joining of two planes on the same object.

Next we consider the match energy. We write $p-d(p)$ for the pixel in $Y$ that corresponds to the pixel $p$ in image $X$ under the disparity $d(p)$.
For color images we construct a nine dimensional feature vector $\Phi^X(p)$ and $\Phi^Y(p)$ for the pixel $p$ in the images $X$ and $Y$
respectively.  The vector $\Phi^X(p)$ consists of three (bias gain corrected) color values plus a six dimensional color gradient vector
and similarly for $\Phi^Y_p$.  We write $\Phi_k^X(p)$ for the $k$th component of the vector $\Phi^X(p)$.  The match energy is defined as
follows where $\lambda_k$ are nine scalar parameters of the match energy.
\begin{equation}
\label{match}
E_M(Z) = \sum_{p \in X} \sum_k \lambda_k(\Phi^x_k(p) - \Phi^y_k(p+d(p))\;)^2
\end{equation}

Finally we consider the texture energy.
At each pixel $p$ we also compute a HOG vector $H(p)$
which is a 24 dimensional feature vector consisting of three 8 dimensional normalized edge orientation histograms --- an 8 dimensional orientation histogram
is computed at three different scales.
The texture energy is defined as follows where $i(p)$ is the superpixel containing pixel $p$ and where
the scalars $\tau_T$, $\lambda_A$, $\lambda_B$, and the vectors $\beta_A$ and $\beta_B$ are parameters
of the energy.  The form of this energy is justified in Chapter~\ref{chap:hog}.
\begin{equation}
\label{orientation}
E_T = \sum_p \min\parens{\tau_T, \begin{array}{ll} & \lambda_A\;\parens{d(p)(\beta_A \cdot H(p)\;) - A_{i(p)}}^2 \\ + & \lambda_B\;\parens{d(p)(\beta_B \cdot H(p)\;) - B_{i(p)}}^2 \end{array}}
\end{equation}

These energy terms can also be interpreted in a probabilistic way as follows.
The combination of the smoothness energy term and the texture energy term can be interpreted as an energy $E_Z(X,Z,\beta_Z)$, which determines the probability $P(Z|X,\beta_Z)$. 
The match energy term can be interpreted as an energy $E_Y(X,Y,Z,\beta_Y)$, which determines $P(Y|X,Z,\beta_Y)$.
The general conditional probability formulation of our model will be discussed in details in Chapter~\ref{uspv-depth-lrn}.

\section{HOG as Surface Orientation Cues}
\label{chap:hog}

\begin{figure*}[h] \centering
\begin{center}
\begin{tabular}{cc}
 \includegraphics[width=0.5\textwidth]{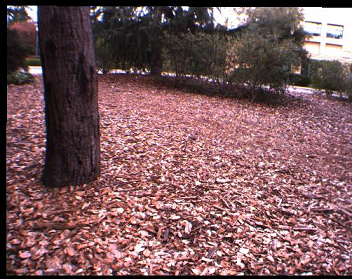} &
 \includegraphics[width=0.5\textwidth]{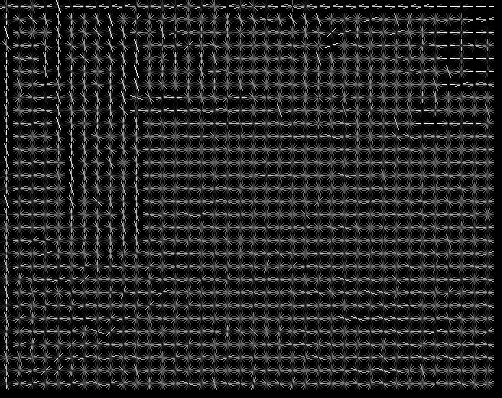}
\end{tabular}
\end{center}
\caption{HOG features for an example image.}
\label{imageHOG}
\end{figure*}

\section{Relationship between HOG and Surface Orientation}
\label{sec:Hog-SurfaceOrient}


In this chapter we discuss the possibility of using a specific type of monocular cues - the HOG feature (Histogram of Oriented Gradients). This feature has recently been extensively applied in computer vision and image processing for the purpose of object detection (\cite{dalal05, Felz08}). The technique counts occurrences of gradient orientation in localized portions of an image. This method is similar to that of edge orientation histograms, scale-invariant feature transform descriptors (or SIFT \cite{SIFT}), and shape contexts \cite{belongie02}, but differs in that it computes on a dense grid of uniformly spaced cells and uses overlapping local contrast normalization for improved performance.
  
First we observe that HOG, which describe the edge distribution, can be related to the orientation of surfaces.
We formulate the statistical relationship between HOG features and surface normal. Based on this derivation, we then argue that HOG features can be used as surface orientation cues, and thus can be incorporated in our existing slanted-plane stereo model (as described in Section~\ref{sec:slantedplane}).

\begin{figure*}[t] \centering
\begin{center}
 \includegraphics[width=0.8\textwidth]{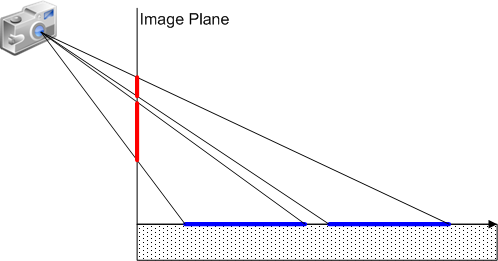}
\end{center}
\caption{As a surface is tilted away
from the camera the edges in the direction of the tilt become foreshortened while the edges
orthogonal to the tilt are not.}
\label{HOGintuition}
\end{figure*}

\begin{figure*}[t] \centering
\begin{center}
\begin{tabular}{cc}
 \includegraphics[width=0.5\textwidth]{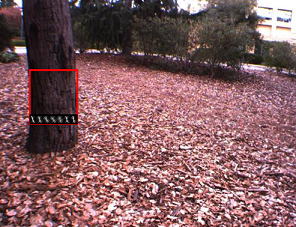} &
 \includegraphics[width=0.5\textwidth]{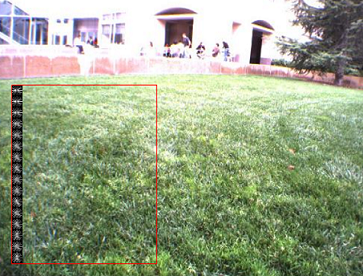} \\
 \includegraphics[width=0.5\textwidth]{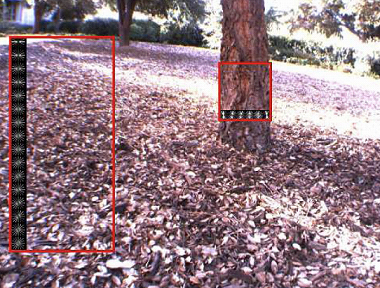} &
 \includegraphics[width=0.5\textwidth]{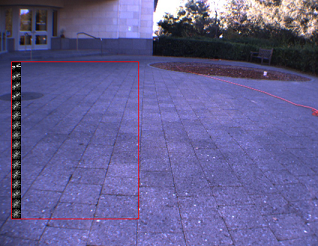}
\end{tabular}
\end{center}
\caption{HOG features for image regions.  The amount of edge as a function of angle, a HOG feature, is averaged over different vertical and horizontal regions on various images.
The different surface orientations of these regions affect the HOG features.  We can see the cylindrical structure of tree trunks and the fact that the ground plane becomes more tilted
as the distance increases.}
\label{image}
\end{figure*}

Here we aim at describing a justification for the idea of using HOG features as surface orientation cues, as mentioned in Chapter~\ref{str-depth-inf}. 
The basic intuition behind HOG as an orientation cue is that as a surface is tilted away from the camera the edges in the direction of the tilt become foreshortened while the
edges orthogonal to the tilt are not.  This changes the edge orientation distribution and therefore the edge orientation distribution can be used as a cue for surface orientation.
This effect is shown in figure ~\ref{image} where the average HOG feature is shown for various regions of tree trunk, forest floor, grass lawn, and patio tile.
The cylindrical shape of the tree trunk is clearly indicated
by the warping of the HOG feature as a function of position on the trunk.

We consider the case of a surface with an isotropic edge distribution --- the amount of edge at any two edge orientations is equal.  We do not expect this assumption to hold in
any particular small surface patch, but we do expect a statistical relationship between HOG features and orientation.  The assumption of isotropic surface textures provides
a departure point for analysis.  We are interested in the expected HOG
feature for the edges projected onto the image plane when the surface is tilted relative to the image plane.

We consider edges to be line segments on the image plane.  Each edge has a length $r$ and an orientation $\Theta$.
The HOG feature is a histogram of edge orientations.  We will normalize the HOG histogram so that it is a probability distribution on orientations.
If $P$ is a probability distribution (or probability density) on edges then the HOG feature $H(\Theta)$ is a probability distribution (or density) on $\Theta$.
More specifically, $H$ is a marginal of $P$ onto $\Theta$ in which edges are weighed by length.
\begin{eqnarray}
\label{HOG}
H(\Theta) & = & \frac{1}{\expect{r}}\;\;P(\Theta)\expect{r|\Theta}
\end{eqnarray}
We consider a surface patch imaged by a perspective camera.
A perspective camera induces the following map from three dimensional coordinates to image plane coordinates.
\begin{equation}
\label{threed}
x' = (fx)/z \;\;\;y'  =  (fy)/z
\end{equation}
We assume a coordinate system on the surface patch such that each point on the surface patch has coordinates $x_s$, $y_s$.
The image plane and surface coordinates can be selected so that we have the following
map from surface coordinates to three dimensional coordinates where $\Psi$ is the angle between surface normal and the image plane normal.
\begin{equation}
\label{intrinsic}
x  =  x_s \;\;\;\; y  =  y_s\cos\Psi \;\;\;\; z = z_0 + y_s \sin \Psi
\end{equation}
Combining (\ref{threed}) and (\ref{intrinsic}), and differentiating with respect to $x_s$ and $y_s$ gives the following with $z = z_0 + y_s \sin \Psi$.
\begin{equation}
\label{linear}
\begin{array}{cccl}
dx' & = & \parens{\frac{f}{z}}dx_s & - \parens{\frac{fx\sin \Psi}{z^2}}\;dy_s \\
dy' & = & \parens{\frac{f\cos \Psi}{z}} dy_s & - \parens{\frac{fy \cos \Psi \sin \Psi}{z^2}} dy_s
\end{array}
\end{equation}
We first consider the region near the center of the image.  At the center of the image we have $x = y = 0$ and (\ref{linear}) becomes
the following.
\begin{equation}
\label{affine}
dx'  = \parens{\frac{f}{z}}dx_s\;\;\;\; dy'  =  \parens{\frac{f\cos \Psi}{z}} dy_s
\end{equation}
We now have the following map from the length $R$ and orientation $\Gamma$ on the surface to length $r$ and orientation $\Theta$ on the image plane.
\begin{eqnarray}
\Theta & = & \tan^{-1}\parens{\sin \Gamma \cos \Psi,\;\cos \Gamma} \nn
\nn
r & = & \frac{fR}{z}\sqrt{\sin^2\Gamma\cos^2\Psi + \cos^2\Gamma} \nonumber
\end{eqnarray}
For $cos \Psi \not = 0$ we have a bijection between $\Gamma$ and $\Theta$.  Furthermore, for fixed $\Theta$ and $\Gamma$ we have a linear relationship between $r$ and $R$.
This gives the following.
\begin{equation}
H(\Theta) \propto \;\;Q(\Gamma(\Theta)\;)\frac{d \Gamma}{d \Theta} \expect{R|\Gamma(\Theta)} \frac{d r}{d R} \nonumber
\end{equation}
We assume that $Q$ is isotropic, i.e., that $Q(\Gamma)\expect{R|\Gamma}$ is a constant independent of $\Gamma$.
We then have the following.
\begin{eqnarray}
H(\Theta) & \propto & \frac{d \Gamma}{d \Theta}\;\sqrt{\cos^2 \Psi \sin^2\Gamma  + \cos^2 \Gamma} \nn
\nn
\frac{d\Gamma}{d\Theta} & = & \frac{1 + \tan^2\Theta}{\cos \Psi\parens{1 + \frac{\tan^2\Theta}{\cos^2\Psi}}} \nonumber
\end{eqnarray}
We have no simple formula for $H(\Theta)$.  However, we can see that $H(\Theta)$ is a continuous function of $\Theta$ satisfying the following.
\begin{equation}
\begin{array}{ccccc}
H(0) & = & H(\pi) & = & \frac{1}{Z}\parens{\frac{1}{\cos\Psi}} \\
\\
H(\pi/2) & = & H(3\pi/2) & = & \frac{1}{Z} \; \cos^2 \Psi
\end{array} \nonumber
\end{equation}
So we get the following where $H_{\min}$ is the minimum of $H(\Theta)$ and $H_{\max}$ is the maximum of $H(\Theta)$.
\begin{equation}
\label{shape}
H_{\min}/H_{\max} = \cos^3 \Psi
\end{equation}

To justify the form of the orientation energy (\ref{orientation}) we first note that in the same coordinate system as (\ref{intrinsic})
the equation for the surface plane can be written as follows.
\begin{equation}
\label{plane}
z = z_0 + y\tan\Psi
\end{equation}
If we let $b$ be the distance between the foci of the two cameras we have that the disparity $d$ equals $bf/z$.
Multiplying (\ref{plane}) by $bf/(zz_0)$ gives the following.
\begin{eqnarray}
\frac{bf}{z_0} & = & \frac{bf}{z} +\parens{\frac{b \tan \Psi}{z_0}}\parens{\frac{fy}{z}} \\
\nn
d_0 & = & d + \parens{\frac{d_0\tan \Psi}{f}}y' \\
\nn
d & = & d_0 - \parens{\frac{d_0\tan \Psi}{f}}y' \\
\nn
B & = & - \frac{d_0\tan \Psi}{f}
\end{eqnarray}
For the pixel $p$ at the center of the image we have $d(p) = d_0$.  We handle pixels outside of the center of the image by considering panning the camera to bring the desired
point to the center and approximating panning the camera by translating the image.  This gives the following general relation between the disparity plane parameter
and the angle $\Psi$ between the ray form the camera and the surface normal.
\begin{equation}
\label{justification}
B  = - \frac{d(p)\tan \Psi}{f}
\end{equation}
In the orientation energy (\ref{orientation}) we interpret $\beta_B \cdot H(p)$ as a predictor of $-(\tan \Psi)/f$ and we multiply by $d(p)$ to get a predictor of $B$.
We do not currently exploit (\ref{shape}). Doing so should result in a more refined texture cue.

\section{HOG Computation and Representation}
\label{sec:Hog-Compute}

\begin{figure*}[t] \centering
\begin{center}
 \includegraphics[width=0.8\textwidth]{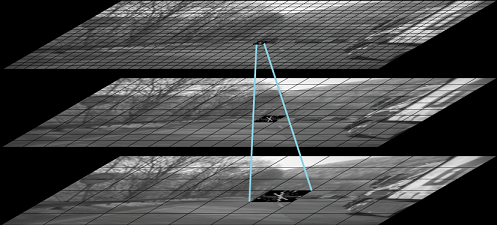}
\end{center}
\caption{We compute HOG features at each image scale for three different scales to obtain a 24-dimensional feature vector at each pixel.}
\label{P-HOG}
\end{figure*}

In order to use HOG features as a probability distribution on orientations, we want to compute HOG as a normalized histogram of edge orientations through the whole image.
Our HOG descriptor at each pixel is a 24-dimensional vector which describes the edge orientation distribution over the local image region around that pixel. (We compute a 8-dimensional HOG feature at each image scale, for 3 different scales to get a 24-dimensional feature vector at each pixel).
The HOG computation includes the following 3 main steps:

\begin{itemize}
	\item \textbf{Gradient Computation: }
	
The first step of calculation is the computation of the gradient values for all pixels in the left image. This is to simply apply the 1-D centered, point discrete derivative mask in one of or both the horizontal and vertical directions: $\left[-1,0,1\right]$ and $\left[-1,0,1\right]^{T}$.

  \item \textbf{Orientation Binning}
The second step of calculation involves creating the cell histograms. The cells themselves can either be square or radial in shape, and the histogram channels are evenly spread over 0 to 180 degrees or 0 to 360 degrees, depending on whether the gradient is "unsigned" or "signed". Here we use square cells, unsigned gradients in conjunction with 8 histogram channels. Each pixel within the cell casts a weighted vote for an orientation-based histogram channel based on the values found in the gradient computation. As for the vote weight, pixel contribution can either be the gradient magnitude itself, or some function of the magnitude; here we use the gradient magnitude.
Specifically, we quantize the gradient orientation of pixels in a cell into 8 bins (corresponding to 8 equal ranges from 0 to 180 degrees). In each bin we sum the magnitude of gradients in that bin. Then we normalize the magnitude of the resulting 8-dimensional vector of the bin.

  \item \textbf{Pyramidal HOG Descriptors}
We generate the HOG feature pyramid by computing HOG features of 3 different scales of the image pyramid (see Figure), corresponding to different sizes of the cell. The sizes of the cell at each scale are 8x8, 16x16 and 32x32 respectively.
In this step we used the integral image trick for the purpose of computational efficiency. 
\end{itemize}

\chapter{Unsupervised CRF Learning for Stereo Vision with Monocular Cues}
\label{uspv-depth-lrn}


In this chapter, we apply the unsupervised CFR model we introduced in Section~\ref{sec:introlearning} to the problem of 
unsupervised learning of a highly parameterized stereo vision model involving
the monocular shape from texture cues - training the model parameters from unlabeled stereo pair training data\cite{hoang09}.
First, we describe in more details our PBP inference algorithm for this CRF stereo model.
We then give a background review on contrastive divergence, a machine learning technique that plays a key component in our unsupervised CRF learning algorithm.
Next, we present our unsupervised CRF learning of the slanted-plane stereo vision model.
We conclude the chapter by reporting experimental results both for learning and inference on two different stereo datasets.

\section{Particle Belief Propagation for Plane Estimation}
\label{sec:planeinference}

\begin{figure*}[t] \centering
\begin{center}
 \includegraphics[width=0.98\textwidth]{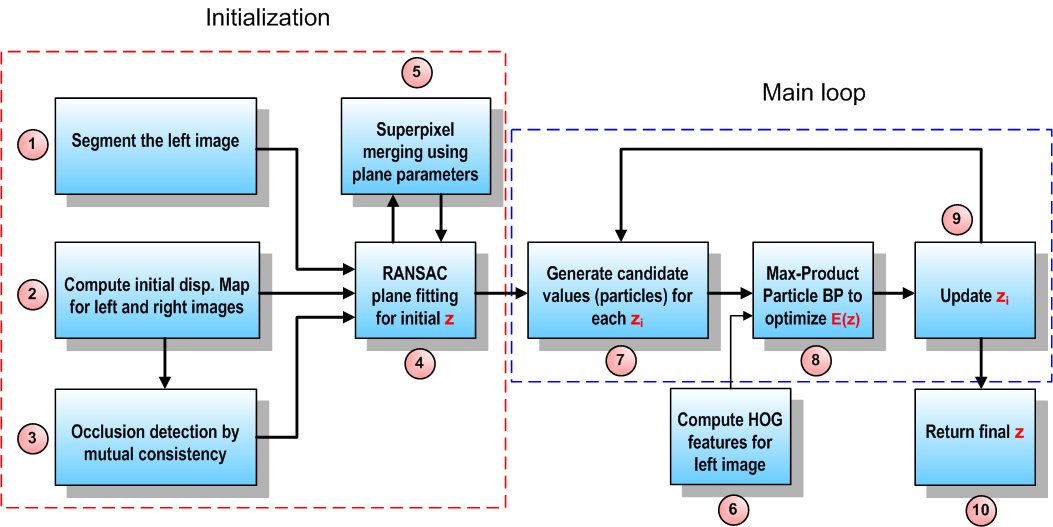}
\end{center}
\caption{Disparity plane inference algorithm using Particle-Belief Propagation.}
\label{fig:PlaneInference}
\end{figure*}

Given a pair of images $(X, Y)$, and a given segmentation of $X$, and a given setting of the model parameters,
the inference problem is to find an assignment $Z$ of plane parameters to superpixels so as to minimize the total energy in (\ref{StereoEq2}). 
The energy defines a Markov random field.  More specifically, the match energy defines the unary potential on each superpixel
independently and the smoothness energy defines the binary potential on pairs of adjacent superpixels.

The complete inference algorithm is illustrated in Figure~\ref{fig:PlaneInference}. The steps of the algorithm include:

\begin{enumerate}
\item We compute a segmentation using the Felzenszwalb-Huttenlocher segmentation algorithm \cite{Pedro04}. Although this segmentation algorithm has a parameter governing the number of superpixels, we do not attempt to tune this parameter with our training algorithm. A more principled approach would include the segmentation itself in the energy functional and tune the segmentation parameter along with the other parameters of the model. However by holding the segmentation parameter fixed, we can treat the segmentation as part of the input data.

\item We run the Felzenszwalb-Huttenlocher's efficient loopy BP stereo algorithm (\cite{Pedro06}) to compute the disparity maps for both images $X$ and $Y$.

\item Mutual consistency check requires that for a particular pixel in the left image, the disparity values between the left and right disparity maps are consistent, i.e:

\begin{eqnarray}
\label{MutualConEq}
d_L(p) = d_R(p - d_L(p))
\end{eqnarray}

The pixel is considered occluded if (\ref{MutualConEq}) does not hold, and considered non-occluded otherwise.

\item The plane fitting is performed in the disparity space, and is applied per superpixel to obtain an initial disparity plane. This is done robustly using RANSAC \cite{RANSAC} on the disparity values of the non-occluded pixels only. (similar to \cite{yang08})

\item This step is optional. In this step we implement a superpixel grouping algorithm  similar to the one used in step 1. However each time we consider merging two adjacent superpixels, we also look at the difference in their disparity planes computed from step 4. Intuitively we prefer merging adjacent superpixels having very similar disparity planes. Note that if used, this step may change the initial segmentation generated by step 1. Then the output is iteratively fed back into the plane fitting in step 4.

\item At each pixel $p$ in the left image $X$, we compute a HOG vector $H(p)$ which is a 24 dimensional feature vector consisting of three 8 dimensional normalized
edge orientation histograms — an 8 dimensional orientation histogram is computed at three different scales.

\item Let $C_i$ be a set of candidate values for $z_i$ derived by repeatedly adding random noise to the current value of $z_i$.

\item Run discrete max-product BP with the finite value set $C_i$ for each node $i$.

\item Set $z_i$ to be the best value for $i$ found in step 8 and repeat.

\item The iteration can be stopped after a fixed number of iterations or when the energy is no longer reduced.
\end{enumerate}

Note that the main loop of the algorithm, including steps 7, 8, 9, is very similar to the Particle-based Belief Propagation algorithm described in Section~\ref{sec:PBP}, which is closely related to previous work in \cite{ihler09} and \cite{koller99}. This can be viewed as the max-product PBP algorithm. In Chapter~\ref{pbp-inference}, we describe in details two other applications of the sum-product PBP inference algorithm to two inference problems in Computer Vision.

\section{Background Review on Contrastive Divergence}
\label{sec:CD}

\subsection{Introduction to Contrastive Divergence}

Assume that our goal is to model the probability of the observed data $x$ using a function of the form $f(x,\beta)$, where $\beta$ is the model parameter vector.
This probability can be defined as follows.

\begin{eqnarray}
\label{eq_cd1}
P(x, \beta) & = & \frac{1}{Z(\beta)} f(x,\beta)
\end{eqnarray}

where $Z(\beta)$, commonly known as the partition function, is defined as

\begin{eqnarray}
\label{eq_cd2}
Z(\beta) & = & \int f(x,\beta) dx
\end{eqnarray}

Given a training dataset $X = \left(x_1, x_2, ..., x_N\right)$, we want to learn the model parameter vector $\beta$ so as to maximize the probability of the training data.

\begin{eqnarray}
\label{eq_cd3}
P(X, \beta) & = & \prod_{i=1}^{N} \frac{1}{Z(\beta)} f(x_i,\beta)
\end{eqnarray}

Taking the negative log of (\ref{eq_cd3}) will give us the following:

\begin{eqnarray}
\label{eq_cd4}
E(X, \beta) & = & \log Z(\beta) - \frac{1}{N}\sum_{i=1}^{N}\log f(x_i,\beta)   
\end{eqnarray}

We denote $E(X, \beta)$, which is the negative log of $P(X, \beta)$, to be the energy function.

The partition function $Z(\beta)$ in equation (\ref{eq_cd2}) is, for many cases, a very complicated function which can be very hard to compute. In cases where this function is intractable, this leads to a situation where we are trying to minimize an energy function that we cannot evaluate.

Contrastive divergence offers a solution to this problem. Briefly speaking, contrastive divergence is a tool that allows us to estimate the gradient of the energy function, even though we cannot evaluate the energy function itself.

\subsection{Formulation of Contrastive Divergence}

As previously mentioned, contrastive divergence provides us a way to estimate the gradient of our energy function $E(X, \beta)$ with respect to the model parameters $\beta$, given the training dataset $X$. Taking the partial derivative of equation (\ref{eq_cd4}) with respect to $\beta$ gives us the following derivation.

\begin{eqnarray}
\label{eq_cd5}
\frac{\partial E(X, \beta)}{\partial \beta} & = & \frac{\partial \log Z(\beta)}{\partial \beta} - \frac{1}{N}\sum_{i=1}^{N}\frac{\partial \log f(x_i,\beta)}{\partial \beta} \\
\nn
& = & \frac{\partial \log Z(\beta)}{\partial \beta} - \left\langle \frac{\partial \log f(x,\beta)}{\partial \beta} \right\rangle_{X}        
\end{eqnarray}

where $\left\langle .\right\rangle_{X}$ denotes the average over the training data X. 
In other words, it is the expectation of $.$ given the training data distribution $X$.

The first term on the right hand side of (\ref{eq_cd5}) involves the partition function $Z(\beta)$. Substituting equation (\ref{eq_cd2}) into this term, we have:

\begin{eqnarray}
\label{eq_cd6}
\frac{\partial \log Z(\beta)}{\partial \beta} & = & \frac{1}{Z(\beta)} \frac{\partial Z(\beta)}{\partial \beta} \\
\nn
& = & \frac{1}{Z(\beta)} \frac{\partial}{\partial \beta} \int f(x,\beta) dx\\
\nn
& = & \frac{1}{Z(\beta)} \int \frac{\partial f(x,\beta)}{\partial \beta} dx\\
\nn
& = & \frac{1}{Z(\beta)} \int f(x,\beta) \frac{\partial \log f(x,\beta)}{\partial \beta} dx\\
\nn
\label{eq_cd61}
& = & \int P(x,\beta) \frac{\partial \log f(x,\beta)}{\partial \beta} dx\\
\nn
\label{eq_cd62}
& = & \left\langle \frac{\partial \log f(x,\beta)}{\partial \beta} \right\rangle_{P(x,\beta)}
\end{eqnarray}

From equation (\ref{eq_cd62}), it is not hard to see that, although the integration in (\ref{eq_cd61}) is algebraically intractable, it can be numerically approximated by drawing samples from the proposal distribution $P(x,\beta)$.
The next problem is that the distribution $P(x,\beta)$ is actually unknown, as we do not know the value of the partition function. In order to resolve this, we can use a Markov Chain Monte Carlo (MCMC) sampling method to simulate draws from the proposal distribution by drawing directly from the target distribution (the training data distribution), then transforming these samples to the proposal distribution. The transformation is done by generating a Markov chain starting from the drawn sample at the target distribution to the proposal distribution. 
Please recall that in the Markov chain, each state $x^{t+1}$ only depends on the previous state $x^t$. At each time step $t$, a new value $x'$ is accepted for the next state $x^{t+1}$ if a random number drawn from $U(0,1)$ satisfies:

$$\alpha < \min \left\{\frac{P(x',\beta)Q(x^t, x')}{P(x^t,\beta)Q(x', x^t)}, 1\right\}$$ 

This involves computing the ratio $\frac{P(x',\beta)}{P(x^t,\beta)}$, which is possible since the partition function cancels out. 
The approximated version of equation (\ref{eq_cd5}) can be written as follows:

\begin{eqnarray}
\label{eq_cd7}
\frac{\partial E(X, \beta)}{\partial \beta} & = & \left\langle \frac{\partial \log f(x,\beta)}{\partial \beta} \right\rangle_{X^{n}} 
- \left\langle \frac{\partial \log f(x,\beta)}{\partial \beta} \right\rangle_{X} 
\end{eqnarray}

where $X^n$ represents the set of samples drawn from the training data after $n$ steps of MCMC.

The use of MCMC sampling strategy, however, can be very inefficient and not practical, since it may take a very long time for the Markov chain to mix and approximately converges. Hinton in \cite{contrastiveA} asserts that only a few MCMC steps would be sufficient to calculate an approximate gradient.
The intuition behind this is that after
a few iterations the data will have moved from the target distribution (i.e. that of the training
data) towards the proposed distribution, and so give an idea in which direction the proposed
distribution should move to better model the training data. Empirically, Hinton has found that
even 1 step of MCMC is often sufficient for the algorithm to converge.

That said, as we use gradient descent in order to minimize our energy function, the contrastive divergence parameter update equation should look like the following:

\begin{eqnarray}
\label{eq_cd8}
\beta_{t+1} & = & \beta_{t} - \gamma \left(\left\langle \frac{\partial \log f(x,\beta)}{\partial \beta} \right\rangle_{X^{1}} 
- \left\langle \frac{\partial \log f(x,\beta)}{\partial \beta} \right\rangle_{X}\right) 
\end{eqnarray}
	
where $\gamma$ is the step size value which is allowed to change at every iteration, based on convergence time and stability.

\section{Our Unsupervised CRF Learning Algorithm}
\label{sec-hardEM}

As described earlier in Section~\ref{sec:slantedplane}, 
our stereo model involves three terms --- a
correspondence energy measuring the degree to which the left and right images agree under the
induced disparity map, a smoothness energy measuring the smoothness of the induced depth map, and an
texture energy measuring the degree to which the surface orientation at each point agrees with a
certain (monocular) texture based surface orientation cue.  For surface orientation cue we use
histograms of oriented gradients (HOG) \cite{Dalal}.
The stereo model itself and the monocular surface cues (as part of the model) are both viewed as the objects being trained.
Our energy functional includes in total 62 parameters - 10 correspondence
parameters, 2 smoothness parameters, and 50 texture parameters.
Among these parameters, we train 50 of them are trained using gradient descent. The gradient of the energy function w.r.t to each parameter is obtained using contrastive divergence, the method described in the previous section. 
The other 12 parameters, including 10 matching parameters ($\lambda_k$ in (\ref{match})) and 2 texture parameters ($\beta_A$ and $\beta_B$ in (\ref{orientation})), are computed by least squares regression.

Our approach to unsupervised learning is based on maximizing conditional likelihood. 
In particular, we consider a general conditional probability model $P_\beta(u|x)$ over arbitrary variables $x$ and $u$ and defined in terms of a parameter vector $\beta$. 
Analogously, CRFs model \cite{Pereira01} defines a conditional probability of the dependent variables $u$ given the exogenous variables $x$ and (importantly) does not model the distribution of the exogenous variables.
In the case of stereo vision one
might expect that it is easier to model the probability distribution of the right image given the
left image than to model a probability distribution over images.

In a closely related earlier work by Zhang and Seitz \cite{Seitz05}, the authors also used a similar CRF model for the classical pixel-based stereo algorithm. Although the training was also based on  maximizing conditional likelihood, they worked with a much lower-dimensional model (5 parameters) and tuned these parameters separately to each single stereo pair as in (\ref{seitzEq}).

\begin{eqnarray}
\label{seitzEq}
\beta^{*}_{i} = \argmax_{\beta}\;\;P_{\beta_{i}}(u_i|x_i)
\end{eqnarray}

The five
parameters are tuned to each individual input stereo pair, although the method could be used
to tune a single parameter setting over a corpus of stereo pairs.
The main difference between their work and ours is that we train highly
parameterized monocular depth cues.  Another difference is that we formulate a
general CRF-like model for unsupervised learning based on maximizing conditional likelihood
and avoid the need
for the independence assumptions used by Zhang and Seitz by using contrastive divergence --- a general
method for optimizing loopy CRFs \cite{contrastiveA,contrastiveB}.

There is also related work on learning highly parameterized monocular depth cues by Saxena et al. in \cite{ashu07,andrew07}, as well as Hoiem et al. in \cite{Hoiem05}.
The main difference between these methods and ours is that we use unsupervised learning while they
use ground truth data to train their system. In Chapter~\ref{experiment-stereo} we will show that our algorithm with unsupervised training outperformed the results in \cite{andrew07}. One might argue that stereo pairs constitute supervised
training of monocular depth cues.  A standard stereo depth algorithm could be used to infer a depth map for each pair
which could then be used in a supervised learning mode to train monocular depth cues.  However, we demonstrate that
training monocular depth cues from stereo pair data improves {\em stereo} depth estimation.  Hence the method can be
legitimately viewed as unsupervised learning of a stereo depth.  Also the general formulation of
unsupervised learning by maximizing conditional likelihood, like the shift from MRFs to CRFs, may have significance
beyond computer vision.

Other related work includes that of Scharstein and Pal \cite{scharstein07} and Kong and Tao \cite{Kong04}.
In these cases somewhat more highly parameterized stereo models are trained using methods developed for general CRFs. 
However, the training uses ground truth depth data rather than unlabeled stereo pairs.

We learn MRF parameters using contrastive divergence \cite{contrastiveA,contrastiveB}, a general MRF learning
algorithm capable of training large models. Another work that also described learning on MRF using contrastive divergence is the Field of Experts system by Roth and Black in \cite{roth05}.
In \cite{roth05}, the authors describe an extension of the traditional MRF model by learning potential functions over extended pixel neighborhoods, which they then implemented with two vision applications: image denoising and image impainting. To training the model parameters, they also used contrastive divergence update, aiming at minimizing the KL-divergence between the model distribution and the data distribution. The MCMC sampling strategy was used to approximate the expectation over model distribution. The data distribution is easy to compute by computing the average over training data. Similar to other aforementioned related work, they also used training data with ground truth label. 

Our training approach is actually more related to the hidden CRF model \cite{Quattoni07} than the standard CRF. Our model includes one more variable, the latent variable $y$ which is not observed in the training data. 
Specifically in our slanted plane stereo model, $x$ denoted segmented left image, $u$ denoted a right image, and $y$ was an assignment of a plane to each superpixel of $x$. $y$ is not in the training data, but it is part of the model.
However, in this chapter we aim at describing our model in a more general level rather than a model specifically applied for stereo vision, as defined by (\ref{genmodel}). 
The conditional probability model $P_\beta(u|x)$ now is defined not only over $x$ and $u$ but also by an arbitrary latent variable $y$.

\begin{equation}
\label{genmodel}
P_{\beta}(u|x) = \sum_y P_{\beta}(u,y|x)
\end{equation}

Given a set of stereo image pairs as training data $(x_1,u_1), \ldots (x_N,u_N)$ conditional EM is an algorithm for locally optimizing the parameter vector $\beta$
so as to maximize the probability of the $u$ values given the $x$ values in the training data, as formulated in equation (\ref{softopt}) (This is exactly equation (\ref{TrainingEq}) from Section~\ref{sec:introlearning}).

\begin{equation}
\label{softopt}
\beta^* = \argmax_{\beta}\;\;\sum_{i=1}^N\;\ln P_{\beta}(u_i|x_i)
\end{equation}

Conditional EM is a straightforward modification of EM and is defined by the following two updates where $\beta$ is initialized with domain specific heuristics.

\begin{eqnarray}
\label{softe}
P_i(y) \!\!\!\!& := \!\!\! & P_{\beta}(y|x_i,u_i) \\
\nn
\label{softm}
\beta \!\!\!\! & := \!\!\! & \argmax_{\beta}\; \sum_{i=1}^N \expectsub{y \sim P_i}{\ln P_{\beta}(u_i,y_i,|x_i)}
\end{eqnarray}

Update (\ref{softe}) is called the E step and update (\ref{softm}) is called the M step.
Hard EM, also known as Viterbi training, works with the single most likely (hard) value of $y$ rather than the (soft) distribution $P_i$ defined by
(\ref{softe}).  Hard conditional EM locally optimizes the following version of (\ref{softopt}). 

\begin{equation}
\label{hardopt}
\beta^* = \argmax_{\beta}\;\;\sum_{i=1}^N\;\max_y\;\ln P_{\beta}(u_i,y|x_i)
\end{equation}

Hard conditional EM is defined to be the process of iterating the updates (\ref{harde}) and (\ref{hardm}) below which can be interpreted as hard versions
of (\ref{softe}) and (\ref{softm}).

\begin{eqnarray}
\label{harde}
y_i & := & \argmax_y P_{\beta}(u_i,y|x_i) \\
\nn
\label{hardm}
\beta & := & \argmax_{\beta}\; \sum_{i=1}^N \ln P_{\beta}(u_i,y_i|x_i) 
\end{eqnarray}

We will call (\ref{harde}) the hard E step and (\ref{hardm}) the hard M step.  Updates (\ref{harde}) and (\ref{hardm}) are both coordinate ascent steps for the
objective defined by (\ref{hardopt}).  However, we refer to (\ref{harde}) and (\ref{hardm}) as hard conditional EM rather than simply ``coordinate ascent'' because
of the clear analogy between (\ref{hardopt}), (\ref{harde}), (\ref{hardm}) and (\ref{softopt}), (\ref{softe}), (\ref{softm}). 


The parameter vector $\beta$ is composed of two components $\beta = (\beta_y,\;\beta_u)$ where $\beta_y$ parameterizes $P(y|x)$ and $\beta_u$ parameterizes $P(u|x,y)$.
Note that the probability model on the right hand side of (\ref{harde}) can be further expanded as follows:

\begin{eqnarray}
\label{factorization1}
y_i & := & \argmax_y P_{\beta_y}(y|x_i)P_{\beta_u}(u_i|x_i,y)
\nn
& := & \argmax_y \ln \left(P_{\beta_y}(y|x_i)P_{\beta_u}(u_i|x_i,y)\right)
\nn
& := & \argmin_y E_{\beta_y}(x_i, y) + E_{\beta_u}(x_i, u_i, y) + \ln Z_{\beta_u}(x, y) + \ln Z_{\beta_y}(x)
\end{eqnarray}

We have that:

\begin{eqnarray}
\label{partition1}
Z_{\beta_y}(x) & = & \sum_y\;e^{(-E_y(x,y,\beta_y)\;)}
\end{eqnarray}

\begin{eqnarray}
\label{partition2}
Z_{\beta_u}(x, y) & = & \sum_u\;e^{(-E_u(x,u,y,\beta_u)\;)}
\end{eqnarray}

where equation (\ref{partition1}) corresponds to the sum of (\ref{smooth}) and (\ref{orientation}), equation (\ref{partition2}) corresponds to (\ref{match}). Since (\ref{partition1}) does not depend on $y$, we can eliminate it from (\ref{factorization1}).
It is less obvious to show that (\ref{partition2}) does not depend on $y$. 

\begin{eqnarray}
\label{partition21}
Z_{\beta_u}(x, y) & = & \sum_u\;e^{(-E_u(x,u,y,\beta_u)\;)}
\nn
& = & \sum_{u}\;e^{-(\sum_{p \in X} (\Phi^x(p) - \Phi^u(p+d(p,y))\;)^2)}
\end{eqnarray}

We assume that the mapping from $x$ to $u$ in (\ref{partition21}) is a bijection, i.e. each pixel $p$ in $x$ only maps to one unique pixel in $u$ (since $x$ and $u$ have the same size, this condition is sufficient to guarantee bijection). 
We can consider the matching cost from equation (\ref{match}) as a distance function between the left image $x$ and one of its prediction $\hat{x}$, where $\hat{x}$ is a function of a pair $(u, y)$. 
Note that (\ref{partition2}) is a sum over all possible values of $u$, and pairing all possible $u$ with $y$ can generate all possible values of $\hat{x}$.
We can rewrite (\ref{partition2}) as follows:

\begin{eqnarray}
\label{partition22}
Z_{\beta_u}(x, y) & = & \sum_u\;e^{(-E_u(x,u,y,\beta_u)\;)}
\nn
& = & \sum_{\hat{x}}\;e^{(-E_{\hat{x}}(x,\hat{x},\beta_u)\;)}
\end{eqnarray}

It follows that (\ref{partition2}) does not depend on $y$ either, and therefore can also be eliminated from (\ref{factorization1}).
We can rewrite (\ref{factorization1}) exactly as:

\begin{eqnarray}
\label{factorization2}
y_i & := & \argmin_y E_{\beta_y}(x_i, y) + E_{\beta_u}(x_i, u_i, y)
\end{eqnarray}

Each of the terms in the right hand side of (\ref{factorization2}) is a CRF. 
In the case of the slanted plane stereo model, the hard E step (\ref{factorization1}) is implemented using a stereo inference algorithm
which computes $y_i$ by minimizing the corresponding energy functionals.
The inference algorithm is described in Section~\ref{sec:planeinference}.  

Our implementation of the hard M step
relies on a factorization of the probability model into two conditional probability models each of which is defined by an energy functional.
Unlike CRFs, we do not require the energy functional to be linear in the model parameters.

\begin{eqnarray}
\label{factorization}
P_{\beta_u,\beta_y}(u,y|x) & = & P_{\beta_y}(y|x)P_{\beta_u}(u|x,y) \\
\nn
\label{genmodela}
P_{\beta_y}(y|x) & =  & \frac{e^{(-E_y(x,y,\beta_y)\;)}}{Z_y(x,\beta_y)} \\
\nn
Z_y(x,\beta_y) & = & \sum_y\;e^{(-E_y(x,y,\beta_y)\;)} \nn
\nn
\label{genmodelb}
P_{\beta_u}(u|x,y) & = & \frac{e^{(-E_u(x,u,y,\beta_u)\;)}}{Z_u(x,y,\beta_u)}  \\
\nn
Z_u(x,y,\beta_u) & = & \sum_u\;e^{(-E_u(x,u,y,\beta_u)\;)}\nonumber
\end{eqnarray}

Given this factorization of the model, the hard M step (\ref{hardm}) can be written as the following pair of updates.

\begin{eqnarray}
\label{hardma}
\beta_y & := & \argmax_{\beta_y} \prod_i P_{\beta_y}(y_i|x_i)
\nn
& := & \argmax_{\beta_y} \sum_i \ln P_{\beta_y}(y_i|x_i)
\nn
& := & \argmin_{\beta_y} \sum_i E_y(x_i,y_i,\beta_y)
\nn
\label{hardmb}
\beta_u & := & \argmax_{\beta_u} \prod_i P_{\beta_u}(u_i|x_i,y_i)
\nn
& := & \argmax_{\beta_u} \sum_i \ln P_{\beta_u}(u_i|x_i,y_i)
\nn
& := & \argmin_{\beta_u} \sum_i E_u(x_i,u_i,y_i,\beta_u)
\end{eqnarray}

Let $L$ abbreviate the quantity being maximized in the right hand side of (\ref{hardma})
and let $E_i(y)$ abbreviate $E_y(x_i,y_i,\beta)$.  We can express the gradient of $L$ as follows.

\begin{equation}
\label{grad}
\nabla_{\beta_y} L = \sum_{i=1}^N \parens{\expectsub{y \sim P_y(y|x_i,\beta)}{\nabla_{\beta_y} E_i(y)} - \nabla_{\beta_y} E_i(y_i)}
\end{equation}

A similar equation holds for (\ref{hardmb}).

The first term in the right hand side of (\ref{grad}) is the expectation over the model distribution, while the second term is the expectation over the training data.
The first term involves $P_{\beta_y}(y|x_i)$, which is a very complicated probability distribution and usually is extremely difficult to compute. However in such cases, we can estimate the expectation of such distributions by taking the average over a sufficient amount of samples, obtained by sampling $y$ from $P_{\beta_y}(y|x_i)$ using an MCMC sampling process.
The regular MCMC sampling method is performed by running a sequence of Metropolis or Metropolis-Hastings sampling steps until the Markov chain mixes - i.e. reaching the stationary distribution. Usually it is a difficult problem to determine how many steps are needed to converge to the stationary distribution within an acceptable error.
Contrastive divergence proposed by G. Hinton et. al. \cite{contrastiveA,contrastiveB} is a technique that allows us to sample $z$ using only one or two MCMC steps rather than a long running MCMC process.
Since we can estimate $\nabla_{\beta_y} L$, we can then optimize (\ref{hardma}), and similarly (\ref{hardmb}), by gradient descent.

For the experiments reported here we use contrastive divergence \cite{contrastiveA,contrastiveB} to sample $y$ rather than a long running MCMC process.
In contrastive divergence we initialize $y$ to be $y_i$ and then perform only a few (one or two) MCMC updates to get a sample of $y$.
Contrastive divergence can be motivated by the observation that if $y_i$ is assumed to be drawn
at random from $P_{\beta}(y|x_i)$ then the expected contrastive divergence update is zero.  So as $\beta$ better fits
the pairs $(x_i,y_i)$ one expects the contrastive divergence gradient estimate to tend to zero.  Furthermore,
because only a few updates are used in the MCMC process, contrastive divergence runs faster and with lower variance than
a longer running MCMC process.
Contrastive divergence yields a consistent estimator for $\beta$ whenever the
expected update direction is the gradient of a convex potential function.

We will demonstrate the experimental results in the next chapter to show that training monocular cues from stereo pair data alone improves stereo depth estimation.
Our unsupervised learning algorithm implicitly trains shape from texture monocular surface orientation cues. 
Moreover, our stereo model with texture cues, only by unsupervised training, outperformed the results in \cite{andrew07}.
Throughout this chapter, stereo vision was considered as a simple setting to investigate unsupervised learning and hence seems a good place to start.
However we believe that our approach to unsupervised learning based on maximizing conditional likelihood can be generalized to other, maybe much more sophisticated models.

\section{Experimental Results}
\label{experiment-stereo}


We implement two training methods in our experiments --- supervised and unsupervised. For each of the supervised and unsupervised training methods we train both a version with texture cues for surface orientation and a version without such cues.
In supervised training we set $z_i$ (for each training image) by fitting a plane in each segment to the ground truth disparities for that segment. We then train the model using a contrastive divergence implementation of the hard M step (\ref{hardm}) which we describe in more detail below.  In supervised training we use only a single setting of $z$ and run one iteration of (\ref{hardm}). In unsupervised training we use the same separation into training and test pairs but do not use ground truth on the training pairs.  Instead we iterate (\ref{harde})
and (\ref{hardm}) six times starting with initial values  for the parameters.

We use the inference algorithm described in section~\ref{sec:planeinference} to implement the hard E step (\ref{harde}). This uses a form of max-product particle belief propagation. Given an assignment $z$ of a plane to each segment, we propose 15 additional candidate planes for each segment by adding Gaussian noise to the plane specified by $z$. The plane parameters $A$ and $B$ have units of pixels of disparity per pixel in the image, and hence are dimensionless.  Typical values of $|A|$ and $|B|$ are from .1 to 1. In the proposal distribution we use Gaussian noise with a standard deviation of .007 for each of $A$ and $B$ and use a deviation of .1 pixels for $C$. We perform six rounds of proposing and selecting.

We implement the hard M step (\ref{hardm}) by first breaking it down into (\ref{hardma}) for training $P(z|x,\beta_z)$ and (\ref{hardmb}) for training $P(y|x,z,\beta_y)$. The form of the match energy (a simple quadratic energy) allows a closed form solution for (\ref{hardmb}).  We implement (\ref{hardma}) by gradient descent using a contrastive divergence approximation of the gradient in (\ref{grad}).  We perform 8 gradient descent parameter updates with a constant learning rate. To estimate the expectation in (\ref{grad}) in each parameter update we generate 10 alternative plane assignments using single MCMC stochastic step starting at $z$ and accepting or rejecting Gaussian noise added once to each plane. The MCMC process proposes a new plane for each segment by adding Gaussian noise and then accepting or rejecting that proposal using the standard Metropolis rejection rule.

\section{Experimental Results with the Middlebury Dataset}


\begin{table*} [h]
  \begin{center}
\begin{tabular}{|l|l|l|l|l|l|l|l|l|l|l|l|l|}
\hline
\multicolumn{3}{|l|}{\scriptsize{Tsukuba}}&\multicolumn{3}{l|}{\scriptsize{Venus}}&\multicolumn{3}{l|}{\scriptsize{Teddy}}&\multicolumn{3}{l|}{\scriptsize{Cones}}&\tiny{Avg.}\\
\cline{1-12} \tiny{nonocc}&\tiny{all}&\tiny{disc}&\tiny{nonocc}&\tiny{all}&\tiny{disc}&\tiny{nonocc}&\tiny{all}&\tiny{disc}&\tiny{nonocc}&\tiny{all}&\tiny{disc}&\tiny{bad}\\
\hline\hline
\scriptsize{2.58}&\scriptsize{4.66}&\scriptsize{11.8}&\scriptsize{0.47}&\scriptsize{0.64} 
&\scriptsize{6.1}&\scriptsize{6.72}&\scriptsize{6.98} &\scriptsize{16.1}&\scriptsize{6.93}&\scriptsize{9.33} &\scriptsize{16.6}&\tiny{7.41}\tiny{\%}\\
\hline
\end{tabular}
  \end{center}
  \caption{Performance on the Middlebury stereo evaluation. The numbers shown are for unsupervised training with texture features.}
\label{resultsb}
\end{table*}

Table~\ref{resultsb} shows the performance of our system on the Middlebury stereo evaluation (version 2).  The numbers shown are for unsupervised training with texture features.
In this case all four images where used as unsupervised training data (ground truth disparities were not used in training). Figure~\ref{fig:stro} shows the inferred depth maps for the Middlebury images at various points in the parameter training.  The figure shows a clear improvement as the parameters are trained.

\begin{figure*} [h]\centering
\begin{tabular}{cccc}
 \includegraphics[width=0.23\textwidth]{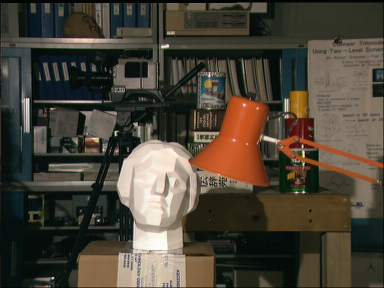} &
 \includegraphics[width=0.23\textwidth]{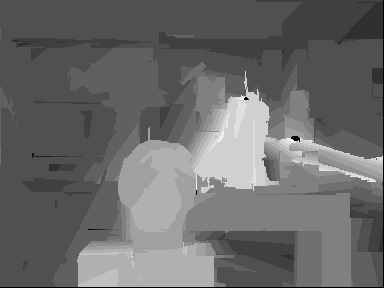} &
 \includegraphics[width=0.23\textwidth]{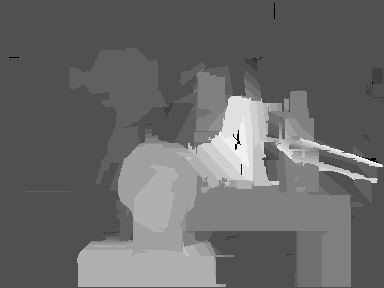} &
 \includegraphics[width=0.23\textwidth]{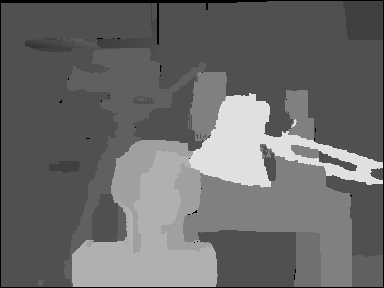} \\
 \includegraphics[width=0.23\textwidth]{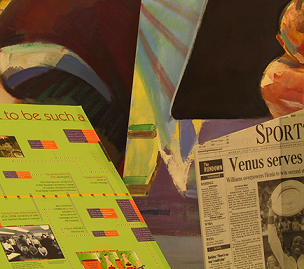} &
 \includegraphics[width=0.23\textwidth]{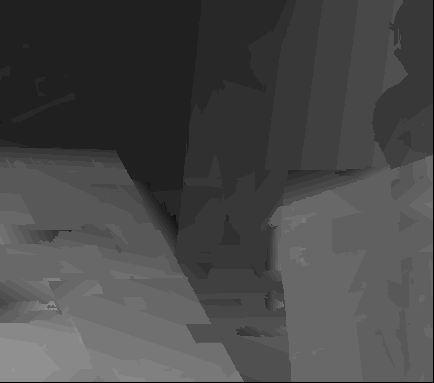} &
 \includegraphics[width=0.23\textwidth]{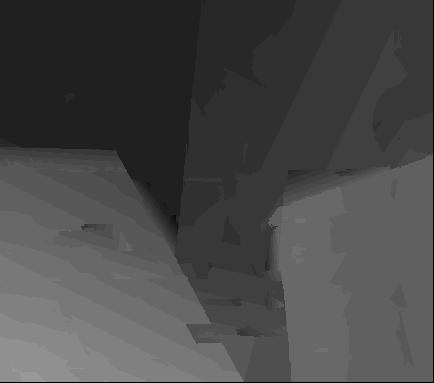} &
 \includegraphics[width=0.23\textwidth]{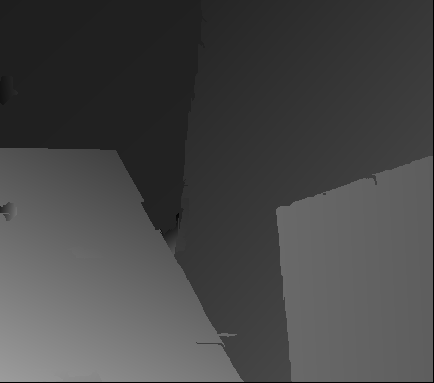} \\
 \includegraphics[width=0.23\textwidth]{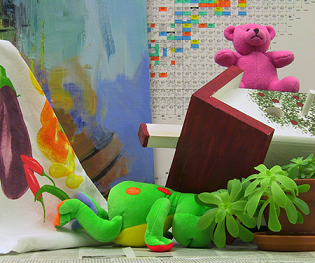} &
 \includegraphics[width=0.23\textwidth]{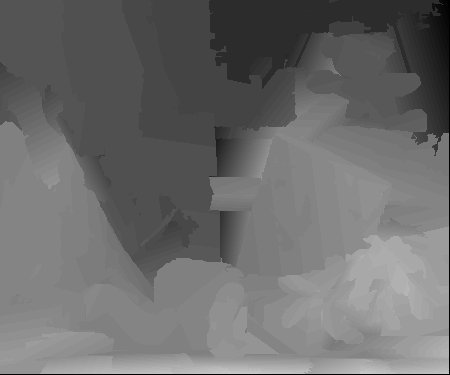} &
 \includegraphics[width=0.23\textwidth]{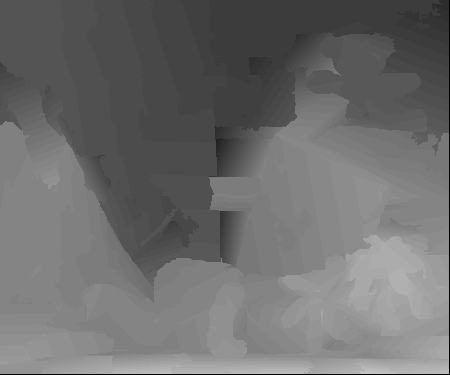} &
 \includegraphics[width=0.23\textwidth]{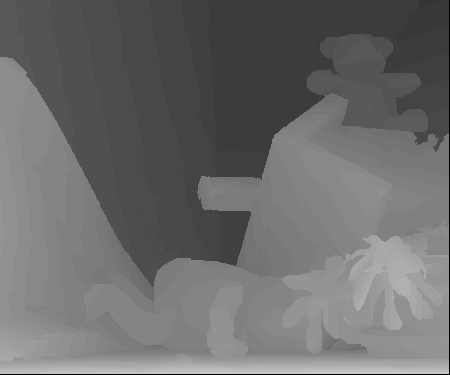} \\
 \includegraphics[width=0.23\textwidth]{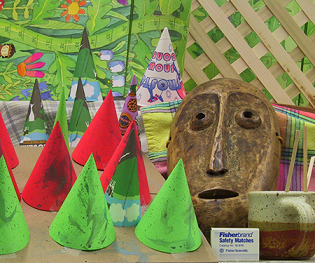} &
 \includegraphics[width=0.23\textwidth]{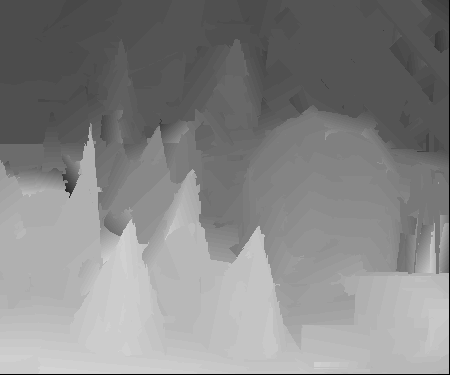} &
 \includegraphics[width=0.23\textwidth]{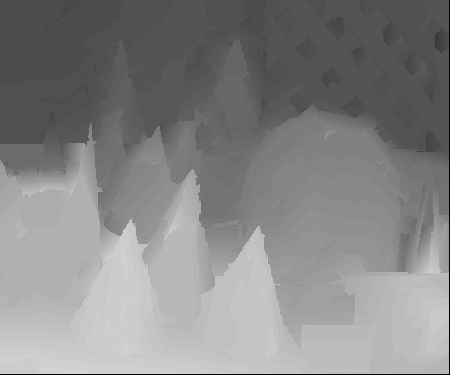} &
 \includegraphics[width=0.23\textwidth]{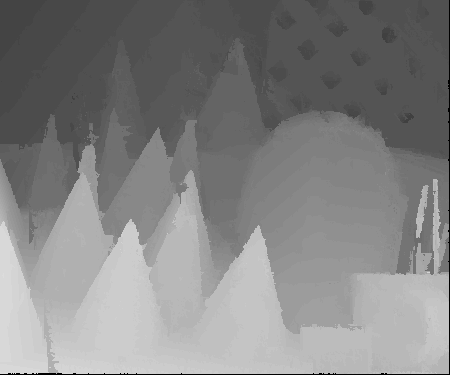} \\
 Left image&Iteration 1&Iteration 3&Iteration 5
\end{tabular}
\caption{Improvement with training on the Middlebury dataset.}
\label{fig:stro}
\end{figure*}

\section{Experimental Results with the Stanford Dataset}

\begin{table} [h]
\begin{center}
\begin{tabular}{|l|l|l|}
\hline
 & RMS & Average\\
 &  Disparity Error &  Error \\
 & (pixels) & {\tiny $|\log_{10}Z - \log_{10}\hat{Z}|$} \\
\hline
Saxena et al. \cite{andrew07} & & .074 \\
\hline
Unsuper., Notexture & 1.286 & .075 \\
\hline
Unsuper., Texture & 1.1608 & .0723\\
\hline
Super., Notexture & 1.1142 & .0709\\
\hline
Super., Texture & 1.0319 & .0686\\
\hline
\end{tabular}
\end{center}
\caption{RMS disparity error (in pixels) and average error (average base 10 logarithm of the multiplicative error) on the Stanford stereo pairs
for four versions of our systems plus the best reported result from~\cite{andrew07} on this data.  Each system was either trained using the ground truth depth
map (supervised) or trained purely from unlabeled stereo pairs (unsupervised) and either used texture cues (Texture) for surface orientation or did not (Notexture).
Note that texture information helps improve the performance in both supervised and unsupervised cases.}
\label{resultsa}
\end{table}

We have also run experiments on a set of stereo image pairs taken from the Stanford color stereo dataset \footnote{http://ai.stanford.edu/~asaxena/learningdepth/data} which has been used to train monocular depth estimation \cite{ashu05,ashu07,andrew07}.  The images cover different types of outdoor scenes (buildings, grass, forests, trees, bushes, etc.) and some indoor scenes. 

First we performed epipolar rectification on this dataset. 
Given a pair of stereo images, rectification computes a
transformation matrix for each image plane such that pairs of conjugate
epipolar lines become collinear and parallel to one
of the image axes (usually the horizontal one). The rectified
images can be thought of as acquired by a new stereo
rig, obtained by rotating the original cameras. The important
advantage of rectification is that computing stereo correspondences is made simpler,
because search is done along the horizontal lines of the rectified
images.

Standard rectification methods such as \cite{Fusiello} rely on detecting and matching a sparse set of feature points to compute the epipolar geometry between the left and right image. The rectification then is achieved usually by minimizing a measure of distortion  - a score function involving only the point correspondences.

Using a rectification kit from Fusiello et al. \cite{Fusiello} failed in 57 out of 257 stereo pairs in the Stanford dataset (The failed cases are cases where the rectification score function exceeded a specified threshold). Therefore we came up with a different rectification approach that we call "`dense rectification"'. 
First we observe that all stereo pairs in the dataset were taken from a stereo rig with almost the same calibration parameters. We then aim at determining one  or only a few rectification solutions for all pairs in the dataset, rather than computing a separate rectification for each pair. Besides, since epipolar rectification is considered an important preprocessing stage of dense stereo matching, the ultimate goal of rectification is also to optimize the dense stereo matching score. Therefore our method strives to find the rectification solutions that directly serve for this purpose: minimizing the dense stereo energy score rather than the feature point correspondence distortion score. To compute the dense stereo energy score for each pair, we simply use the efficient stereo algorithm in \cite{Pedro06}. Our approach can be summarized as follows:

\begin{itemize}	
\item  Run rectification code from Fusiello et al. \cite{Fusiello} on the whole dataset.
\item  Use stereo energy to pick the best 200 image pairs. (Consider the other 57 images outliers)
\item  Do K-means clustering on the 200 transformation matrices of these 200 image pairs with $(K = 2, 3, 4)$.
\item  Pick best K = 2. Compute the 2 mean transformation matrices.
\item  Apply each of the transformation matrices to rectify the 257 images. For each image we pick the rectified one with better stereo energy score (instead of the rectification score).
\item  We obtain 257 rectified image pairs.	
\end{itemize}

With our rectification algorithm, we successfully rectified all stereo image pairs in the dataset. The rectified dataset has been put online for public use  \footnote{http://ttic.uchicago.edu/~ntrinh/shared/stanford/}. Our results are directly comparable to \cite{andrew07}, i.e we use the same training set and test set (193 pairs for training, 64 pairs for testing). Table~\ref{resultsa} shows that we outperform their results.
 
\begin{figure*} [h]\centering
\begin{tabular}{ccc}
 \includegraphics[width=0.32\textwidth]{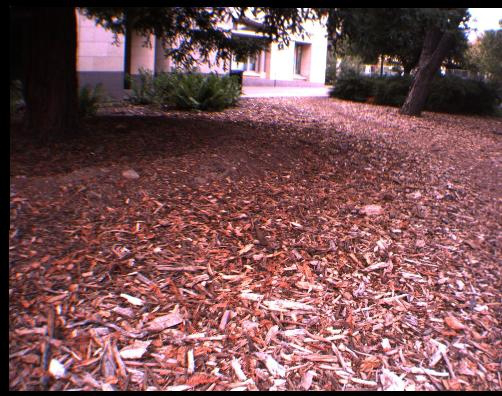} &
 \includegraphics[width=0.32\textwidth]{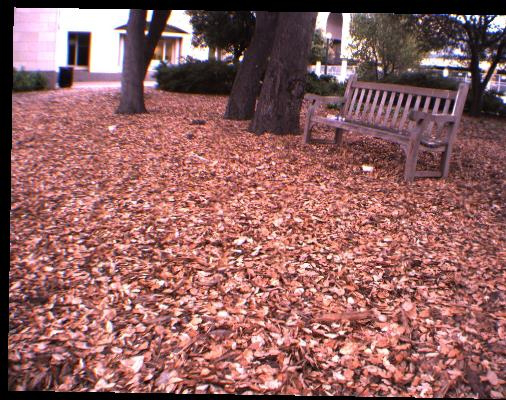} &
 \includegraphics[width=0.32\textwidth]{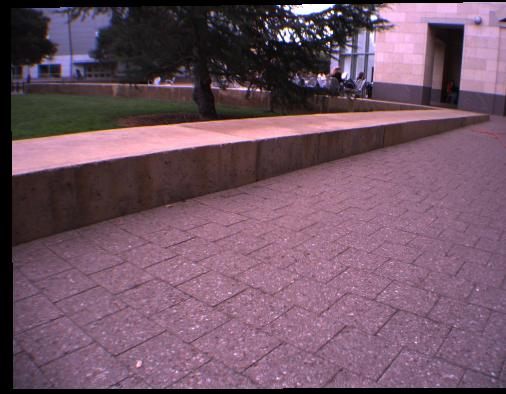} \\
 \includegraphics[width=0.32\textwidth]{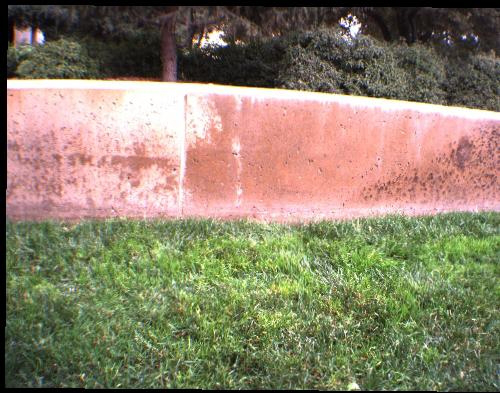} &
 \includegraphics[width=0.32\textwidth]{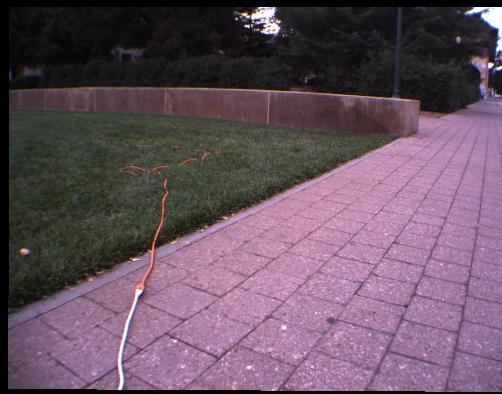} &
 \includegraphics[width=0.32\textwidth]{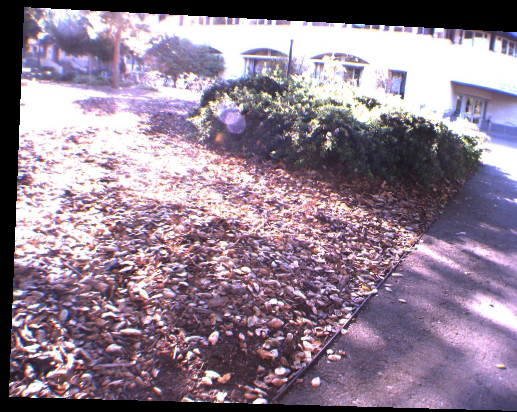} \\
 \includegraphics[width=0.32\textwidth]{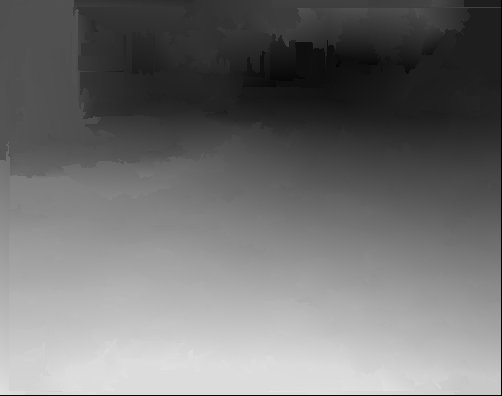} &
 \includegraphics[width=0.32\textwidth]{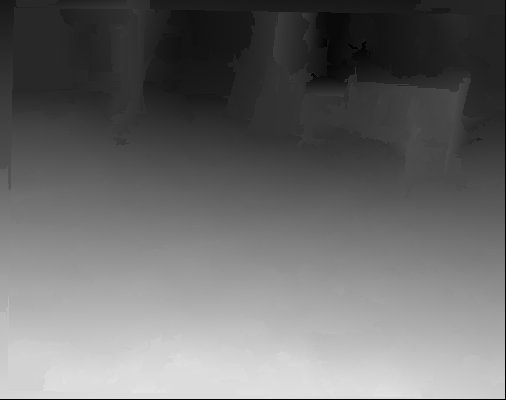} &
 \includegraphics[width=0.32\textwidth]{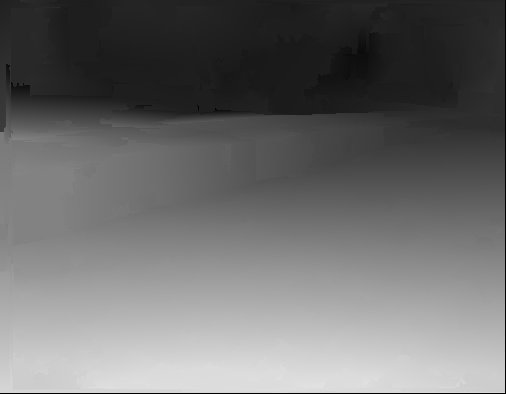} \\
 \includegraphics[width=0.32\textwidth]{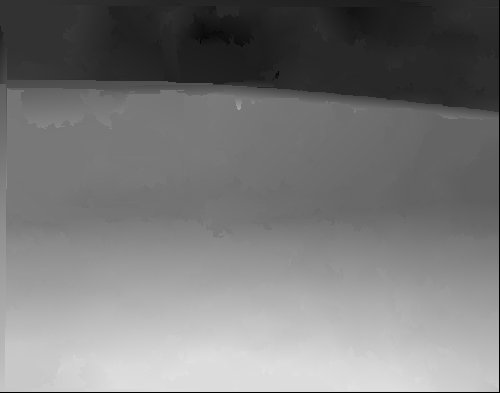} &
 \includegraphics[width=0.32\textwidth]{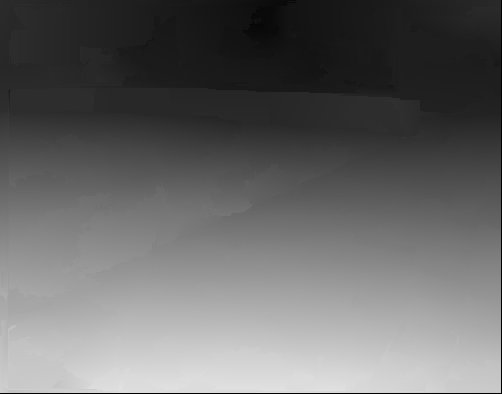} &
 \includegraphics[width=0.32\textwidth]{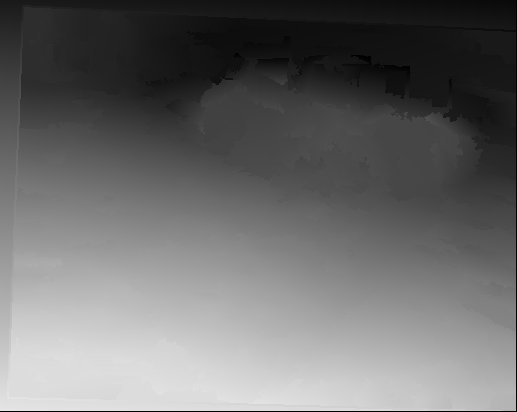}
\end{tabular}
\caption{Several depth map results on the rectified Stanford stereo dataset. The first 2 row are some example images. The second 2 rows are our output depth maps.}
\label{fig:stroStanford}
\end{figure*}

\section{Conclusion}

In many applications we would like to be able to build systems that learn
from data collected from mechanical devices such as microphones and cameras.  Stereo vision provides perhaps the simplest setting in which
to study unsupervised learning.  We have formulated an approach to unsupervised learning based on maximizing conditional likelihood
and demonstrated its use for unsupervised learning of stereo depth with monocular depth cues.
Ultimately we are interested in learning highly parameterized sophisticated models including, perhaps, models of surface types,
shape from shading, albedo smoothness priors, lighting smoothness priors, and even object pose models.  We believe that unsupervised learning
based on maximizing conditional likelihood can be scaled
to much more sophisticated models than those demonstrated here.



\chapter{Unsupervised CRF Learning for Monocular Depth Estimation}
\label{mno-depth-est}

In this chapter we present our unsupervised CRF learning approach to the problem of monocular depth estimation (MDE). MDE is the problem of predicting the depth at each pixel of a single input image. This is a much more challenging problem than stereo vision. There exist an intrinsic ambiguity between local image information and the 3-D location of the real-world object, due to perspective projection. In other words, a point in the image plane can correspond to any 3-D point lying on a line connecting the camera center and that image point.

In recent work by Saxena et al. \cite{ashu05,ashu07,andrew07} an MDE system is trained from images paired with laser range finder depth maps.
Here we are interested in training an MDE system using only stereo pairs --- pairs of images taken from slightly different angles.
Stereo pair training for MDE is interesting for a variety of reasons.
First, in some cases it may be easier to obtain stereo pairs than to obtain images paired with laser
range finder data.  Second, learning MDE from stereo pairs can be viewed as a first step toward
learning MDE from video sequences.
Video is easily obtained and widely available and could provide an enormous volume of training data.
Third, stereo pair training for MDE serves as a case study in unsupervised two-view learning which might provide a model for
this form of learning in other applications.

We train a monocular depth estimator using the same stereo pair database as used in \cite{andrew07} but without the use of ground truth depth maps.
For learning we use basically the same unsupervised CRF learning algorithm we applied for Stereo Vision with Monocular Cues in Chapter~\ref{uspv-depth-lrn}.
Our hard EM algorithm first estimates depth using stereo alone (E step), then trains the monocular predictor (M step), then re-estimates depth using both stereo and monocular cues, then retrains the monocular predictor, and so on.
While most of the performance of the monocular predictor is achieved by training on the initial stereo depth estimates, a modest improvement is seen for both view prediction and for ground truth prediction at later EM iterations.  The performance on ground-truth prediction is similar to that reported in \cite{andrew07} in spite of the absence of ground truth in training.

In this chapter, first we present our depth estimation model with monocular depth cues. 
Next, we introduce the notion of view prediction and its relation to the conditional likelihood that the learning algorithm wants to maximize.
We then describe the application of our unsupervised CRF learning algorithm for this specific model, aiming at training the MDE.
The last section demonstrates experimental results.

\section{The model}

Our model is simply an extension of the standard dense stereo model. 

\begin{eqnarray}
\label{eqn:EnergyMDE}
E_w(I^{(1)},I^{(2)},d) & = & \left\{\begin{array}{l} 
\lambda_d \sum_{p} \min\parens{||Y_i^{(1)}(p)-Y_i^{(2)}(p+d_i(p))||^2,\tau_d} \\
+ \lambda_s \sum_{p,q\in N(p)} min(\left|d(p) - d(q)\right|,\tau_{s}) \\
+ \sum_{p}(d(p)-w\cdot X^{(1)}(p))^2
\end{array}\right.
\end{eqnarray}

where $(I^{(1)},I^{(2)})$ is a stereo pair, $d$ is the depth map of $I^{(1)}$, $d(p)$ is the depth value assigned to pixel $p$, 
$Y_i^{(1)}(p)$ is the image feature vector corresponding to pixel $p$ in $I^{(1)}$,
$X^{(1)}(p)$ is the monocular feature vector of $p$, $N(p)$ denotes the set of neighbors of pixel $p$,
and $w$ is the MDE parameter. 
For this current model, we assume that other parameters $\lambda_d, \lambda_s, \tau_d, \tau_d$ are fixed, and the only parameter we want to learn is $w$. 

Note that (\ref{eqn:EnergyMDE}) defines an MRF, in which the features are stereo, smoothness and monocular features, the hidden labels are the pixel depth values. The first term in (\ref{eqn:EnergyMDE}) is the match energy measuring how well the disparity assignment $d$ agrees
with the input image pair, the second term is the smoothness energy enforcing the smoothness assumption, and the third term is the MDE term.
For MDE we want to construct a disparity map $d$ given only the first view $I^{(1)}$. 
We assume a feature vector $X^{(1)}(p)$ at each pixel $p$ of $I^{(1)}$. A particular choice of features will be discussed
in Section~\ref{sec:experiments} but we can think of $X^{(1)}(p)$ as containing a constant feature (a bias feature),
information about the $y$ component of the pixel $p$, and the value of various filters applied to image $I^{(1)}$ at the pixel $p$.
Our MDE energy term now is the sum squared difference between the assigned disparity $d(p)$
and a locally predicted disparity $w \cdot X(w)$.

We now define a monocular disparity predictor $d^*_w(I^{(1)})$ determined by a parameters vector $w$ as follows.

\begin{align} \label{eqn:Energya}
 d_w^{\ast}(I^{(1)}) &= \argmin_{d}\;\;(\sum_{p}(d(p)-w\cdot X^{(1)}(p))^2 + \lambda_s \sum_{p,q\in N(p)} min(\left|d(p) - d(q)\right|,\tau_{s})
\end{align}

The second term here is exactly the robust $L_1$ smoothing term in (\ref{eqn:EnergyMDE}). (\ref{eqn:Energya}) itself also defines an MRF, without the stereo feature.

\section{Stereo Pair View Prediction and Conditional Likelihood}
\label{sec:stereo-view-prediction}

Here we introduce the notion of view prediction.
In view prediction the goal is to predict the second view given the first. View prediction training
can be expressed with the following schema where the distortion and complexity functions are application specific.

\begin{equation}
\label{colabelinga}
f^* = \argmin_{f \in {\cal F}} \; \sum_{i=1}^M \mathrm{Distortion}\parens{f\parens{x^{(1)}_i},x^{(2)}_i} + \lambda \mathrm{Complexity}(f)
\end{equation}

In the system developed here we take $f\parens{x^{(1)}}$ to be an image derived by applying an inferred disparity map to the image $x^{(1)}$
and the distortion function is simply the pixel-wise squared error (the $L_2$ distance) between $x^{(2)}$ and $f\parens{x^{(1)}}$.
A log-loss version of view prediction can be formulated as follows where $P_w(x^{(2)} |x^{(1)})$ is a conditional probability (or probability density) parameterized by $w$.

\begin{equation}
\label{colabelingb}
w^* = \argmin_w \; \sum_{i=1}^M \ln\frac{1}{P_w\parens{x^{(2)}|x^{(1)}}} + \lambda \mathrm{Complexity}(w)
\end{equation}

For stereo pairs we might define $P_w\parens{x^{(2)}|x^{(1)}}$ by inferring a disparity map from $x^{(1)}$ under parameter setting $w$ and then using the inferred disparity map to predict a probability density in color space at each pixel of $x^{(2)}$.
Note that view prediction training as defined by (\ref{colabelinga}) or (\ref{colabelingb}) is meaningful without labels.

We consider stereo pairs of the form $(I^{(1)},I^{(2)})$ where $I^{(1)}$ and $I^{(2)}$ are images each of which is an assignment of a color vector to pixels.  We write $I^{(j),c}(p)$ for the color $c$ component at pixel $p$
in image $I^{(j)}$. 
Let $d$ denote a disparity map --- a function that assigns a disparity $d(p)$ to
each pixel $p$.  Disparity defines a mapping between the two images. The position $(x,y)$ in $I^{(1)}$ corresponds to the position
$(x,y+d(x,y))$ in image $I^{(2)}$.  We let $p$ range over pixels where each pixel is defined by a pair of integer coordinates $(x,y)$
and we write $p + d(p)$ as an abbreviation for $(x,y+d(x,y))$ where $(x,y)$ is the pixel $p$.  We assume that $d(p)$ is integral.
We will mostly work with disparity $d$ rather than depth $Z$ where $d$ and $Z$ are related
by $Z = bf/d$ where $b$ is the baseline (the distance between the two cameras) and $f$ is the focal length.
We assume that $b$ and $f$ are known.

Given a disparity map $d$ and a reference image $I^{(1)}$ we construct a prediction $\hat{I}^{(2)}$ which approximately satisfies the
following equation.

\begin{align} \label{eqn:gainbias}
\hat{I}^{(2),c}(p + d(p)) = \frac{\sigma^{(2),c}}{\sigma^{(1),c}}(I^{(1),c}(p) - \bar{I}^{(1),c}) + \bar{I}^{(2),c}
\end{align}

Here $\bar{I}^{(1),c}$ and $\sigma^{(1),c}$ are the mean and standard deviation of all the color $c$
pixel values in image $I^{(1)}$. The quantities $\bar{I}^{(2),c}$ and $\sigma^{(2),c}$ are defined similarly for image $I^{(2)}$.
The use of pixel means and variances for the two images corrects for different biases and gains,
in each color channel separately, of the two cameras.  We could have assumed knowledge of camera bias
and gain of the two cameras rather than knowledge of pixel mean and variance of $I^{(2)}$.
But in any case the mean and variance of the pixels in $I^{(2)}$ is only a very small amount of information about $I^{(2)}$.
Note that equation (\ref{eqn:gainbias}) both overspecifies and underspecifies $\hat{I}^{(2)}$ as a function of $I^{(1)}$ and $d$.
Overdetermined values in $\hat{I}^{(2)}$ are set to the pixel value in $I^{(1)}$ with largest disparity (the closest point) and
missing values in $\hat{I}^{(2)}$ are filled in from neighboring values preferring values derived from pixels with low disparity (farthest points).

The overall view predictor $f_w\parens{I^{(1)}}$ is then defined by (\ref{eqn:gainbias})
applied to $I^{(1)}$ and $d^*_w\parens{I^{(1)}}$. We will work with the following distortion function where $P$ is the number of pixels
and $C$ is the number of colors.

\begin{equation}
\label{eqn:distortion}
\mathrm{Distortion}(\hat{I}^{(2)},I^{(2)}) = \frac{1}{PC} \sum_{p,c} \frac{(\hat{I}^{(2),c}(p) - I^{(2),c}(p))^2}{\parens{\sigma^{(2),c}}^2}
\end{equation}

Here we have defined distortion to be a percentage of variance.  Note that if we take $\hat{I}^{(2)}$ to be simply the mean pixel value of $I^{(2)}$
then we get exactly 100\% distortion.

We have pointed out that the view prediction error (\ref{eqn:distortion}) corresponds directly to the conditional probability $P_w\parens{I^{(2)}|I^{(1)}}$, i.e. the probability of the right image given the left image. This probability is exactly the conditional probability in our unsupervised CRF model:

\begin{eqnarray}
\label{TrainingEqMDE}
\beta^* = \argmax_{\beta}\;\;\sum_{i=1}^N\;\ln P_{\beta}(u_i|x_i)
\end{eqnarray}

where $u$ corresponds to the right image $I^{(2)}$, $x$ corresponds to the left image $I^{(1)}$, and $\beta$ is the model parameter, which corresponds to $w$.

\section{Unsupervised CRF Learning for MDE Model}
\label{sec:bootstrap}

We are now interested in training the parameter vector $w$ so as to minimize the expected distortion (\ref{eqn:distortion})
over fresh stereo pairs.  We assume a collection of training
pairs 
$$\parens{I^{(1)}_1,I^{(2)}_1},\ldots,\parens{I^{(1)}_N,I^{(2)}_N}$$.  

We now formulate the hard EM algorithm which first estimates disparity
using a classical (untrained) stereo algorithm (E step), then trains a monocular estimator from the inferred disparity map (M step), then re-estimates disparity
using both binocular and monocular cues, retrains the monocular predictor, and so on. We formulate this EM bootstrap algorithm as coordinate
descent on a joint energy function of both the disparity map $d$ and the parameters $w$.  Formulating this algorithm as coordinate descent guarantees convergence.

The joint monocular and binocular energy function was defined in (\ref{eqn:EnergyMDE}). 
In our experiments we take $Y_i^{(j)}(p)$ to consist of the color vector at $p$ and the gradient of each color channel at $p$
for a total of $3 + 6 = 9$ feature values for three colors.\footnote{Before computing color vector and gradient features the images are normalized
to remove bias and gain effects by subtracting the mean color and then dividing each color channel by the standard deviation of that color channel.}
The hard EM learning algorithm is then defined by the following optimization problem.

\begin{eqnarray}
\label{eqn:Energyb}
w^* & = & \argmin_w \; \sum_{i=1}^N \;\min_{d}  E_w(I_i^{(1)},I_i^{(2)},d)
\end{eqnarray}

Our training algorithm is coordinate descent on (\ref{eqn:Energyb}).  We alternatively optimize the ``coordinates''
$w$ and $d_i$.  We initialize $d_i$ to be the disparity map minimizing the first two terms of (\ref{eqn:EnergyMDE}).
This corresponds to classical (untrained) binocular inference of the disparity map.  Given an initial value
for the disparity maps $d_i$ we alternate the following two updates.

\begin{eqnarray}
\label{updatea}
w & := & \argmin_w\;\sum_{i=1}^N E_w(I_i^{(1)},I_i^{(2)},d_i) \\
 & = & \argmin_w\;\sum_{i=1}^N \sum_p (d_i(p) - w \cdot X_i^{(1)}(p))^2
\nn
\label{updateb}
d_i & :=  & \argmin_d \;E_w(I_i^{(1)},I_i^{(2)},d)
\end{eqnarray}

Note that (\ref{updatea}) is a least squares regression problem that can be solved exactly with standard codes.
Also note that regularization of $w$ can easily be incorporated into (\ref{updatea}).
We approximately solve (\ref{updateb}) using the efficient loopy BP algorithm of Felzenszwalb and Huttenlocher \cite{Pedro06}.
To relate this to the learning algorithm described earlier in Chapter~\ref{uspv-depth-lrn}, 
please note that equation (\ref{updatea}) corresponds exactly to equation (\ref{hardm}),
and (\ref{updateb}) corresponds exactly to equation (\ref{harde}).

For the experiments reported in Section\ref{sec:experiments}, parameters of the algorithm other than $w$ are set by hand.

\section{Experimental Results} \label{sec:experiments}

\begin{table*}[t]
  \begin{center}
\begin{tabular}{|l|l|l|l|l|l|}
\hline
 &&\multicolumn{4}{l|}{\hspace{30pt} Number of iterations}\\
\cline{3-6}
 &&1&2&3&4\\
 \hline
Baseline&View Prediction Error&0.458&-&-&-\\
&Ground Truth RMS&8.644&-&-&-\\
 \hline
\textbf{Monocular}&View Prediction Error&0.428&0.419&0.418&0.418\\
&Ground Truth RMS&1.933&1.921&1.921&1.921\\
\hline
\end{tabular}
  \end{center}
  \caption{Quantitative comparison between the ground plane baseline and our model using the average testing error per pixel as functions of the number of iterations. For each model we report: (a) The average view prediction error, measuring the predicted right frame and the ground truth right frame, i.e. the quantity on the left side of (\ref{eqn:distortion}). (b) The RMS disparity error compared with the ground truth (in pixels), measuring the average difference per pixel in disparity value between the predicted disparity and the ground truth disparity.}
  \label{tab:vperr}
\end{table*}

\begin{figure*}[t] \centering
\begin{tabular}{cccc}
 \includegraphics[width=0.2\textwidth]{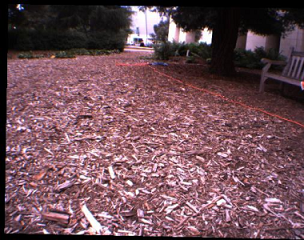} &
 \includegraphics[width=0.2\textwidth]{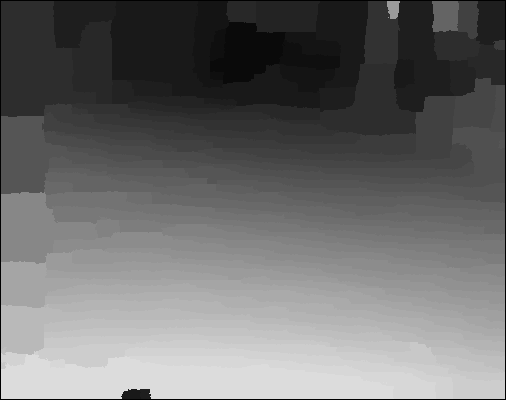} &
 \includegraphics[width=0.2\textwidth]{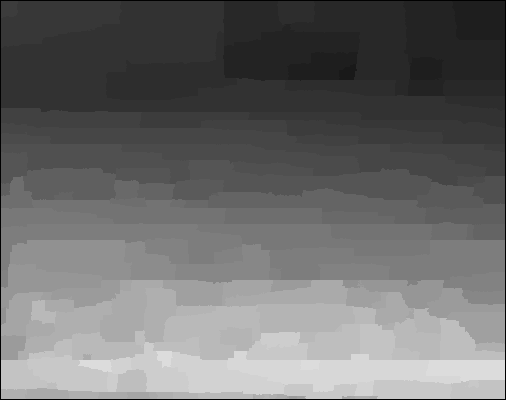} &
 \includegraphics[width=0.2\textwidth]{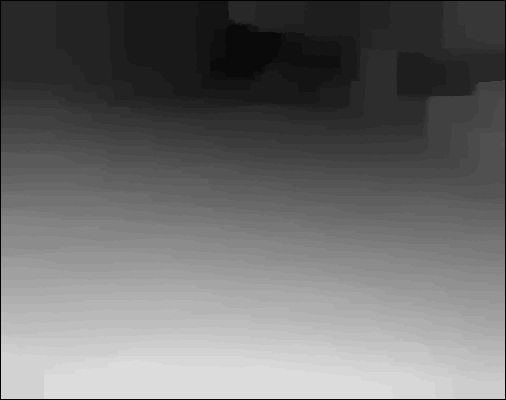} \\
 \includegraphics[width=0.2\textwidth]{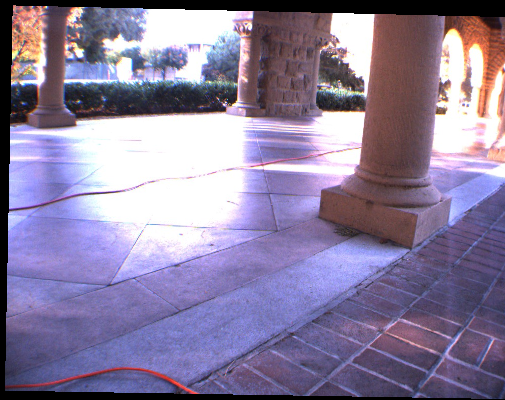} &
 \includegraphics[width=0.2\textwidth]{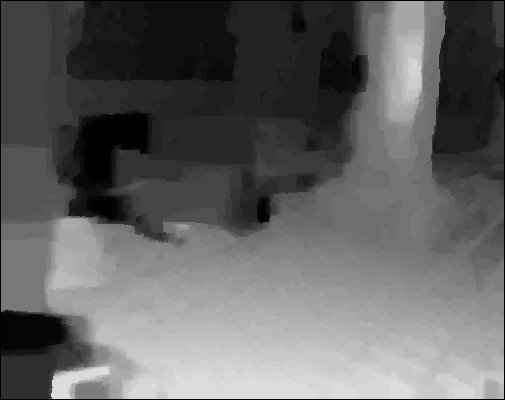} &
 \includegraphics[width=0.2\textwidth]{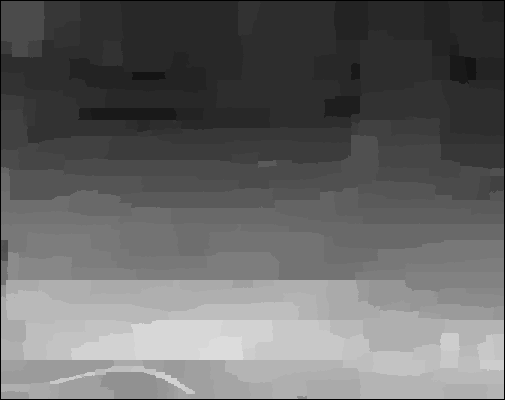} &
 \includegraphics[width=0.2\textwidth]{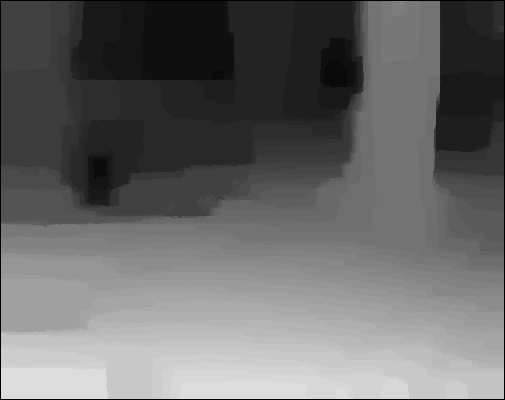} \\
 \includegraphics[width=0.2\textwidth]{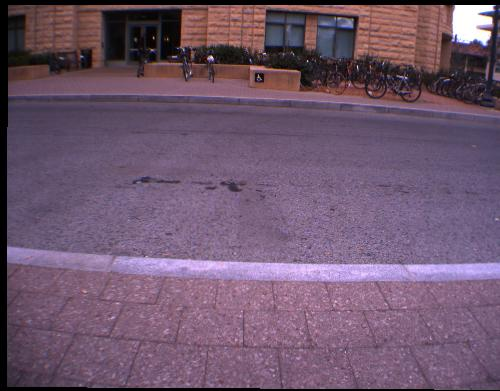} &
 \includegraphics[width=0.2\textwidth]{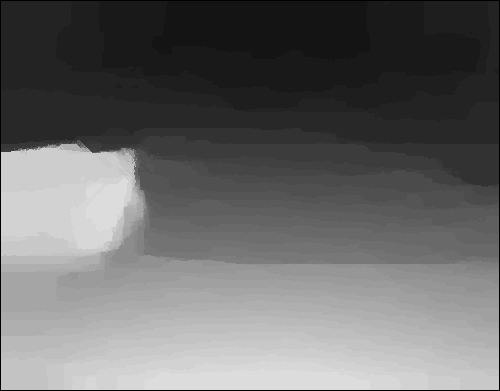} &
 \includegraphics[width=0.2\textwidth]{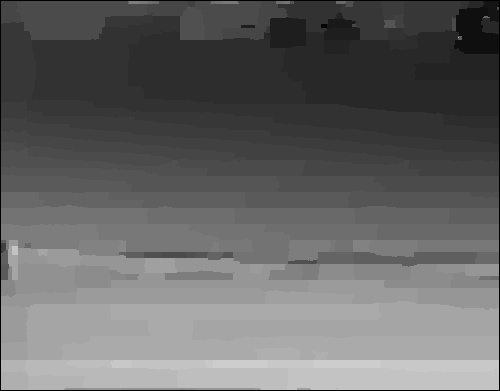} &
 \includegraphics[width=0.2\textwidth]{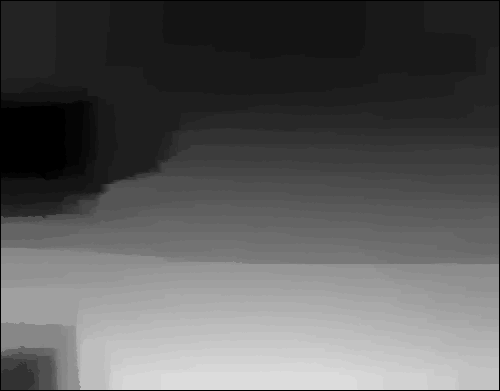} \\
 Left image&Stereo depthmaps&Monocular depthmaps&Mono+Stereo depthmaps
\end{tabular}
\caption{Resulting disparity maps from (a) the initial stereo algorithm, (b) the trained monocular predictor, (c) the combined monocular and binocular predictor.} 
\label{fig:im}
\end{figure*}

We ran the EM bootstrap training algorithm on a set of rectified stereo pairs
taken from a color stereo dataset.\footnote{The dataset is available at http://ai.stanford.edu/~asaxena/learningdepth/data.}
The images cover different types of outdoor scenes (buildings, grass, forests,
trees, bushes, etc.), and some indoor scenes. The images were
epipolar rectified using a rectification kit from Fusiello et al. \cite{Fusiello}.
Initial disparity maps were computed for all images using Felzenszwalb and Huttenlocher's efficient
loopy BP algorithm on the first two energy terms of~\eqref{eqn:EnergyMDE}.
By examining the resulting rectifications and depth maps it was clear that some rectifications
had failed.  At this point we removed from the dataset all pairs for which the energy value
achieved by loopy BP was above a specified threshold.  This left 204 out of an original 250 stereo pairs
on which to train the monocular depth estimator.

The monocular feature vector $X^{(1)}(p)$ consists of local 
image features computed at 3 different scales, which capture the 
texture and color cues (similar to \cite{andrew07}).
We also used the pixel image coordinates as features and included 1 constant feature for a bias term.
We have $X^{(1)}(p) \in R^{54}$ (17 image features at each 
scale for 3 scales, plus 2 spatial features and 1 bias feature).
Following \cite{andrew07}
we made disjoint copies of this feature set for different height levels in the image
so that the monocular predictor is trained separately for each height level.
We use 40 discrete height levels.

Training the monocular predictor separately at different height levels exploits the fact that
these images were taken from a single robot with a fairly stable ground plain orientation relative to the cameras.
The stability of the ground plane in these images suggests a baseline monocular prediction in which the
disparity at each height level is simply predicted to be the average disparity for that height level across all training data
as estimated in the learning algorithm.  We will call this the ground plane baseline (even though it is not really a ground plane).
The ground plane baseline is stronger than the baseline used in \cite{ashu07}.

We performed K-fold cross-validation using hard EM stereo pair training and view prediction testing.
For each fold, $80\%$ of the data was used for training and $20\%$ was used for
holdout testing.  For testing, monocular inference was performed using loopy BP
to minimize the energy function described in~\eqref{eqn:Energya}.
Some of the resulting monocular disparity maps for holdout test stereo pairs are shown in
Figure~\ref{fig:im}, demonstrating the ability of our monocular
predictor to predict depth from different scenes.  For the test pairs we also computed disparity maps
using binocular only and combined monocular and binocular cues. The disparity prediction system combining both monocular and
stereo cues appears to improve the accuracy of the
disparity obtained from the initial stereo algorithm, although we did not have quantitative ground truth for the
test pairs.

The hard EM algorithm
was run for four iterations of bootstrapping resulting in four weight vectors.
Each monocular feature weight vector was tested by two measures on test data.
First, we measured the view prediction error using formula (\ref{eqn:distortion}) on the holdout stereo pairs.
Second, we computed an error relative to ground truth on images from a second data set of 63 single images
paired with laser range finder depth maps.\footnote{We picked all the images with prefix
  \textit{'img-stats'} from the dataset described in
  \cite{ashu07}.}
We related disparity $d$ to depth $Z$ using $Z = c/d$ where the constant $c$ is determined by a fit to the
entire data set (the same value of $c$ is used for all test images). We reported RMS disparity error which is the square root of the average over pixels of $(d - c/Z)^2$.
The results, averaged across all folds, are shown in Table~\ref{tab:vperr}.

Table~\ref{tab:vperr} shows that our algorithm converges quickly after only a
few iterations. The prediction error improves for at least one iteration of training. 
This again suggests that the monocular cues improve the 
disparity estimation relative to binocular cues only.
The table shows a dramatic improvement in RMS disparity error over the ground plane baseline. 
We obtained the average error in disparity estimate of 2 pixels, much better than the performance of the ground plane baseline (8.6 pixels). 

\begin{figure*}[t] \centering
\begin{tabular}{ccc}
 \includegraphics[width=0.25\textwidth]{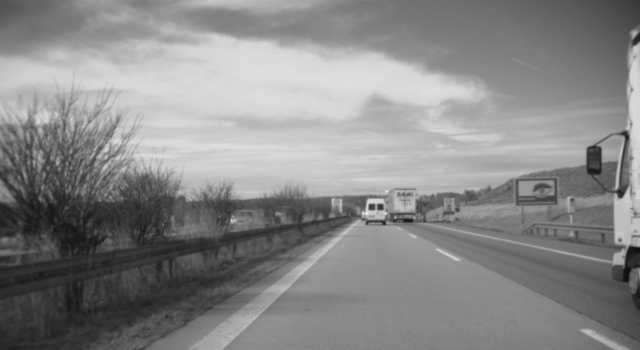} &
 \includegraphics[width=0.25\textwidth]{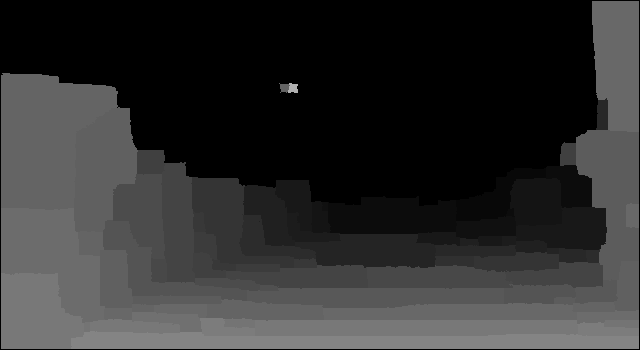} &
 \includegraphics[width=0.25\textwidth]{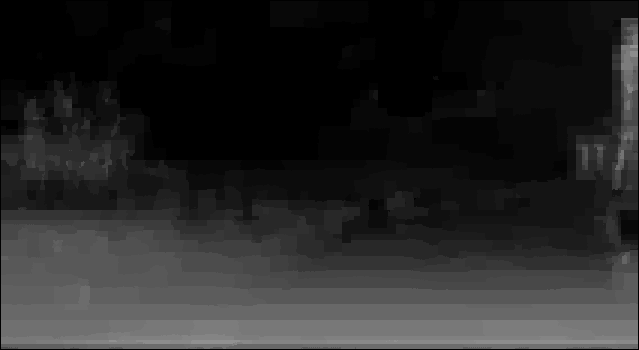}\\
 \includegraphics[width=0.25\textwidth]{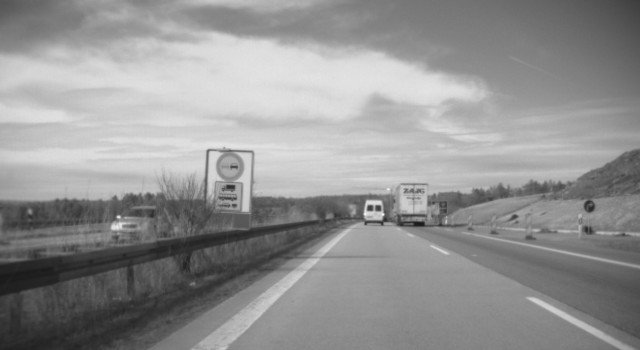} &
 \includegraphics[width=0.25\textwidth]{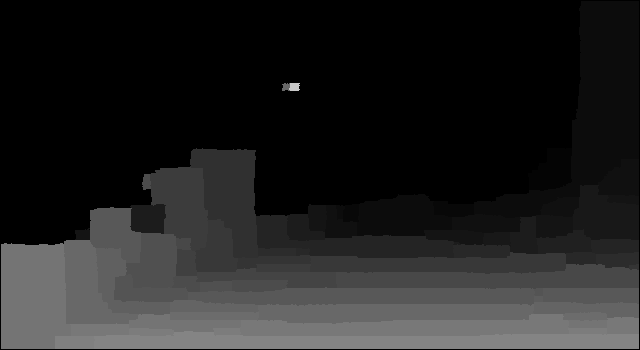} &
 \includegraphics[width=0.25\textwidth]{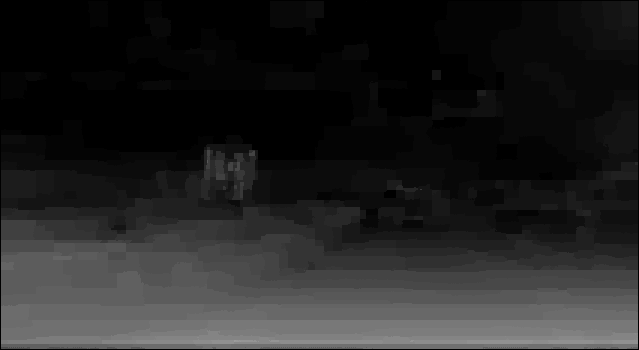}\\
 \includegraphics[width=0.25\textwidth]{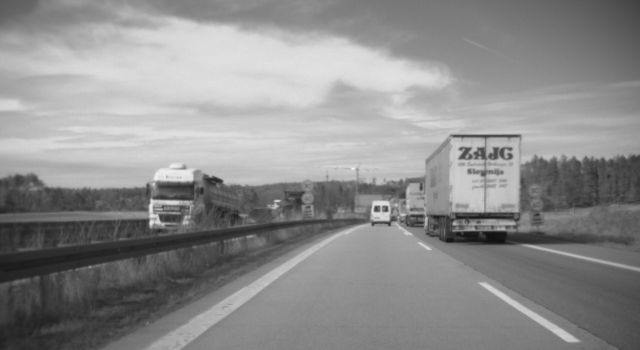} &
 \includegraphics[width=0.25\textwidth]{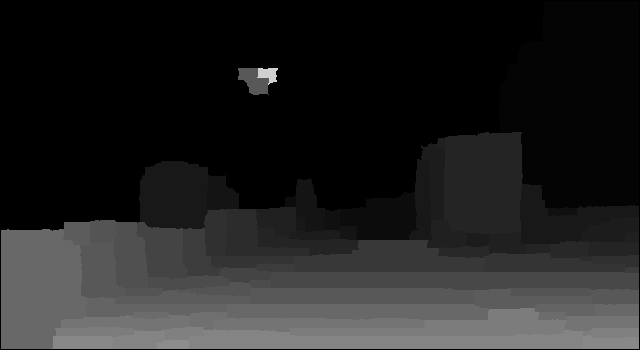} &
 \includegraphics[width=0.25\textwidth]{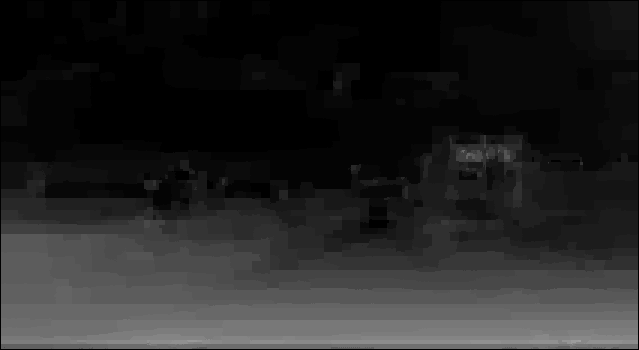}\\
 \includegraphics[width=0.25\textwidth]{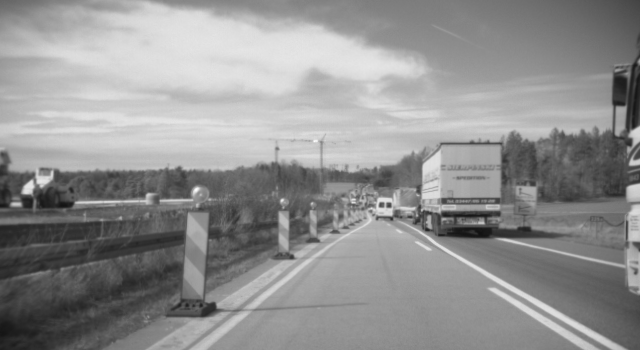} &
 \includegraphics[width=0.25\textwidth]{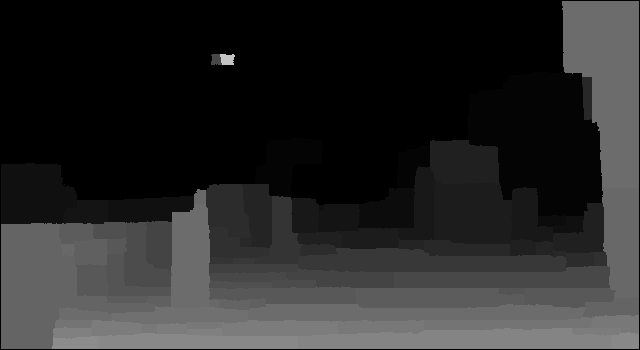} &
 \includegraphics[width=0.25\textwidth]{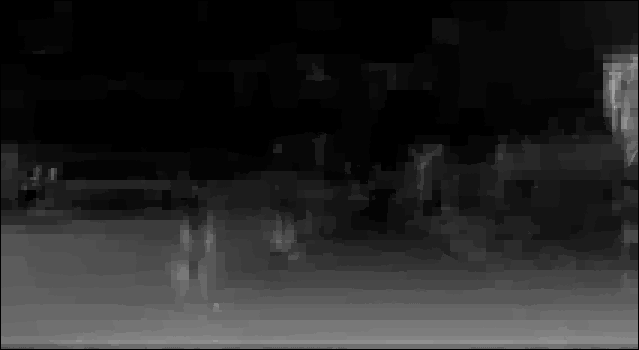}\\  
 Left image&Stereo depthmaps&Monocular depthmaps
\end{tabular}
\caption{MDE results on some testing images} 
\label{fig:im1}
\end{figure*} 

We also run our hard EM learning algorithm on the real-world road driving rectified stereo sequences acquired and provided by Daimler AG \cite{Klette07}. The dataset is the composition of seven sequences, each of them presenting different traffic, road and lighting conditions. Ground truth depth is not available. For this dataset, we trained our MDE model on six sequences and tested on the remaining sequence. As these are grayscale images, we used 15 image features at each scale. Again for testing, monocular inference was performed using loopy BP
to minimize the energy function described in~\eqref{eqn:Energya}.
We demonstrated the resulting monocular disparity maps of a few frames in the testing sequence in figure~\ref{fig:im1}. The results show that the monocular model is able to provide good depth predictions from different scenes in general, although there were still image parts which were not generalized well.

We have emphasized view prediction as a way of evaluating monocular depth estimation in the absence of any ground truth depth labels. We relate the learning problem using view prediction error to the problem of maximizing conditional likelihood in hard conditional EM learning.
In Chapter~\ref{depth-motion}, view prediction is employed again as an evaluation metrics for a structure and motion estimation framework. 
We have implemented our unsupervised CRF learning algorithm based on hard conditional EM for training MDE from stereo pairs only. The algorithm has shown to perform well by both view-prediction measures and by comparisons with ground truth labels, even though ground truth labels were not used in training.  Furthermore, performance improved (modestly) over several iterations of learning which indicates that the monocular cues used in the EM process improve depth estimation when both monocular and binocular cues are used (as is reported in \cite{andrew07}). In the second experiment, although the desired qualitative analysis could not be performed, we have demonstrated that MDE can also be used for Vision-based driving assistance.

%
%

\chapter{Depth and Motion Estimation from Stereo Sequences}
\label{depth-motion}

We introduce a unified framework for scene structure and motion estimation on road-driving stereo sequences \cite{hoang10}. 
This framework is based on the slanted-plane scene model that has become widely popular in the stereo vision community.
Our algorithm iteratively and alternately solves for scene structure and motion. Surface estimation is done using
our own slanted-plane stereo algorithm. Motion estimation is achieved by solving a MRF labeling problem
using Loopy Belief Propagation.
We show that with some specific assumptions about the motion of the camera and the scene, the motion estimation problem can be reduced to a 1D search problem along the epipolar lines.
We also propose a novel evaluation metrics, based on the notion of view prediction error previously introduced in Chapter~\ref{mno-depth-est}. This metrics can be used to evaluate the performance of structure and motion estimation algorithms on stereo sequences without ground truth data.
Experimental results on road-driving stereo sequences demonstrate that our algorithm successfully improves the view prediction error although it was not designed to directly optimize this quantity.

\section{Introduction}
\label{sec:introDepthMotion}

Multi-view video sequence is the richest form of image data for scene reconstruction. A subtype in this form that we consider very interesting are stereo video sequences. 
So far, to our knowledge, this useful data type has hardly been investigated and exploited for the purpose of recovering scene structure and motion. 
Some of the very few research work that has discussed this issue was introduced by D. Min et al. \cite{min06} and F. Huguet et al. \cite{huguet07}, in which the authors presented variational methods for scene flow estimation from calibrated stereo image sequences.
This data type is interesting for two reasons. First, as opposed to a multi-view video sequence, a stereo sequence can easily be captured using only one moving calibrated binocular camera. Second, this data type provides us with enough constraints to conveniently compute location and motion of scene points simultaneously, since coupling dense stereo matching with motion estimation helps decrease the number of unknowns per image point. More specifically, for stereo sequences, we can formulate structure and motion estimation as an energy minimization problem, in which the model is either an extension of a stereo vision model (\cite{Sze02}) to also handle scene point motion, or an extension of a Structure from Motion or optical flow model (\cite{simon09}) under a stereo setup. One example of using this latter formulation is the work presented in \cite{yzhang01}.

In this section, we follow the former approach.
Based on our slanted-plane stereo model, we develop a new algorithm for structure and motion estimation on road-driving stereo sequences. 
This framework is based on the slanted-plane scene model that has become widely popular in the stereo vision community \cite{scharstein01}.
Our algorithm iteratively and alternately solves for scene structure and motion. Surface estimation is done using
our own slanted-plane stereo algorithm. Motion estimation is achieved by solving a MRF labeling problem
using Loopy Belief Propagation.
We show that with some specific assumptions about the motion of the camera and the scene, the motion estimation problem can be reduced to a 1D search problem along the epipolar lines.
We also propose a novel evaluation metrics, based on the notion of view prediction error. 
Again, it is also much easier and more affordable to collect unlabeled stereo video data than to collect stereo video with associated ground truth (which usually requires using multiple sensors followed by a data fusion step). 
We argue that view prediction error can be used to evaluate the performance of structure and motion estimation algorithms on stereo sequences without ground truth data.
Experimental results on road-driving stereo sequences support our argument by demonstrating that our algorithm successfully improve the view prediction error although it was not designed to directly optimize this quantity. 
We believe view prediction can be used as a quantitative performance measure on unlabeled multi-view datasets in a variety of applications.

\section{Proposed approach for Structure and Motion}
\label{sec:algoDepthMotion}

\begin{figure}[h]
\centering
\includegraphics[width = 0.8 \textwidth]{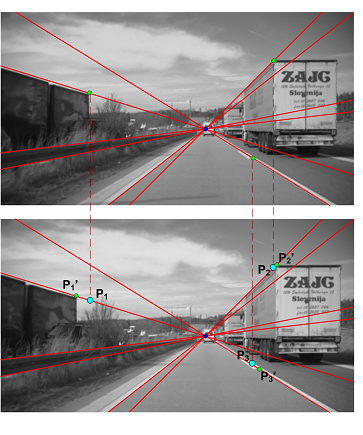}
\caption{As the camera moves forward, each pixel moves along its corresponding epipolar line. The distance it moves depends on both their depth and velocity. Since the truck on the left has highest velocity w.r.t the camera, $P_1$ moves the longest distance. On the other hand, the truck on the right has almost zero velocity w.r.t the camera, hence $P_2$ almost stands still. $P_3$ reflects the motion of the camera w.r.t the road (the scene background).}
\label{fig:epipole}
\end{figure}

In this section, we present a detailed description of our unified formulation for scene structure and motion. At each time step $t$ in the video sequence we consider the stereo pair at time $t$ and at time $t+1$ (Figure~\ref{fig:fourframes}). We only observe the three frames: $I_L^{t}$, $I_R^{t}$ and $I_L^{t+1}$ and the frame $I_R^{t+1}$ is unobserved. We want to estimate the 3-D location and motion for all pixels in $I_L^{t}$. 
To simplify the derivation of our framework, we use the following assumptions: the camera's viewing direction is the same direction of the Z axis, and all motion vectors in the scene are parallel to the Z axis. Later in section~\ref{sec:experimentDepthMotion} we show that our approach still delivers good estimates even in sequences where these assumptions are violated, e.g: when there is small camera rotation, or rotation of moving objects. 

\subsection{Connection between Disparity and Motion}
Our assumptions about the motion of the camera and the scene guarantee the following constraints:

\begin{itemize}
	\item{The image location of the epipoles in all frames of the sequence is constant:  As the epipole is simply the projection of the next camera center on the previous frame, this is obvious since the camera always moves along its viewing direction.}
	\item{From one frame to the next, a pixel translates along the epipolar line connecting that pixel and the epipole: since all 3D points in the scene move in the same direction with the Z axis, a 3D point always stay in the same 3D line which is parallel to the Z axis. Therefore the epipolar line, which is the projection of this 3D line on the image plane, stays constant, given that the epipole also stays constant. (Figure~\ref{fig:epipole})}	
\end{itemize}

\begin{figure}[h]
\centering
\includegraphics[width = 0.6 \textwidth]{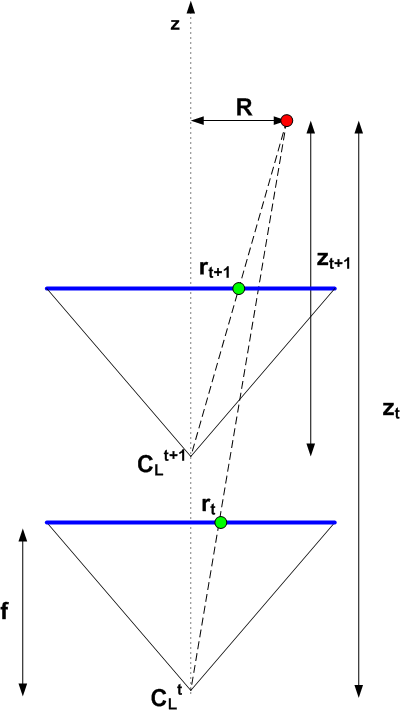}
\caption{The geometric interpretation of equation (\ref{flow1}).}
\label{fig:DispMotion}
\end{figure}

Consider a 3D point $P$ in the scene. Let $R$ be the distance from the point to the Z axis. Let $r_t$ be the 2D distance from the projection of $P$ to the epipole in the video frame at time $t$. We can derive the following mathematical relation between $r_t$, $r_{t+1}$, $v_t$ and $d_t$ as follows (This relationship is also illustrated in Figure~\ref{fig:DispMotion}):

\begin{eqnarray*}
r_t & = & \frac{Rf}{z_t} = \frac{Rf}{fh/d_t} = \frac{Rd_t}{h} \\
r_{t+1} & = & \frac{Rf}{z_{t+1}} = \frac{Rf}{z_t - \Delta z_t} \\
& = & \frac{Rf}{fh/d_t - \Delta z_t} = \frac{Rd_t}{h(1 - d_t\Delta z_t/fh)}
\end{eqnarray*}

\begin{equation}
\label{flow1}
\frac{r_{t+1}}{r_t} = \frac{1}{1 - d_t\Delta z_t/fh} = \frac{1}{1 - d_t v_t}
\end{equation}

\begin{equation}
\label{disp1}
\frac{d_{t+1}}{d_t} = \frac{r_{t+1}}{r_t} = \frac{1}{1 - d_t v_t}
\end{equation}

where: $f$ is the focal length, $z_t$ is the depth of $P$ at time $t$ (the actual distance of $P$ w.r.t the camera), $h$ is the stereo baseline, $d_t$ is the disparity of the projected pixel of $P$, $v_t$ is the velocity value of $P$. Note that in the derivation above we make use of the following relationship between $d_t$ and $z_t$: $z_t = fh/d_t$.

\subsection{3-frame Model for Depth and Motion estimation}

At each time step $t$ in the video sequence we consider the stereo pair at time $t$ and the left frame at time $t+1$. The frames are labeled as in Figure~\ref{fig:fourframes}. By equation (\ref{flow1}), we can see that if we know $d_t$ and $v_t$, then we can compute $r_{t+1}$, which means solving for the correspondence between $I_L^{t}$ and $I_L^{t+1}$. In other words, we converted the 2D optical flow field problem to the stereo disparity estimation problem and a 1D velocity assignment problem. We define the following energy functions: 

\begin{figure}[h]
\centering
\includegraphics[width = 0.9 \textwidth]{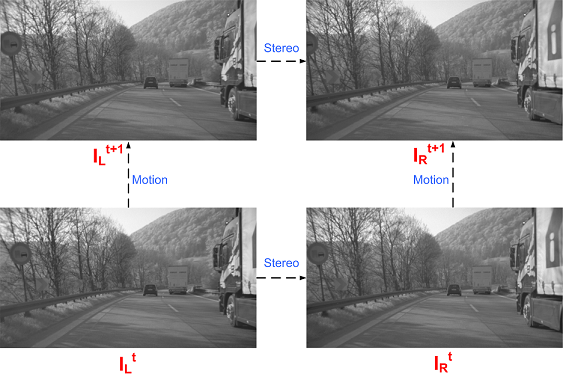}
\caption{}
\label{fig:fourframes}
\end{figure}

\begin{eqnarray}
\label{globalenergy}
E & = & E_{Stereo} + E_{S}^{v} + E_M^{v}\\
\nn
\label{smoothvel}
E_{S}^{v} & = & \sum_{P,Q}\min\left(\tau_v,\lambda_v\left|v_P-v_Q\right|\right) \\
\nn
\label{matching13}
E_M^{v} & = & \sum_{p}\sum_{k}\lambda_k\parens{\begin{array}{ll} & \phi_{k}^{t}(p) \\ - & \phi_{k}^{t+1}(p + Q(p, d_p, v_p))\end{array}}^{2}
\end{eqnarray}

where $E_{Stereo}$ is the function in (\ref{StereoEq2}), $Q(p, d_p, v_p)$ is the function mapping a pixel in $I_L^{t}$ to a pixel in $I_L^{t+1}$, and $\phi^{t}(p)$ is the k-dimensional feature vector corresponding to pixel $p$ in frame at time $t$.
The objective is to find the optimal co-assignment for depth and velocity that minimize the global energy function in (\ref{globalenergy}).
$$(d^{*}, v^{*}) = \argmin_{(d,v)}(E)$$

The optimization of $E_{Stereo}$ is introduced in section~\ref{sec:planeinference}. 

Minimizing the sum of the other two energy terms: $E_{Motion} = E_{S}^{v} + E_M^{v}$ can be considered another MRF labeling problem: we want to assign a velocity value to each superpixel such that $E_{Motion}$ is minimized. The first term $E_{S}^{v}$ penalizes the difference in velocity between two adjacent superpixels, the second term $E_M^{v}$ is the data cost of assigning a velocity $v_p$ to pixel $p$, taking into account the image data agreement between frame $I_L^{t}$ and frame $I_L^{t+1}$. We solve this MRF labeling problem by Loopy Belief Propagation. We used the max-product BP algorithm with conceptually parallel updates. In our actual implementation however, they were performed sequentially.

Finally, once we have obtained an initial estimate of $v_t$, we can fix $v_t$ and optimize the function $E_{Stereo}^{3frame} = E_{Stereo} + E_M^{v}$. Note that this function has 1 more data term compared to the previous stereo function as we can now incorporate data from $I_L^{t+1}$.
We construct an iterative algorithm alternately optimizing the disparity $d_t$ and the velocity $v_t$ for frame $I_L^{t}$. Here is the outline of our algorithm:

\begin{enumerate}
	\item {Compute the epipole at $I_L^{t}$ using a version of the 8-point algorithm.}	
	\item {Estimate the disparity map $d_t$ of $I_L^{t}$ using our stereo algorithm.}	
	\item {Use SIFT feature matching to obtain a set of initial velocity values: Each pair of matched SIFT features give us 1 initial velocity.}
	\item {Estimate $v_t$ given $d_t$: Run Loopy BP to optimize $E_{Motion}$ and assign a velocity value to each superpixel in $I_L^{t}$.}
	\item {Fixing $v_t$, reestimate $d_t$ by optimizing $E_{stereo}^{3frame}$.}
	\item {Repeat from Step 4.}
\end{enumerate}

\section{Experimental Results}
\label{sec:experimentDepthMotion}

\subsection{Evaluation Metrics}

We use view prediction as evaluation metrics for our experiments here. The idea of using view prediction error has been investigated in Section~\ref{sec:stereo-view-prediction}, in which we described stereo pair view prediction. 
View prediction is a general notion that has been motivated by the two-view learning approach in machine learning \cite{avrim98}. In two-view learning, the goal is to predict the second view given the first. In our context the goal of view prediction is to predict the unobserved data $y$ given the observed data $x$ and the latent variables $w$.We can express a probabilistic interpretation of view prediction as follows:
$$w^{*} = \argmin_{w}\sum_{i = 1}^{m}\ln 1/P_w(y|x)$$

where $P_w(y|x)$ is the conditional probability of the unobserved data given the observed data parametrized by $w$.
As our algorithm never observed $I_R^{t+1}$, in our problem setting we might define $P_w(y|x)$ as the $P(I_R^{t+1}|I_L^{t}, I_R^{t}, I_L^{t+1},d_t, v_t)$. The idea is that the more accurate we estimate $(d_t, v_t)$, the more accurate we can predict $I_R^{t+1}$.

\begin{figure}[h] 
\centering
\includegraphics[width=0.9 \textwidth]{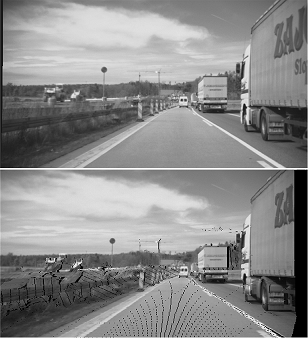}
\caption{Top: original fourth frame $I_R^{t+1}$. Bottom: predicted fourth frame $\hat{I}_R^{t+1}$. The black pixels are missing values caused by bilinear interpolation. These pixels are excluded when computing the view prediction error.}
\label{fig:viewprediction}
\end{figure}

Given the estimated depth and velocity $d_t, v_t$, we can compute the prediction $\hat{I}_R^{t+1}$ as follows:
$$I_L^{t} \stackrel{d_t, v_t}{\rightarrow} I_L^{t+1} \stackrel{d_{t+1}}{\rightarrow} \hat{I}_R^{t+1}$$

where $d_{t+1}$ is computed using (\ref{disp1}). 
An example of a predicted fourth view is shown in Figure~\ref{fig:viewprediction}. 

The view prediction error is defined as the pixel RMS between $I_R^{t+1}$ and $\hat{I}_R^{t+1}$:

\begin{eqnarray}
\label{predicterr}
Err(I_R^{t+1}, \hat{I}_R^{t+1}) & = & \sqrt{\frac{\sum_{p = 1}^{N}(I_R^{t+1}(p) - \hat{I}_R^{t+1}(p))^2}{N}}
\end{eqnarray}

We can see a clear analogy between equation (\ref{predicterr}) and equation (\ref{eqn:distortion}) from Section ~\ref{sec:stereo-view-prediction}. The difference is that the computation of the prediction $\hat{I}_R^{t+1}$ here is more complicated, involving both the estimated disparity and velocity.
In the absence of ground truth data, it is natural to use view prediction error as a useful tool for quantitative analysis, i.e. to measure how good the estimation of latent variables is. In the specific setting of our problem, the latent variables $w^{*} = (d^{*}, v^{*})$.

\subsection{Results on road-driving stereo sequences}
\begin{figure*}[t] \centering
\begin{center}
 \includegraphics[width=0.9\textwidth]{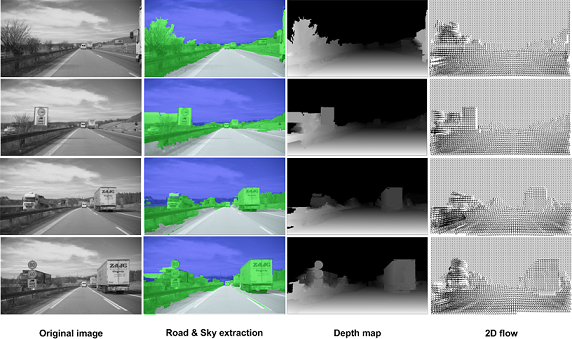}
\end{center}
\caption{Results on a Daimler road driving sequence.}
\label{fig:daimler}
\end{figure*}

We tested our algorithm on 7 gray scale road-driving stereo sequences provided by Daimler AG. Each sequence contains from 250 to 300 rectified, bias gain corrected stereo pairs, taken under different light condition and road setting. Ground truth depth and motion is not available for these datasets.
The algorithm estimates the dense map of disparity and velocity value for each left frame in the sequence. Figure~\ref{fig:daimler} demonstrates our results on several frames in a sequence.

%
%
  
%
Table~\ref{tab:score} shows the average view prediction error (\ref{predicterr}) over seven video sequences after 3 iterations of the algorithm. The results shows that the algorithm successfully improved the estimation over time. The improvement stops after three iterations. Note that the pixel intensity values are scaled to be in the range of $[0-1]$.

\begin{table} [h]
\begin{center}
\begin{tabular}{|l|l|}
\hline
 & Average View Prediction Error\\
\hline
Iter 1 & 0.0692 \\
\hline
Iter 2 & 0.0622 \\
\hline
Iter 3 & 0.0621 \\
\hline
\end{tabular}
\end{center}
\caption{Average view prediction error (in pixel intensity value) on 7 Daimler road-driving stereo sequences after each iteration of the algorithm.}
\label{tab:score}
\end{table}

\section{Unsupervised Learning of the Three-frame Model}
\label{sec:UnsupervisedDepthMotion}

In Chapter ~\ref{uspv-depth-lrn}, we demonstrated our algorithm for unsupervised learning of a highly parameterized stereo vision model involving
the shape from texture cues - training the model parameters from unlabeled stereo pair training data.
Our approach to unsupervised learning is based on maximizing conditional likelihood.

Throughout Chapter ~\ref{uspv-depth-lrn}, we used stereo
vision as a simple setting to investigate unsupervised learning. However we also argued that our approach to unsupervised learning
based on maximizing conditional likelihood can be generalized to other, maybe much more sophisticated models.

In this chapter, we formulate our unsupervised learning algorithm specifically for the model described in Section ~\ref{sec:algoDepthMotion}. 
We rewrite the 3-frame depth and motion estimation model here:

\begin{eqnarray}
\label{globalenergy2}
E & = & E_{Stereo} + E_{S}^{v} + E_M^{v}\\
\nn
\label{smoothvel2}
E_{S}^{v} & = & \sum_{P,Q}\min\left(\tau_v,\lambda_v\left|v_P-v_Q\right|\right) \\
\nn
\label{matching13_2}
E_M^{v} & = & \sum_{p}\sum_{k}\lambda_k\parens{\begin{array}{ll} & \phi_{k}^{t}(p) \\ - & \phi_{k}^{t+1}(p + Q(p, d_p, v_p))\end{array}}^{2}
\end{eqnarray}

\subsection{Unsupervised training using Hard Conditional EM/Bootstrap training}

In our three-frame model, we denote $x$ to be the frame $I_L^{t}$, $y = (y^{(1)}, y^{(2)})$, where $y^{(1)}$ is the frame $I_R^{t}$, $y^{(2)}$ is the frame $I_R^{t+1}$, and $z = (d, v)$ to be the latent variable, i.e the assignment of disparity and velocity to all pixels in $x$.

We are interested in training the parameter vector $\beta = (\beta_d, \beta_v)$ of the model in (\ref{globalenergy2}) so as to optimize the conditional probability of $y$ given $x$.

\begin{eqnarray}
\label{softopt2}
\beta^* & = & \argmax_{\beta}\;\;\sum_{i=1}^N\;\ln P_{\beta}(y_i|x_i)\\
\nn
& = & \argmax_{\beta}\;\;\sum_{i=1}^N\;\ln P_{\beta}(y_{i}^{(1)},y_{i}^{(2)}|x_i)
\end{eqnarray}

Hard conditional EM locally optimizes the following version of (\ref{softopt2}).

\begin{eqnarray}
\label{hardopt2}
\beta^* & = & \argmax_{\beta}\;\;\sum_{i=1}^N\;\max_z\;\ln P_{\beta}(y_i,z|x_i)\\
\nn
& = & \argmax_{\beta}\;\;\sum_{i=1}^N\;\max_{(d,v)}\;\ln P_{\beta}(y_i^{(1)},y_i^{(2)},d,v|x_i)
\end{eqnarray}

Specifically, hard conditional EM iterates the following two updates.

\begin{eqnarray}
\label{harde2}
z_i & := & \argmax_z P_{\beta}(y_i,z|x_i) \\
(d_i, v_i) & := & \argmax_{(d,v)} P_{\beta}(y_i^{(1)},y_i^{(2)},d,v|x_i) \\
\nn
\label{hardm2}
\beta & := & \argmax_{\beta}\; \sum_{i=1}^N \ln P_{\beta}(y_i,z_i|x_i) \\
\nn
(\beta_d, \beta_v) & := & \argmax_{(\beta_d, \beta_v)}\; \sum_{i=1}^N \ln P_{(\beta_d, \beta_v)}(y_i^{(1)},y_i^{(2)},d_i,v_i|x_i) 
\end{eqnarray}

Equation (\ref{harde2}) is the hard E step and equation (\ref{hardm2}) is the hard M step.
In the case of our three-frame model, the hard E step is implemented using the inference algorithm described in Section~\ref{sec:algoDepthMotion}. This iterative algorithm 
finds the optimal co-assignment for depth $d$ and velocity $v$ that minimize the global energy function in (\ref{globalenergy}).

To implement the hard M step, we first rewrite the probability in the right hand side of (\ref{hardm2}) as follows.

\begin{eqnarray}
\label{factorization3}
P_{(\beta_d, \beta_v)}(y^{(1)},y^{(2)},d,v|x) & = & P_{\beta_d}(y^{(1)},d|x) P_{\beta_v}(y^{(2)},v|d,x)
\end{eqnarray}

The first term, $P_{\beta_d}(y^{(1)},d|x)$ can be further factorized as demonstrated in (\ref{factorization}), (\ref{genmodela}) and (\ref{genmodelb}).

The second term, $P_{\beta_v}(y^{(2)},v|d,x)$ can also be factorized in a very similar way.

\begin{eqnarray}
\label{factorization4}
P_{\beta_v, \beta_y}(y^{(2)},v|d,x) & = & P_{\beta_y}(y^{(2)}|v,d,x) P_{\beta_v}(v|d,x)\\
\nn
\label{genmodela4}
P_{\beta_y}(y^{(2)}|v,d,x) & = & \frac{e^{(-E_M^v(x,y,v,\beta_y)\;)}}{Z_y(x,v,\beta_y)} \\
\nn
Z_y(x,v,\beta_y) & = & \sum_y\;e^{(-E_M^v(x,y,v,\beta_y)\;)} \nn
\nn
\label{genmodelb4}
P_{\beta_v}(v|d,x) & = & \frac{e^{(-E_S^v(x,v,\beta_v)\;)}}{Z_v(x,\beta_v)}  \\
\nn
Z_v(x,\beta_v) & = & \sum_v\;e^{(-E_S^v(x,v,\beta_v)\;)}\nonumber
\end{eqnarray}

Given this factorization, the hard M step in equation (\ref{hardm2}) then can be implemented the same way as described in equations (\ref{hardma}), (\ref{hardmb}) and (\ref{grad}).

Specifically, let $E_i(v)$ abbreviate $E_S^v(x,v,\beta_v)$, the contrastive divergence update equation for $\beta_v$ can be expressed as follows.

\begin{equation}
\label{grad}
\nabla_{\beta_v} L = \sum_{i=1}^N \parens{\expectsub{v \sim P_v(v|x_i,\beta)}{\nabla_{\beta_v} E_i(v)} - \nabla_{\beta_v} E_i(v_i)}
\end{equation}

A similar equation holds for $\beta_y$ and $E_M^v(x,y,v,\beta_y)$.

\subsection{Unsupervised training using fourth view prediction}

Here we are interested in training the parameter vector of the model in (\ref{globalenergy2}) so as to minimize the fourth view prediction error (\ref{predicterr}).
First, assume that we have a way to compute the gradient of the view prediction error function (\ref{predicterr}) with respect to the latent variables $d, v$. Using contrastive divergence can be one of the possibilities. We call this gradient $G$.
Second, we can also assume that the gradient of the latent variables $d$ and $v$ with respect to the model parameters $\beta$ can also be computed. Let $g$ be this gradient.

By applying the chain rule, the gradient of the view prediction error function with respect to $\beta$ can be computed as follows.

\begin{equation}
\frac{\partial Err(I_R^{t+1}, \hat{I}_R^{t+1})}{\partial \beta} = G^{T}g
\end{equation}

We can then use gradient descent to find the optimal $\beta^{*}$ optimizing the fourth view prediction error (\ref{predicterr}).

\section{Conclusion}
\label{sec:concludeDepthMotion}
In this chapter we emphasized view prediction as a way of evaluating scene structure and motion estimation from stereo video data in the absence of any ground truth labels.
We introduced a new algorithm for structure and motion estimation on road-driving stereo sequences. Based on specific assumptions about the motion of the camera and the scene, we can reduced the 2D optical flow problem to a 1D velocity value problem. Our algorithm iteratively and alternately solve for structure and motion. Scene structure estimation is done using our own plane-based stereo algorithm. Velocity estimation is completed by solving a MRF labeling problem using Loopy BP. Experiments on road-driving stereo sequences showed encouraging results, even with video sequences where the scene and camera motion do not fully comply with our assumptions.

Performance analysis was done using our novel evaluation metrics based on the notion of view prediction error. 
We argue that this evaluation metrics is quite appropriate for algorithms working with stereo sequences as well as multiple view image data, when ground truth data is not available. 
Our experimental results support this argument.
We used hand-tuned parameters for our model in this chapter. Ideally, these parameters should be estimated by an automated method, usually through learning using a labeled training data set. With view prediction error, we believe the problem we solve in this chapter can be another setting where we can apply the unsupervised learning approach based on maximizing conditional likelihood. In other words, the model parameters can be learned using only unlabeled stereo video data.

Since we could relate the learning problem using view prediction error to the problem of maximizing conditional likelihood in hard conditional EM learning, as demonstrated in Chapter~\ref{mno-depth-est}, in addition to the fact that the problem we address here is also modeled as CRFs, there is obviously the possibility of using this problem as another setting where we can apply our unsupervised CRF learning approach.
It is hoped that a general notion of view prediction will eventually facilitate unsupervised learning in a variety of cases with mutual information between views. 

\chapter{Conclusion}
\label{conclude}

In this thesis, we demonstrated machine learning techniques to build computer vision systems 3-D scene geometry recovery from images.
Scene structure can be recovered from different types of input image data. In this thesis we focused on three data types: two-view stereo pairs, single image, and stereo video sequences.
Building these systems requires algorithms for doing inference as well as learning
the parameters of MRFs or CRFs. 

For inference in systems with continuous valued variables, or
discrete-valued variables with very large domains, it is impossible to directly use a standard
discrete MRF inference techniques such as Loopy BP, graph cuts or tree-reweighted message passing.
For such systems, we proposed to use a generic Particle-based Belief Propagation (PBP) algorithm closely related
to previous work, which we then formulated specifically for our MRF labeling problems.
Although we only described a specific use of
this generic PBP algorithm, we believe it can be used as an approximate inference scheme
for a wide variety of problems that can be formulated by a probabilistic graphical model,
especially those containing many random variables with very large or continuous domains.
Our approach creates a set of particles for each variable, representing samples from the posterior marginal.
The algorithm then continues to improve the current marginal estimate by constructing
a new sampling distribution and draw new sets of particles. It's shown by experiments
that the algorithm is consistent, i.e. approaches the true values of the message and belief
functions with finite samples. Besides accuracy, PBP algorithm also provides good
estimate for properties of the distribution, and a representation of state uncertainty.
Although an "adaptive" choice for the sampling distribution, and such iterative resampling
processes require further work to analyze, our results seem to support the
notion of sampling from the current marginal estimates themselves, whether from fitted
distributions or via a series of MCMC steps.

Learning in MRF involves estimating the optimal parameters of the energy model. 
For learning, unlike previous work, we trained our system without using ground-truth labeled data. 
We introduce the unsupervised CRF learning algorithm, based on maximizing conditional likelihood using hard conditional EM.
We demonstrated the application of our unsupervised CRF learning algorithm for different scene geometry problems

In unsupervised learning one usually formulates a parameterized probability model and seeks parameter values maximizing
the likelihood of the unlabeled training data. Specifically for stereo vision, the model defines the conditional probability of the right image given the left.
We introduced a slanted-plane stereo vision model in which we used
a fixed over-segmentation to segment the left image into coherent regions called superpixels.
We then assign a disparity plane for each superpixel. We formulated the problem of
inferring plane parameters as a MRF labeling problem, which can be solved by an energy
minimization method. The MRF is a graphical model in which superpixels define nodes
and the adjacency between superpixels define edges. Our stereo energy function balances
between a data matching term and a smoothness term.
We then used our learning algorithm to train this highly parameterized stereo vision model involving
the shape from texture cues.
We showed that this unsupervised learning algorithm implicitly trains
shape from texture monocular surface orientation cues. We demonstrated that training monocular cues from stereo pair data improved stereo depth estimation. Our stereo model with texture cues, only by unsupervised training, outperformed the results in related work on the same stereo dataset.
Stereo vision provides perhaps the simplest setting in which to study unsupervised learning. 
We have formulated an approach to unsupervised learning based on maximizing conditional likelihood and demonstrated
its use for unsupervised learning of stereo depth with monocular depth cues. Ultimately
we are interested in learning highly parameterized sophisticated models including,
perhaps, models of surface types, shape from shading, albedo smoothness priors, lighting
smoothness priors, and even object pose models. We believe that unsupervised learning
based on maximizing conditional likelihood can be scaled to much more sophisticated models
than those demonstrated in this paper.

We also applied our unsupervised CRF learning approach to the problem of monocular depth estimation (MDE).
We learned the MDE model using stereo pairs without ground truth depth maps. 
While most of the performance of the monocular predictor is achieved by training on the initial stereo depth estimates, a modest improvement is seen for both view prediction and for ground truth prediction at later EM iterations.  The MDE performance is similar to that reported in previous work on the same dataset despite the absence of ground truth in learning.

In this thesis, we also addressed the use of an interesting image data type - stereo video sequences. Specifically, for stereo sequences, we can formulate structure and motion estimation as an energy minimization problem, in which the model is either an extension of a dense stereo vision model to also handle scene point motion.
In this thesis, based on the constructed slanted-plane stereo model, we introduced a new algorithm for structure and motion estimation on road-driving stereo sequences. 
Based on specific assumptions about the motion of the camera and the scene, we can reduce the 2D optical flow problem to a 1D velocity value problem.
Our algorithm iteratively and alternately solve for structure and motion. Surface estimation is done using our own slanted-plane stereo algorithm. Velocity estimation is achieved by solving a MRF labeling problem using Loopy BP. Performance analysis was done using our novel evaluation metrics based on the notion of view prediction error.
Experiments on road-driving stereo sequences showed encouraging results, even with video sequences where the scene and camera motion do not fully comply with our assumptions.
We emphasized view prediction as a way of evaluating scene structure and motion estimation from stereo video data in the absence of any ground truth labels.
We argue that this evaluation metrics is quite appropriate for algorithms working with stereo sequences as well as multiple view image data, when ground truth data is
not available. Our experimental results support this argument. We used hand-tuned parameters for our model in this paper. Ideally, these parameters should be estimated
by an automated method, usually through learning using a labeled training data set. Our thesis addressed a theoretical discussion regarding this issue, although we did not actually implement the idea. With view prediction error, we believe the scene structure and motion estimation from stereo video problem can be another setting where we can apply our unsupervised learning approach based on maximizing conditional likelihood. In other words, the model parameters can be learned using only unlabeled stereo video data.

{
\bibliographystyle{plainnat}
\pagebreak
\bibliography{main}

\begin{thebibliography}{59}
\providecommand{\natexlab}[1]{#1}
\providecommand{\url}[1]{\texttt{#1}}
\expandafter\ifx\csname urlstyle\endcsname\relax
  \providecommand{\doi}[1]{doi: #1}\else
  \providecommand{\doi}{doi: \begingroup \urlstyle{rm}\Url}\fi

\bibitem[Arulampalam et~al.(2002)Arulampalam, Maskell, Gordon, and
  Clapp]{arulampalam02}
M.~S. Arulampalam, S.~Maskell, N.~Gordon, and T.~Clapp.
\newblock A tutorial on particle filters for online nonlinear/non-{G}aussian
  {B}ayesian tracking.
\newblock \emph{IEEE Trans. SP}, 50\penalty0 (2):\penalty0 174--188, February
  2002.

\bibitem[Ashutosh~Saxena(2005)]{ashu05}
Andrew Y.~Ng Ashutosh~Saxena, Sung H.~Chung.
\newblock Learning depth from single monocular images.
\newblock In \emph{Advances in Neural Information Processing Systems}, 2005.

\bibitem[Ashutosh~Saxena and Ng(2007{\natexlab{a}})]{andrew07}
Jamie~Schulte Ashutosh~Saxena and Andrew~Y. Ng.
\newblock Depth estimation using monocular and stereo cues.
\newblock In \emph{IJCAI}, 2007{\natexlab{a}}.

\bibitem[Ashutosh~Saxena and Ng(2007{\natexlab{b}})]{ashu07}
Min~Sun Ashutosh~Saxena and Andrew~Y. Ng.
\newblock 3-d depth reconstruction from a single still image.
\newblock \emph{International Journal of Computer Vision}, 2007{\natexlab{b}}.

\bibitem[Birchfield and Tomasi(1999)]{BirchfieldTomasi}
Stan Birchfield and Carlo Tomasi.
\newblock Multiway cut for stereo and motion with slanted surfaces.
\newblock In \emph{International Conference on Computer Vision}, 1999.

\bibitem[Blum and Mitchell(1998)]{avrim98}
Avrim Blum and Tom Mitchell.
\newblock Combining labeled and unlabeled data with co-training.
\newblock In \emph{Proceedings of the Workshop on Computational Learning
  Theory}, 1998.

\bibitem[Carreira-Perpi{\~n}{\'a}n and Hinton(2005)]{contrastiveB}
M.~A. Carreira-Perpi{\~n}{\'a}n and G.E. Hinton.
\newblock On contrastive divergence learning.
\newblock In \emph{10th Int. Workshop on Artificial Intelligence and Statistics
  (AISTATS 2005)}, 2005.

\bibitem[Dalal and Triggs(2005{\natexlab{a}})]{Dalal}
N.~Dalal and B.~Triggs.
\newblock Histograms of oriented gradients for human detection.
\newblock In \emph{IEEE Computer Society Conference on Computer Vision and
  Pattern Recognition}, pages I: 886--893, 2005{\natexlab{a}}.

\bibitem[Dalal and Triggs(2005{\natexlab{b}})]{dalal05}
Navneet Dalal and Bill Triggs.
\newblock Histograms of oriented gradients for human detection.
\newblock In \emph{IEEE Computer Society Conference on Computer Vision and
  Pattern Recognition}, 2005{\natexlab{b}}.

\bibitem[Derek~Hoiem(2005)]{Hoiem05}
Martial~Hebert Derek~Hoiem, Alexei A.~Efros.
\newblock Geometric context from a single image.
\newblock In \emph{International Conference on Computer Vision}, 2005.

\bibitem[Doucet et~al.(2001)Doucet, de~Freitas, and Gordon]{doucet01}
A.~Doucet, N.~de~Freitas, and N.~Gordon, editors.
\newblock \emph{Sequential {M}onte {C}arlo Methods in Practice}.
\newblock Springer-Verlag, New York, 2001.

\bibitem[Felzenszwalb et~al.(2008)Felzenszwalb, McAllester, and
  Ramanan]{Felz08}
Pedro Felzenszwalb, David McAllester, and Deva Ramanan.
\newblock A discriminatively trained, multiscale, deformable part model.
\newblock In \emph{IEEE Computer Society Conference on Computer Vision and
  Pattern Recognition}, 2008.

\bibitem[Felzenszwalb and Huttenlocher(2004)]{Pedro04}
Pedro~F. Felzenszwalb and Daniel~P. Huttenlocher.
\newblock Efficient graph-based image segmentation.
\newblock \emph{International Journal of Computer Vision}, 59\penalty0 (2),
  September 2004.

\bibitem[Felzenszwalb and Huttenlocher(2006)]{Pedro06}
Pedro~F. Felzenszwalb and Daniel~P. Huttenlocher.
\newblock Efficient belief propagation for early vision.
\newblock \emph{Int. J. Comput. Vision}, 70\penalty0 (1):\penalty0 41--54,
  2006.
\newblock ISSN 0920-5691.
\newblock \doi{http://dx.doi.org/10.1007/s11263-006-7899-4}.

\bibitem[Fischler and Bolles(1981)]{RANSAC}
M.A. Fischler and R.C. Bolles.
\newblock Random sample consensus: A paradigm for model fitting with
  applications to image analysis and automated cartography.
\newblock \emph{Communications of the ACM}, 24\penalty0 (6):\penalty0 381--395,
  June 1981.

\bibitem[Geman and Geman(1984)]{geman84}
S.~Geman and D.~Geman.
\newblock Stochastic relaxation, gibbs distributions, and the bayesian
  restoration of images.
\newblock \emph{IEEE Transactions on Pattern Analysis and Machine
  Intelligence}, pages 721--741, 1984.

\bibitem[Godsill and Clapp(2001)]{godsill01}
S.~Godsill and T.~Clapp.
\newblock Improvement strategies for {M}onte {C}arlo particle filters.
\newblock In J.~F. G. De~Freitas A.~Doucet and N.~J. Gordon, editors,
  \emph{Sequential {M}onte {C}arlo Methods in Practice}. Springer-Verlag, New
  York, 2001.

\bibitem[Hammersley and Clifford(1971)]{HCTheorem}
JM~Hammersley and P~Clifford.
\newblock Markov fields on finite graphs and lattices.
\newblock Unpublished Manuscript, 1971.

\bibitem[Hartley and Zisserman(2004)]{Hartley04}
R.~I. Hartley and A.~Zisserman.
\newblock \emph{Multiple View Geometry in Computer Vision}.
\newblock Cambridge University Press, 2nd edition, 2004.

\bibitem[Hastings()]{hastings70}
W.~K Hastings.
\newblock Monte carlo sampling methods using markov chains and their
  applications.
\newblock \emph{Biometrika}, pages 97--109.

\bibitem[Hinton(2002)]{contrastiveA}
Geoffrey~E. Hinton.
\newblock Training products of experts by minimizing contrastive divergence.
\newblock \emph{Neural Computation}, 14\penalty0 (8):\penalty0 1771--1800,
  2002.

\bibitem[Huguet and Devernay(2007)]{huguet07}
F.~Huguet and F.~Devernay.
\newblock A variational method for scene flow estimation from stereo sequences.
\newblock In \emph{International Conference on Computer Vision}, 2007.

\bibitem[Ihler and McAllester(2009)]{ihler09}
A.~Ihler and D.~McAllester.
\newblock Particle belief propagation.
\newblock In \emph{International Conference on Artificial Intelligence and
  Statistics}, 2009.

\bibitem[Irsara and Fusiello(2006)]{Fusiello}
L.~Irsara and A.~Fusiello.
\newblock Quasi-euclidean uncalibrated epipolar rectification.
\newblock In \emph{Rapporto di Ricerca RR 43/2006, Dipartimento di Informatica
  - Universit{\`a} di Verona}, 2006.

\bibitem[J.~Yedidia and Weiss(2000)]{yedidia00}
W.~Freeman J.~Yedidia and Y.~Weiss.
\newblock Generalized belief propagation.
\newblock In \emph{Advances in Neural Information Processing Systems}, 2000.

\bibitem[Kaucic et~al.(2001)Kaucic, Hartley, and Dano]{kaucic01}
R.~Kaucic, R.~Hartley, and N.~Dano.
\newblock Plane-based projective reconstruction.
\newblock In \emph{International Conference on Computer Vision}. IEEE Computer
  Society, 2001.

\bibitem[Khan et~al.(2005)Khan, Balch, and Dellaert]{khan05}
Z.~Khan, T.~Balch, and F.~Dellaert.
\newblock {MCMC}-based particle filtering for tracking a variable number of
  interacting targets.
\newblock \emph{IEEE Transactions on Pattern Analysis and Machine
  Intelligence}, pages 1805--1918, November 2005.

\bibitem[Kindermann and Snell(1980)]{MRF}
Ross Kindermann and J.~Laurie Snell.
\newblock \emph{Markov Random Fields and Their Applications}.
\newblock Americal Mathematical Society, 1980.

\bibitem[Klaus et~al.(2006)Klaus, Sormann, and Karner]{KSK06}
A.~Klaus, M.~Sormann, and K.~Karner.
\newblock Segment-based stereo matching using belief propagation and a
  self-adapting dissimilarity measure.
\newblock In \emph{International Conference on Pattern Recognition}, 2006.

\bibitem[Koller et~al.(1999)Koller, Lerner, and Angelov]{koller99}
D.~Koller, U.~Lerner, and D.~Angelov.
\newblock A general algorithm for approximate inference and its application to
  hybrid {B}ayes nets.
\newblock In \emph{Conference on Uncertainty in Artificial Intelligence 15},
  pages 324--333, 1999.

\bibitem[Kong and Tao(2004)]{Kong04}
Dan Kong and Hai Tao.
\newblock A method for learning matching errors in stereo computation.
\newblock In \emph{BMVC}, 2004.

\bibitem[Lafferty et~al.(2001)Lafferty, McCallum, and Pereira]{Pereira01}
John Lafferty, Andrew McCallum, and Fernando Pereira.
\newblock Conditional random fields: {P}robabilistic models for segmenting and
  labeling sequence data.
\newblock In \emph{International Conference on Machine Learning}, pages
  282--289. Morgan Kaufmann, San Francisco, CA, 2001.

\bibitem[Li(1995)]{MRFVision}
Stan~Z. Li.
\newblock \emph{Markov Random Field Modeling in Computer Vision}.
\newblock Springer-Vewrlag, 1995.

\bibitem[Liu and Klette(2007)]{Klette07}
Zhifeng Liu and Reinhard Klette.
\newblock Performance evaluation of stereo and motion analysis on rectified
  image sequences.
\newblock In \emph{CITR-TR-207, The University of Auckland, ISSN 1178-3524},
  2007.

\bibitem[Lourakis and Argyros(2004)]{lourakis04}
M.I.A. Lourakis and A.A. Argyros.
\newblock The design and implementation of a generic sparse bundle adjustment
  software package based on the levenberg-marquardt algorithm.
\newblock Technical Report 340, Institute of Computer Science - FORTH,
  Heraklion, Crete, Greece, Aug. 2004.
\newblock Available from \verb+http://www.ics.forth.gr/~lourakis/sba+.

\bibitem[Lowe(2004)]{SIFT}
David~G. Lowe.
\newblock Distinctive image features from scale-invariant keypoints.
\newblock \emph{International Journal of Computer Vision}, 60:\penalty0
  91--110, 2004.

\bibitem[Metropolis and Ulam(1949)]{metropolis49}
N.~Metropolis and S.~Ulam.
\newblock The monte carlo method.
\newblock \emph{J. Amer. Statist. Assoc}, 1949.

\bibitem[Min and Sohn(2006)]{min06}
D.~Min and K.~Sohn.
\newblock Edge-preserving simultaneous joint motion disparity estimation.
\newblock In \emph{International Conference on Pattern Recognition}, 2006.

\bibitem[Neal et~al.(2003)Neal, Beal, and Roweis]{neal03}
R.~M. Neal, M.~J. Beal, and S.~T. Roweis.
\newblock Inferring state sequences for non-linear systems with embedded hidden
  {M}arkov models.
\newblock In \emph{Advances in Neural Information Processing Systems 16}, 2003.

\bibitem[Pearl(1988)]{pearl88}
J.~Pearl.
\newblock \emph{Probabilistic Reasoning in Intelligent Systems}.
\newblock Morgan Kaufman, San Mateo, 1988.

\bibitem[Q.~Yang and Nistér(2008)]{yang08}
R.~Yang H.~Stewénius Q.~Yang, L.~Wang and D.~Nistér.
\newblock Stereo matching with color-weighted correlation, hierarchical belief
  propagation and occlusion handling.
\newblock \emph{IEEE Transactions on Pattern Analysis and Machine
  Intelligence}, 31\penalty0 (3), 2008.

\bibitem[Quattoni et~al.(2007)Quattoni, Wang, Morency, Collins, and
  Darrell]{Quattoni07}
Ariadna Quattoni, Sybor Wang, Louis-Philippe Morency, Michael Collins, and
  Trevor Darrell.
\newblock Hidden conditional random fields.
\newblock \emph{IEEE Transactions on Pattern Analysis and Machine
  Intelligence}, 29\penalty0 (10):\penalty0 1848--1852, October 2007.

\bibitem[R.~Zhang and Shah.(1999)]{zhang99}
J.~Cryer R.~Zhang, P.~Tsai and M.~Shah.
\newblock Shape from shading: A survey.
\newblock \emph{IEEE Transactions on Pattern Analysis and Machine
  Intelligence}, pages 690--706, 1999.

\bibitem[Roth and Black(2005)]{roth05}
Stefan Roth and Michael~J. Black.
\newblock Fields of experts: A framework for learning image priors.
\newblock In \emph{IEEE Computer Society Conference on Computer Vision and
  Pattern Recognition}, 2005.

\bibitem[S.~Belongie and Puzicha(2002)]{belongie02}
J.~Malik S.~Belongie and J.~Puzicha.
\newblock Shape matching and object recognition using shape contexts.
\newblock \emph{IEEE Transactions on Pattern Analysis and Machine
  Intelligence}, 2002.

\bibitem[Scharstein and Szeliski(2002)]{Sze02}
D.~Scharstein and R.~Szeliski.
\newblock A taxonomy and evaluation of dense two-frame stereo correspondence
  algorithms.
\newblock \emph{International Journal of Computer Vision}, 2002.

\bibitem[Scharstein et~al.(2001)Scharstein, Szeliski, and Zabih]{scharstein01}
D.~Scharstein, R.~Szeliski, and R.~Zabih.
\newblock http://vision.middlebury.edu/stereo/, 2001.

\bibitem[Scharstein and Pal(2007)]{scharstein07}
Daniel Scharstein and Chris Pal.
\newblock Learning conditional random fields for stereo.
\newblock In \emph{IEEE Computer Society Conference on Computer Vision and
  Pattern Recognition}, 2007.

\bibitem[Simon~Baker and Szeliski(2009)]{simon09}
J.P. Lewis Stefan Roth Michael J.~Black Simon~Baker, Daniel~Scharstein and
  Richard Szeliski.
\newblock A database and evaluation methodology for optical flow.
\newblock Technical Report MSR-TR-2009-179, December 2009.

\bibitem[Snavely et~al.(2006)Snavely, Seitz, and Szeliski]{Szeliski06}
N.~Snavely, S.~M. Seitz, and R.~Szeliski.
\newblock Photo tourism: Exploring photo collections in {3D}.
\newblock In \emph{ACM Transactions on Graphics}, pages 25(3):835--846, August
  2006.

\bibitem[Sun et~al.(2003)Sun, Zheng, and Shum]{Jian03}
Jian Sun, Nan-Ning Zheng, and Heung-Yeung Shum.
\newblock Stereo matching using belief propagation.
\newblock \emph{IEEE Transactions on Pattern Analysis and Machine
  Intelligence}, 25\penalty0 (7):\penalty0 787--800, 2003.
\newblock ISSN 0162-8828.
\newblock \doi{http://dx.doi.org/10.1109/TPAMI.2003.1206509}.

\bibitem[Triggs et~al.(1999)Triggs, McLauchlan, Hartley, and
  Fitzgibbon]{Trigg99}
B.~Triggs, P.~McLauchlan, R.~Hartley, and A.~Fitzgibbon.
\newblock Bundle adjustment - a modern synthesis.
\newblock In \emph{Vision Algorithms'99}, 1999.

\bibitem[Trinh and McAllester(2009)]{hoang09}
Hoang Trinh and David McAllester.
\newblock Unsupervised learning of stereo vision with monocular cues.
\newblock In \emph{British Machine Vision Conference}, 2009.

\bibitem[Trinh and McAllester(2010)]{hoang10}
Hoang Trinh and David McAllester.
\newblock Structure and motion from road-driving stereo sequences.
\newblock In \emph{IEEE Workshop on 3D Information Extraction for Video
  Analysis and Mining - CVPR 2010}, 2010.

\bibitem[van~der Merwe et~al.(2001)van~der Merwe, de~Freitas, Doucet, and
  Wan]{vandermerwe00}
R.~van~der Merwe, N.~de~Freitas, A.~Doucet, and E.~Wan.
\newblock The unscented particle filter.
\newblock In \emph{Advances in Neural Information Processing Systems 13},
  December 2001.

\bibitem[Wang and Zheng(2008)]{WangZheng08}
Z.~Wang and Z.~Zheng.
\newblock A region based stereo matching algorithm using cooperative
  optimization.
\newblock In \emph{IEEE Computer Society Conference on Computer Vision and
  Pattern Recognition}, 2008.

\bibitem[Y.~Boykov and Zabih(2001)]{boykov01}
O.~Veksler Y.~Boykov and R.~Zabih.
\newblock Fast approximate energy minimization via graph cuts.
\newblock \emph{IEEE Transactions on Pattern Analysis and Machine
  Intelligence}, pages 1222--1239, 2001.

\bibitem[Zhang and Seitz(2007)]{Seitz05}
Li~Zhang and Steven~M. Seitz.
\newblock Estimating optimal parameters for mrf stereo from a single image
  pair.
\newblock \emph{IEEE Transactions on Pattern Analysis and Machine
  Intelligence}, 29\penalty0 (2), 2007.
\newblock based on "Parameter Estimation for MRF Stereo", CVPR 2005.

\bibitem[Zhang and Kambhamettu(2001)]{yzhang01}
Y.~Zhang and C.~Kambhamettu.
\newblock On 3d scene flow and structure estimation.
\newblock In \emph{IEEE Computer Society Conference on Computer Vision and
  Pattern Recognition}, 2001.

\end{thebibliography}
}

\end{document}